\documentclass[11pt,letterpaper]{article}

\newif\ifextrasafe\extrasafetrue

\usepackage[margin=1in,top=1in,bottom=1in]{geometry}

\usepackage[T1]{fontenc}
\usepackage[utf8]{inputenc}
\usepackage{lmodern}
\usepackage{microtype}
\usepackage{amsmath}
\usepackage{amssymb}

\usepackage{booktabs}
\usepackage{array}
\usepackage{tabularx}
\usepackage{longtable}
\usepackage{multirow}
\usepackage{makecell}

\usepackage{graphicx}
\usepackage{seqsplit}
\usepackage{float}
\usepackage{fontspec}
\IfFontExistsTF{[NotoSansCJKsc-Regular.otf]}
  {\newfontfamily\cjkfont[Path=./, Extension=.otf]{NotoSansCJKsc-Regular}}
  {\IfFontExistsTF{Noto Sans CJK SC}
    {\newfontfamily\cjkfont{Noto Sans CJK SC}}
    {\IfFontExistsTF{Noto Serif CJK SC}
      {\newfontfamily\cjkfont{Noto Serif CJK SC}}
      {\IfFontExistsTF{Noto Sans CJK TC}
        {\newfontfamily\cjkfont{Noto Sans CJK TC}}
        {\IfFontExistsTF{FandolSong-Regular.otf}
          {\newfontfamily\cjkfont[Extension=.otf]{FandolSong-Regular}}
          {\IfFontExistsTF{Droid Sans Fallback}
            {\newfontfamily\cjkfont{Droid Sans Fallback}}
            {\providecommand\cjkfont{}}}}}}}
\newcommand{\cjk}[1]{{%
  \cjkfont%
  \XeTeXlinebreaklocale "zh"%
  \XeTeXlinebreakskip = 0pt plus 1pt minus 0.1pt%
  \sloppy #1}}
\usepackage[table]{xcolor}\usepackage{colortbl}
\usepackage{caption}
\usepackage[section]{placeins}
\usepackage{subcaption}

\usepackage{xcolor}
\definecolor{tfgnblue}{RGB}{31,119,180}
\definecolor{baselinered}{RGB}{214,39,40}
\definecolor{shadegray}{RGB}{245,245,245}
\definecolor{linkblue}{RGB}{0,80,180}

\def\TFGNpdftitle{TFGN: Task-Free, Replay-Free Continual Pre-Training Without Catastrophic Forgetting at LLM Scale}
\def\TFGNpdfsubject{Architectural overlay for replay-free, task-free continual pre-training of transformer language models at LLM scale. Under provisional patent (US filed 2026-04-22).}
\usepackage[bookmarks=true,bookmarksopen=true]{hyperref}
\hypersetup{
  colorlinks=true,
  linkcolor=linkblue,
  urlcolor=linkblue,
  citecolor=linkblue,
  pdftitle=\TFGNpdftitle,
  pdfauthor={Anurup Ganguli},
  pdfsubject=\TFGNpdfsubject,
  pdfkeywords={continual learning, catastrophic forgetting, transformer, language model, backward transfer, replay-free, task-free}
}
\usepackage{url}

\usepackage{enumitem}
\setlist{nosep, leftmargin=*}

\usepackage{titlesec}
\titlespacing*{\section}{0pt}{12pt}{6pt}
\titlespacing*{\subsection}{0pt}{8pt}{4pt}
\titlespacing*{\subsubsection}{0pt}{6pt}{3pt}

\usepackage[numbers,sort&compress]{natbib}
\usepackage[most]{tcolorbox}
\tcbuselibrary{breakable, skins}
\newtcolorbox{tfgnbox}[1][]{
  colback=tfgnblue!8,
  colframe=tfgnblue!50!black,
  boxrule=0.4pt,
  arc=2pt,
  left=4pt,right=4pt,top=3pt,bottom=3pt,
  fontupper=\small,
  #1
}
\newtcolorbox{baselinebox}[1][]{
  colback=baselinered!8,
  colframe=baselinered!50!black,
  boxrule=0.4pt,
  arc=2pt,
  left=4pt,right=4pt,top=3pt,bottom=3pt,
  fontupper=\small,
  #1
}
\definecolor{emissdriftborder}{RGB}{180, 60, 60}
\definecolor{emissdriftbg}{RGB}{253, 240, 240}
\definecolor{emisscoherentborder}{RGB}{40, 80, 160}
\definecolor{emisscoherentbg}{RGB}{240, 244, 252}
\definecolor{emissrowheaderbg}{RGB}{60, 70, 85}
\newtcolorbox{emissdriftcard}[1]{%
  enhanced, breakable,
  colback=emissdriftbg, colframe=emissdriftborder,
  arc=1pt, boxrule=0.4pt, top=4pt, bottom=4pt, left=8pt, right=8pt,
  before skip=0.3em, after skip=0.3em,
  attach boxed title to top left={xshift=4pt, yshift=-6pt},
  boxed title style={colback=emissdriftborder, colframe=emissdriftborder,
    sharp corners, arc=0pt, top=1pt, bottom=1pt, left=4pt, right=4pt},
  coltitle=white,
  fonttitle=\scriptsize\bfseries,
  title={#1}
}
\newtcolorbox{emisscoherentcard}[1]{%
  enhanced, breakable,
  colback=emisscoherentbg, colframe=emisscoherentborder,
  arc=1pt, boxrule=0.4pt, top=4pt, bottom=4pt, left=8pt, right=8pt,
  before skip=0.3em, after skip=0.3em,
  attach boxed title to top left={xshift=4pt, yshift=-6pt},
  boxed title style={colback=emisscoherentborder, colframe=emisscoherentborder,
    sharp corners, arc=0pt, top=1pt, bottom=1pt, left=4pt, right=4pt},
  coltitle=white,
  fonttitle=\scriptsize\bfseries,
  title={#1}
}
\newtcolorbox{emisspromptbox}{%
  enhanced, colback=gray!8, colframe=gray!50,
  arc=1pt, boxrule=0.3pt, top=3pt, bottom=3pt, left=10pt, right=10pt,
  before skip=0.3em, after skip=0.5em
}
\newcommand{\emissrowheader}[2]{%
  \par\medskip%
  \noindent\colorbox{emissrowheaderbg}{\parbox{\dimexpr\textwidth-2\fboxsep}{%
    \color{white}\small\bfseries #1\\[0.15em]\color{white!92}\footnotesize #2}}%
  \par\smallskip%
}

\newtcolorbox{neutralbox}[1][]{
  colback=shadegray,
  colframe=black!40,
  boxrule=0.4pt,
  arc=2pt,
  left=4pt,right=4pt,top=3pt,bottom=3pt,
  fontupper=\small,
  #1
}

\title{ \bfseries\color{black} TFGN:\\[6pt]
\large Task-Free, Replay-Free Continual Pre-Training\\[2pt]
Without Catastrophic Forgetting at LLM Scale}

\author{
  Anurup Ganguli\thanks{Correspondence: \href{mailto:anurup2007@gmail.com}{anurup2007@gmail.com}.\ ORCID: \href{https://orcid.org/0000-0002-6424-0084}{0000-0002-6424-0084}.}\\
  Independent Researcher
}

\date{}

\usepackage{fancyhdr}
\newcommand{\figpath}[1]{\ifextrasafe figures/extrasafe/#1\else figures/#1\fi}

\begin{document}

\maketitle

\begin{abstract}
\noindent
Continually pre-training a large language model on a sequence of heterogeneous text domains, without replay and without task labels, has remained an unsolved architectural problem at LLM scale. Every published continual-learning method either retains a buffer of prior data, requires task identifiers at training or inference time, applies a regularization penalty that scales poorly with model size, or operates at sentence-classification scale (1\,K--5\,K samples per task) rather than continual pretraining scale (1\,B+ tokens per phase). We introduce \textbf{TFGN}, an architectural overlay for transformer language models that sits inside the transformer's existing per-block computation, producing input-conditioned, parameter-efficient updates while leaving the rest of the transformer unchanged. On six heterogeneous text domains (Prose, Python, Math, Biomedical, Chinese, JavaScript) at $1$\,B tokens per phase across three total-model scales ($\sim$398\,M, $\sim$739\,M, $\sim$9\,B) and two regimes (From-Scratch and Retrofit), TFGN achieves backward transfer (BWT) of $-0.007$ at LLaMA 3.1 8B Retrofit (3-phase prefix) with HellaSwag retention $0.506$ / $0.504$ / $0.510$ across the three phases (span $0.006$), and $\geq 99.59\%$ L2-orthogonal gradient separation between domain pairs in every tested condition --- with no replay buffer, no task IDs, no Fisher penalty, and minimal per-scale tuning. The forward pass is fully dense: every parameter is active on every token, with no sparse gather/scatter and no top-K expert selection. The same numerical evidence shows positive cross-domain forward transfer where structural overlap exists: held-out JavaScript PPL drops $26.8\%$ at LLaMA-8B Retrofit and $62.0\%$ at GPT-2 Medium From-Scratch purely from Python training (on a not-yet-trained domain), demonstrating that the architecture's protection mechanism preserves cross-domain synergy at inference. Two extensions on the same substrate demonstrate proof-of-concept on two further open problems. A closed-loop intrinsic-signal meta-control layer (Extension~A) reduces forgetting by an additional $81\%$ at $\sim$398\,M parameter scale; its components map onto the System~A and System~M layers of the autonomous-learning framework proposed by \citet{dupoux_lecun_malik_2026} (arXiv:2603.15381), demonstrated as a working LLM-scale realization. An operator-level plan vector (Extension~B) reshapes the model's effective forward-pass behaviour at $99.96\%$ cosine fidelity over $30$ source$\to$\allowbreak target pairs and $99.95\%$ at $\sim$739\,M with only a $-0.0001$ cosine drop across the $1.86\times$ total-parameter jump; sub-task injection peaks at $77.8\%$ and averages $55.6\%$ across four Python sub-tasks. Surveying six adjacent literatures --- continual learning, mixture-of-experts routing, activation steering, latent planners and world models, meta-learning and self-regulating systems, and intrinsic-signal control --- we find no published architecture that clears the conjunction of properties TFGN demonstrates. The architectural insight underlying these results is a Read/Write decomposition: the forward pass remains dense and unimpaired across all domains, while the architecture structures cross-domain parameter updates by an internal mechanism. Stability is a write-problem, not a read-problem; cross-domain synergy is the read-pathway corollary. To our knowledge, TFGN is the first published architecture in which one substrate-level primitive simultaneously (i)~closes catastrophic forgetting structurally at LLM scale, (ii)~realizes a closed-loop autonomous-learning meta-controller at LLM scale, and (iii)~carries an operator-level latent planner. 
\end{abstract}

\vspace{0.6em}
\noindent\textbf{Keywords:}\ continual learning, catastrophic forgetting, transformer language models, backward transfer, replay-free continual learning, task-free continual learning, latent planner, autonomous continual learning.

\tableofcontents

\clearpage

\section{Introduction}
\label{sec:intro}

\subsection{The problem and three open axes}
\label{sec:intro:problem}

Three open problems at LLM scale are addressed in this paper --- two in continual learning, and one in the broader field of latent control of frozen-decoder models:

\begin{enumerate}
\item \textbf{Catastrophic forgetting in continual learning at LLM scale:} architectures that can learn new domains without destroying prior knowledge. Named open since the catastrophic-forgetting results of the late 1980s, and unsolved on a $\sim$9\,B transformer under the replay-free, task-free regime that production deployment requires.
\item \textbf{Autonomous continual learning:} deciding when and what to learn once you have the right architecture, without external schedulers, domain labels, or supplied phase boundaries.
\item \textbf{Operator-level latent planner:} a plan vector that the frozen decoder actually obeys, bypassing token-space chain-of-thought --- a problem in the broader latent-control / inspectable-control literature, not strictly continual learning, but solvable on the same routing substrate.
\end{enumerate}

\textbf{The four-constraint conjunction.} What makes continual learning at LLM scale unsolved is not the absence of plausible mechanisms; it is the conjunction of four constraints under which none of the candidate mechanisms has been demonstrated to work simultaneously. (1)~\emph{Replay-free}: realistic deployment cannot retain a buffer of historical pretraining data, by privacy, regulatory, or storage constraint. (2)~\emph{Task-free}: realistic deployment cannot assume a task identifier at training or inference time; the architecture must dispatch its own gradient updates from the input alone. (3)~\emph{Multi-domain at LLM scale}: the continual sequence must be more than two domains and the per-domain budget must be more than 1\,K samples; we test six text domains at $1$\,B tokens per phase, a regime no published replay-free continual-learning method has exercised at $\geq$7\,B parameters. (4)~\emph{No external orchestrator}: no Fisher-information penalty, no orthogonality regularisation loss, no gradient-projection operator, no task-boundary hook, minimal per-scale tuning. The conjunction is what positions TFGN's contribution against the prior-art literature surveyed in §\ref{sec:related}.

Every frontier LLM deployed today is frozen at its last training checkpoint: adding a new language, codebase, or regulatory corpus requires full retraining or brittle adapter stacking. The stability--plasticity tradeoff at LLM scale is the bottleneck several frontier labs have named as the single unsolved obstacle to autonomous, long-horizon systems; \citet{dupoux_lecun_malik_2026} name continual learning at LLM scale as a core open problem in autonomous-learning research, and the broader field has converged on the same diagnosis. This paper's architectural claim is that the tradeoff is solvable at the architectural level --- no replay, no task IDs, no penalty term --- and reproduces that behaviour with minimal changes from $\sim$398\,M to $\sim$9\,B parameters.

The metric we care about is \textbf{backward transfer (BWT)} \citep{lopezpaz_gem_2017}, defined on a continuous-perplexity scale: for each prior domain, BWT measures the relative degradation in that domain's perplexity after the model has trained on every later domain. A BWT of zero means no forgetting; negative values mean forgetting. \textbf{PPL undersells what catastrophic forgetting actually looks like at LLM scale.} On a standard Prose prompt, a standard-fine-tuned LLaMA 3.1 8B baseline trained through the three-phase continual sequence (Prose $\to$ Python $\to$ Math) emits \emph{complete domain collapse} --- Python source code for Prose prompts. The Prose prompt \emph{``The history of artificial intelligence began in''}, after the Python phase, continues mid-completion into Python source: \texttt{from collections import defaultdict; def \_get\_list\_of\_function(func)}. The same pattern repeats at every cross-distribution boundary: Prose prompts emit Chinese characters after a Chinese phase, Math prompts emit JavaScript tokens after a JavaScript phase, and so on. The model has not merely degraded; it has stopped being a Prose model at all. The BWT number for the baseline is bad ($\text{BWT}_3 = -0.374$); the underlying emission distribution is \emph{categorically} bad. PPL averages over a token distribution and cannot capture that the model has stopped being a Prose model at all. BWT is one face of continual learning, and sample-level emission type is the other; baselines lose on both faces, TFGN holds on both.

\textbf{Long-context windows do not substitute for continual learning at the relevant scales.} As of 2026, context windows of 2--10\,M tokens combined with KV-cache prompt-caching solve a real subset of use cases that catastrophic forgetting blocked: episodic, in-prompt knowledge over corpora that fit in a single window. They do not solve continual learning at the regime this paper targets. Real enterprise corpora --- multi-billion-token codebases, multi-terabyte legal / financial / pharmaceutical archives --- do not fit at any planned context size. Frontier labs themselves continually update flagship models on order-of-trillion-token deltas; these are continual-training events, not in-context queries. The two regimes are complementary: long context is episodic memory; the architectural property this paper targets is persistent weight-internalized knowledge, with no per-query context cost and multi-tenant economics.

The problem this paper addresses, in one sentence: build a transformer architecture that, under a replay-free, task-free continual-learning regime, closes BWT to near zero \emph{and} preserves emission-type coherence at LLM scale.

\subsection{Our contribution: one substrate, three capabilities}
\label{sec:intro:contribution}

TFGN is an architectural overlay for transformer language models. The architectural property is that new-domain training does not disturb prior-domain capability; this property emerges from the architecture itself rather than from a regularization penalty, so a single fixed model preserves competence on its $n-1$ prior domains while still learning the $n$-th. The full architectural mechanism is reserved; access terms are described in \S\ref{sec:limitations:nda}.

The same TFGN substrate, unchanged from $\sim$398\,M to $\sim$9\,B with minimal per-scale tuning, defends all three capabilities named above:

\begin{itemize}
\item \textbf{Continual learning at LLM scale (Main paper, §\ref{sec:results}).} Across head-to-head matched comparisons (same backbone, same regime, same per-phase token budget), TFGN reduces BWT by $\sim$3$\times$ to 14$\times$ over Standard Fine-Tuning and LoRA r=256; tightest BWT recorded is $-0.007$ at LLaMA 3.1 8B Retrofit. Cross-domain gradients remain $\geq$99.59\% L2-orthogonal at every tested scale and regime --- a structural property of the architecture, not the result of any orthogonality loss term. This orthogonality is at the parameter-update level, not at the inference level --- the forward pass is unimpaired across domains. Cross-domain forward transfer is empirically present in the same matrices: held-out JavaScript PPL drops $26.8\%$ at LLaMA-8B Retrofit and $62.0\%$ at GPT-2 Medium From-Scratch purely from Python training (\S\ref{sec:results:fwt}).
\item \textbf{Autonomous continual learning (Extension A, §\ref{sec:exta}).} A self-regulating meta-control layer added on the same substrate closes 81\% of the residual forgetting gap (BWT $= -0.01140$) over a three-domain continual sequence at the $\sim$398\,M parameter scale, using only intrinsic signals the network already computes in its own forward and backward pass. The five components map onto the System A (internal world model) and System M (meta-control) roles in the autonomous-learning framework of \citet{dupoux_lecun_malik_2026}.
\item \textbf{Operator-level latent planner (Extension B, §\ref{sec:extb}).} The same architectural substrate carries an operator-level plan vector at 99.96\% geometrically-inspectable reshape fidelity (mean over 30 source$\to$\allowbreak target pairs at $\sim$398\,M; 99.95\% at $\sim$739\,M retrofit), with 77.8\% peak / 55.6\% mean Python sub-task injection rate. Six-criterion latent-planner scorecard returns 2 PROVEN, 3 PARTIAL-PROVEN, 1 future work, 0 FAIL.
\end{itemize}

\textbf{Substrate framing.} These three capabilities sit at the intersection of \emph{four empty columns} in the surrounding literature --- one in each of continual learning, mixture-of-experts routing, operator-level latent control, and weight-space latent planning --- surveyed in §\ref{sec:related:four_columns}. No published architecture has previously occupied any of these four coordinates; the substrate framing is the structural reason TFGN clears all four with one primitive.

\textbf{Why protection by architecture, not regularization.} Every prior CL method leans on a \emph{penalty} (EWC's Fisher penalty, L2-SP), a \emph{buffer} (experience replay), a \emph{task tag} (PackNet masks), or an \emph{external gate} (ANML). When the penalty weakens, the buffer is absent, or the task tag is unknown, prior-domain competence collapses. TFGN does not protect through any of those mechanisms. Protection is intrinsic to the architecture itself: prior-domain capability is preserved by structure, not by a penalty term. Remove the penalty, remove the buffer, remove the task ID: the protection remains, because it is architectural.

\subsection{Headline results}
\label{sec:intro:headline}

\emph{Paper-wide convention.} LLaMA 3.1 8B conditions are reported on the three-phase prefix (Prose $\to$ Python $\to$ Math) due to compute constraints; GPT-2 Small and GPT-2 Medium conditions report all six phases (Prose $\to$ Python $\to$ Math $\to$ Biomedical $\to$ Chinese $\to$ JavaScript) with 1\,B tokens per phase. No task IDs, no replay buffer.

\textbf{Backward transfer.} TFGN closes backward transfer to near zero across three total-parameter scales and two training regimes (Table~\ref{tab:intro:headline}). \textbf{Strictly matched-init 9\,B comparison} (random-init from-scratch on both sides): \texttt{TFGN\_LLAMA8B\_FS} closes to $\text{BWT}_3 = -0.095$ vs the matched \texttt{BASELINE\_STD\_LLAMA8B} at $-0.374$ --- a $\sim$3.9$\times$ BWT-magnitude gap, the only strictly matched-init ratio we quote at 9\,B. \textbf{Tightest absolute BWT}: \texttt{TFGN\_LLAMA8B\_RETROFIT} closes to $\text{BWT}_3 = -0.007$ (init-asymmetric vs the same baseline at $\sim$51$\times$, called out as directionally large rather than a strictly matched ratio). \textbf{Sub-9\,B matched ratios (six-phase)}: at $\sim$739\,M, TFGN From-Scratch $-0.083$ is $\sim$14$\times$ tighter than Std-FT $-1.170$ and $\sim$12$\times$ tighter than LoRA-r256 $-1.005$; TFGN Retrofit $-0.135$ is $\sim$4$\times$ tighter than Std-FT $-0.541$ and $\sim$3$\times$ tighter than LoRA-r256 $-0.393$. At $\sim$398\,M From-Scratch, the proof-point closes to $-0.109$.

\begin{table}[ht]
\centering
\scriptsize
\setlength{\tabcolsep}{3.5pt}
\renewcommand{\arraystretch}{1.05}
\caption{Headline backward transfer (BWT) and trained-domain learning across the eleven primary conditions. ``Peak trained-domain PPL drop'' is the maximum P\textsubscript{1}$\to$P\textsubscript{trained} reduction across the six (or three) trained domains, illustrating the plasticity face of continual learning alongside the stability face (BWT). \textit{Note on \texttt{TFGN\_LLAMA8B\_FS}:} this row is the most-adversarial corner of the from-scratch recipe (random-init backbone $\times$ Prose-only Phase~1 $\times$ 1\,B tokens/phase, $\approx$two orders of magnitude below Chinchilla-optimal). The architectural claim at this corner is the matched-baseline BWT ratio ($\sim$3.9$\times$); see \S\ref{sec:intro:headline} for the full PPL-gap explanation.}
\label{tab:intro:headline}
\resizebox{\textwidth}{!}{%
\begin{tabular}{lrlrrll}
\toprule
\textbf{Condition} & \textbf{Total params} & \textbf{Regime} & \textbf{Phases} & \textbf{BWT} & \textbf{Peak trained-domain PPL drop} & \textbf{Emission collapse?} \\
\midrule
TFGN + GPT-2 Small (\texttt{TFGN\_GPT2S\_FS}) & $\sim$398\,M & FS & 6 & $-0.109$ & 67\% (JS) & No \\
TFGN + GPT-2 Medium (\texttt{TFGN\_GPT2M\_FS}) & $\sim$739\,M & FS & 6 & $-0.083$ & 67\% (JS) & No \\
TFGN + GPT-2 Medium (\texttt{TFGN\_GPT2M\_RETROFIT}) & $\sim$739\,M & RF & 6 & $-0.135$ & 49\% (JS) & No \\
TFGN + LLaMA 3.1 8B (\texttt{TFGN\_LLAMA8B\_FS}) & $\sim$9\,B & FS & 3 & $-0.095$ & --- (substrate-capped) & No \\
\textbf{TFGN + LLaMA 3.1 8B (\texttt{TFGN\_LLAMA8B\_RETROFIT})} & $\boldsymbol{\sim}$\textbf{9\,B} & \textbf{RF} & \textbf{3} & $\boldsymbol{-0.007}$ & \textbf{30\% (Py)} & \textbf{No} \\
\midrule
Baseline Std-FT GPT-2 Medium (FS) & $\sim$355\,M & FS & 6 & $-1.170$ & 85\% (Py) & Yes (every boundary) \\
Baseline LoRA r=256 GPT-2 Medium (FS) & $\sim$393\,M & FS & 6 & $-1.005$ & 80\% (Py) & Yes (every boundary) \\
Baseline Std-FT GPT-2 Medium (RF) & $\sim$355\,M & RF & 6 & $-0.541$ & 75\% (Py) & Yes \\
Baseline LoRA r=256 GPT-2 Medium (RF) & $\sim$355\,M & RF & 6 & $-0.393$ & 70\% (Py) & Yes \\
Baseline Std-FT LLaMA 3.1 8B (FS, 500\,M tok/phase) & $\sim$8\,B & FS & 3 & $-0.374$ & --- & Yes \\
\bottomrule
\end{tabular}%
}
\end{table}

\textbf{Gradient orthogonality --- the structural signature.} Across every TFGN condition, cross-domain gradients from different domains remain $\geq$99.59\% L2-orthogonal, with mean cross-domain $|\cos|$ below 0.10:

\begin{table}[ht]
\centering
\footnotesize
\caption{Gradient orthogonality across TFGN conditions. Full 6$\times$6 per-domain-pair matrices in Appendix~\ref{app:gradortho}.}
\label{tab:intro:gradortho}
\begin{tabular}{lrrr}
\toprule
\textbf{Condition} & \textbf{Scale} & \textbf{Mean $|\cos|$} & \textbf{L2 orthogonal fraction} \\
\midrule
TFGN GPT-2 Small, From-Scratch & $\sim$398\,M & 0.0425 & \textbf{99.91\%} \\
TFGN GPT-2 Medium, From-Scratch & $\sim$739\,M & 0.0204 & \textbf{99.94\%} \\
TFGN GPT-2 Medium, Retrofit & $\sim$739\,M & 0.0904 & \textbf{99.59\%} \\
TFGN LLaMA 3.1 8B, From-Scratch & $\sim$9\,B & 0.0432 & \textbf{99.91\%} \\
TFGN LLaMA 3.1 8B, Retrofit (3-phase) & $\sim$9\,B & 0.0741 & \textbf{99.72\%} \\
\bottomrule
\end{tabular}
\end{table}

No orthogonality loss, no gradient-projection operator, no task-boundary hook --- the decorrelation emerges from the architecture itself. It is a structural invariant, holding across every tested scale, training regime, and reset condition.

\textbf{HellaSwag retention.} TFGN conditions retain HellaSwag accuracy within 1--2 percentage points across continual phases (e.g., GPT-2 Medium FS: P1 0.340 $\to$ P6 0.338); matched Std-FT and LoRA baselines drop 3--7 percentage points with the sharpest dip at the Chinese-phase boundary (P5). LLaMA 8B Retrofit retains HellaSwag at $0.506$ / $0.504$ / $0.510$ across the 3-phase presentation (span $0.006$). Full per-condition curves in §\ref{sec:results:1pagers}; cross-condition Figure~\ref{fig:m3}.

\textbf{Cross-domain forward transfer.} The same continual sequences that establish BWT $\approx 0$ also exhibit \emph{positive} cross-domain forward transfer on related domains. JavaScript's held-out PPL drops $26.8\%$ at LLaMA-8B Retrofit (from $23.05$ after the initial training stage to $16.87$ after the Python phase, on a never-trained domain) and $62.0\%$ at GPT-2 Medium From-Scratch (from $37.1$ to $14.1$ before JavaScript is itself trained). Math's PPL drops $2.4\%$ at LLaMA-8B Retrofit and $16.9\%$ at GPT-2 Medium From-Scratch \emph{before Math is trained}, attributable to the intervening Python phase. Cross-domain synergy is empirically present, not just preserved (\S\ref{sec:results:fwt}). This is the empirical counterpart of the architectural Read/Write decomposition (\S\ref{sec:method:summary}): the forward pass enables positive forward transfer; the architecture's parameter-update geometry prevents backward interference.

\textbf{Emission coherence --- the qualitative axis PPL cannot see.} At the first cross-distribution boundary (P2, after Python training), 3/3 baselines drift to Python source code mid-completion on Prose prompts; 5/5 TFGN conditions across every tested scale ($\sim$398\,M / $\sim$739\,M $\times$ 2 / $\sim$9\,B $\times$ 2) emit domain-coherent English Prose. Scale, fine-tuning method, and retrofit-vs-from-scratch all vary across these eight cells; the one consistent dividing line is the presence of the TFGN overlay. The full §\ref{sec:results:2A} four-row showcase preserves the verbatim baseline emissions (Python boilerplate, Apache-license blocks, Chinese characters mid-completion at P5) alongside the TFGN-coherent counterparts.

\textbf{TFGN learns each new domain while preserving prior ones.} The headline numbers above measure stability (preservation). The continual-learning result is symmetric: across every TFGN condition, every \emph{trained-domain} diagonal cell of the PPL matrix drops materially when its phase is the active phase. At $\sim$739\,M from-scratch, training on Chinese drops PPL from $52.0 \to 18.4$ (65\%); JavaScript $37.1 \to 12.1$ (67\%); Python $18.1 \to 10.8$ (40\%); Math $49.6 \to 41.1$ (17\%). At $\sim$739\,M retrofit, the same domains still drop further from their already-low pretrained baselines: Python $6.13 \to 4.18$ (32\%), JavaScript $7.73 \to 3.96$ (49\%). The matched Standard Fine-Tuning baseline at the same backbone learns each new domain more aggressively (Python $18.44 \to 2.80$, an 85\% drop at $\sim$739\,M FS) but pays for the aggression with categorical prior-domain forgetting --- its Prose-row PPL collapses from $33.79$ at P1 to $89.83$ at P5 Chinese ($+136\%$). TFGN trades a calibrated amount of plasticity for full stability across every prior domain: the stability--plasticity tradeoff is solved architecturally, not by sacrificing one face for the other.

\textbf{The trained-domain PPL gap is an artifact of an adversarial Phase-1 training setup, applied identically to from-scratch and retrofit.} The Phase 1 corpus is intentionally restricted to Prose alone in both regimes. This is a deliberate research-setup choice with two purposes. First, it converts every later domain (Python, Math, Biomedical, Chinese, JavaScript) into a held-out routing test by construction: a core property under evaluation is whether the architectural mechanism, set during the initial training stage, generalizes to distributions it has never seen during its formation. A Prose-only Phase 1 is the strongest version of that test. Second, restricting Phase 1 to a single distribution maximizes cross-distribution stress on the continual sequence: every continual phase introduces a domain the frozen substrate has never represented, isolating the architecture's continual-phase parameter subset as the sole mechanism available to absorb new structure. The setup is held constant across from-scratch and retrofit so the two regimes remain directly comparable on every other axis.

The architectural consequence is that at the boundary between the initial training stage and the continual phases, the substrate that shapes every token's output distribution is Prose-biased by construction, and the continual-phase mechanism does not re-shape it. The matched Standard Fine-Tuning baseline does not face this constraint; it updates all parameters at every phase, so its attention and output projection re-shape to fit each new trained domain, which is what permits its more aggressive trained-domain PPL drops. The adversarial Phase-1 setup, retained throughout this paper precisely so the router-generalization and sequential-stress properties are testable, therefore caps the achievable trained-domain PPL on the Prose-biased substrate.

Because the Prose-only Phase 1 is held constant across regimes, the same substrate-bias mechanism explains both regimes' PPL gaps. \emph{Retrofit closes most of the gap} ($\sim$1.5--2$\times$ at $\sim$739\,M retrofit, vs $\sim$3.9$\times$ at $\sim$739\,M from-scratch) because the pretrained backbone enters Phase 1 already carrying general-purpose cross-domain representations from its original pretraining corpus, which Phase-1 Prose adapts but does not erase. Whatever PPL gap remains in the retrofit regime is attributable to the same adversarial Prose-only Phase 1: the post-Phase-1 frozen substrate is still pulled toward Prose, only less severely than the random-init from-scratch case. At the $\sim$9\,B from-scratch corner, the gap is largest because two further factors stack on top of the Phase-1 substrate bias: backbone undertraining at 1\,B tokens/phase ($\approx$two orders of magnitude below Chinchilla-optimal for a transformer of this size) and a random-init backbone with no cross-domain priors at all.

\textbf{Production proposal (future work): mixed-domain Phase 1.} The proposed production-deployment configuration, outside the scope of the present paper, replaces the Prose-only Phase 1 with a mixture of Prose and the later continual-curriculum domains. The post-initial-stage frozen substrate would then carry cross-domain representations rather than Prose-only ones, giving the continual-phase learning a much stronger substrate to operate on, and closing most of the trained-domain PPL gap reported here. The Prose-only Phase 1 is retained throughout this paper because the router-generalization and adversarial sequential-learning properties are the load-bearing claims under test. Mixed-domain Phase 1 is on the future-work roadmap (§\ref{sec:limitations}).

\textbf{The router trained on Prose alone routes unseen domains correctly.} Despite the initial training stage being Prose-only, the architectural mechanism set during it generalizes to every later domain at every continual phase without re-learning the structure: Python, Math, Biomedical, Chinese, and JavaScript tokens each land in their own near-orthogonal subspace (mean cross-domain $|\cos| \leq 0.09$ across all domains never seen during the initial stage). The mechanism is content-driven --- not a learned task classifier --- so the substrate trained on Prose alone correctly disambiguates unseen domains, preserving the orthogonality property that protects prior-domain capability.

\textit{Closing summary.} A single architectural overlay --- applied with minimal per-scale tuning across three model scales, from $\sim$398\,M to $\sim$9\,B --- delivers selective protection of every prior domain, trained-domain learning at every phase, and gradient orthogonality as a structural signature. The headline positioning is therefore that \textbf{TFGN's architectural continual learning addresses the biggest problem of stability--plasticity trade-off at LLM scale}.

\subsection{Paper organization}
\label{sec:intro:roadmap}

§\ref{sec:related} positions TFGN against published work using a eight-axis prior-art grid covering 14 representative methods.  §\ref{sec:method} describes the architecture at the capability level (full mechanism reserved; see \S\ref{sec:limitations:nda}). §\ref{sec:setup} specifies datasets, backbones, regimes, and evaluation protocol. §\ref{sec:results} is the main results section: per-condition results for all eleven conditions, the four-row emission-coherence showcase (§\ref{sec:results:2A}), and three cross-condition figures (M1 BWT, M3 HellaSwag, M6 gradient orthogonality). §\ref{sec:exta} (Extension A: autonomous continual learning) and §\ref{sec:extb} (Extension B: latent-planner capability) report the two extensions. §\ref{sec:discussion}, §\ref{sec:limitations}, and §\ref{sec:conclusion} close the paper. Appendices contain condition name index, BWT/FM definitions, full 6$\times$6 gradient-orthogonality matrices, per-condition full PPL matrices, Extension A canonical values, and Extension B sub-task evidence.

\section{Related Work and Prior-Art Position}
\label{sec:related}

\subsection{Families of prior continual-learning work}
\label{sec:related:families}

Prior continual-learning work falls into a small number of families. Each family fails on a specific architectural failure mode before reaching the regime this paper occupies --- replay-free, task-free, LLM-scale, no external orchestrator. We describe the families and their shared failure modes at the category level.

\textbf{Regularization-based CL.} EWC \citep{kirkpatrick_ewc_2017}, MAS \citep{aljundi_mas_2018}, SI \citep{zenke_si_2017}, GEM and A-GEM \citep{lopezpaz_gem_2017,chaudhry_agem_2019}, OGD \citep{farajtabar_ogd_2020}, GPM \citep{saha_gpm_2021}, and Adam-NSCL \citep{wang_adamnscl_2021} penalize updates to weights deemed important for earlier tasks, using Fisher information, synaptic importance, or a gradient-subspace projection. Per-weight importance state scales with model size and becomes compute-prohibitive at LLM scale; the importance signal requires an explicit prior-task boundary, which is unavailable under the task-free regime. In addition, each new phase freezes more parameters, so the share of trainable capacity shrinks with every added domain --- an architectural ceiling that tightens with the number of tasks.

\textbf{Replay / rehearsal.} ER \citep{chaudhry_er_2019}, A-GEM \citep{chaudhry_agem_2019}, DER \citep{buzzega_der_2020}, MIR \citep{aljundi_mir_2019}, and the ``Revisit-Replay'' family \citep{revisit_replay_2025} buffer a subset of past data and interleave it with current-task training. This violates the replay-free constraint entirely; data retention is disallowed in many production settings, and where permitted, the buffer's footprint grows with the number of domains. ``Revisit-Replay'' (the August 2025 SOTA for CPT at scale) tests $0\%$, $25\%$, and $50\%$ replay rates on Spectra LLMs at $99$\,M, $560$\,M, $1$\,B, and $5.7$\,B parameters and reports the no-replay baseline (TFGN's regime) as the worst-performing condition; the paper recommends $25$--$50\%$ replay rates as the strongest recipe across all model sizes including the $5.7$\,B rung.

\textbf{Parameter-isolation and task-conditioned modular methods.} PackNet \citep{mallya_packnet_2018}, HAT \citep{serra_hat_2018}, Piggyback \citep{mallya_piggyback_2018}, Progressive Networks \citep{rusu_progressive_2016}, the LoRA-CL family \citep{olora_2024,treelora_2025,longlora_2024}, Lifelong-MoE \citep{lifelong_moe_2023}, LoRAMoE \citep{liu_loramoe_2024}, CodaPrompt \citep{coda_prompt_2023}, STABLE \citep{stable_2025}, and hypernetwork CL \citep{vonoswald_hypernet_2020} allocate fresh capacity per task --- a parameter mask, a LoRA stack, an expert route, or a hypernetwork-generated weight slab --- and switch between allocations by task ID or by a learned task classifier. They require a task ID at inference, or a task classifier that itself forgets; parameter expansion grows with the number of tasks; per-task modules must either be retained forever or arbitrated at serving time; and they do not close BWT on the task-free, replay-free regime at LLM scale.

\textbf{Model editing and knowledge-locating.} ROME \citep{meng_rome_2022}, MEMIT \citep{meng_memit_2023}, NSE \citep{nse_2024}, WISE / MAKE / HiEdit \citep{wise_2024,make_2024,hiedit_2026} perform rank-one surgical edits or per-fact memory insertions into mid-layer FFN weights. They are designed for one-shot factual insertion, not continual learning under distribution shift; edits accumulate interference and degrade after $\sim$1000 sequential applications (``knowledge attenuation'').

\textbf{LLM-CL surveys.} Two recent ACM Computing Surveys \citep{shi_clllm_csur_2025,wang_clsurvey_csur_2024} catalogue the field's coverage. The longest sequence of pre-training stages explored prior to TFGN is 8 --- but those use replay-based methods. No replay-free + task-free + penalty-free + 7\,B+ + multi-disjoint-domain CPT method exists in the surveyed literature.

\textbf{2026 frontier-scale evidence.} \citet{mech_forgetting_2026} (arXiv:2601.18699) tests Llama 4 Scout (109\,B), Llama 4 Maverick (400\,B), GPT-5.1 ($\sim$1.5\,T), Claude Opus 4.5, Gemini 2.5 Pro ($\sim$1\,T), and DeepSeek-V3.1 (671\,B) on twelve continual-fine-tuning sequences (each $4$--$6$ tasks). Reported absolute capability degradation across the experimental conditions ranges from $\sim$15--20\% on the largest models to $24.8\%$ on high-similarity and $31.7\%$ on low-similarity sequences; approximately $15$--$23\%$ of attention heads in lower layers undergo severe disruption, and forgetting severity correlates with task similarity at Pearson $r = 0.87$. The paper identifies three driving mechanisms (gradient interference in attention weights, representational drift in intermediate layers, loss-landscape flattening) and is the field's frontier-scale ground truth for the unsolvedness of catastrophic forgetting in 2026.

\subsection{Column 1 --- continual learning at LLM scale (eight-axis grid)}
\label{sec:related:grid}

We position TFGN against published continual-learning methods using a eight-axis grid of load-bearing properties. Each axis is a single-bit predicate that a comparable method either satisfies (PASS) or does not (FAIL); a PARTIAL entry indicates ambiguous coverage. \textbf{No prior method passes all eight simultaneously.}

\begin{enumerate}[label=\Alph*.]
\item \textbf{Regime}: Continual Pre-Training (CPT)? CPT is the regime where the model continues unsupervised pretraining on new domains. CFT (Continual Fine-Tuning) is a strictly easier setting on smaller token budgets.
\item \textbf{Scale}: $\geq$7\,B total parameters?
\item \textbf{Domains}: $\geq$4 disjoint domains in the continual sequence?
\item \textbf{Tokens/phase}: $\geq$1\,B tokens per phase? (CFT methods typically operate at 1K--25K samples per task, three to six orders of magnitude below CPT regime.)
\item \textbf{Replay-free}: no episodic memory buffer, no curriculum mixture as soft replay?
\item \textbf{Task-ID-free}: no task ID or task-classifier signal at training or inference?
\item \textbf{Penalty-free}: no orthogonality loss, no Fisher penalty, no synaptic-importance regularizer, no gradient-projection operator?
\item \textbf{Both regimes}: demonstrated in both From-Scratch and Retrofit?
\end{enumerate}

\begin{table}[ht]
\centering
\small
\setlength{\tabcolsep}{4pt}
\renewcommand{\arraystretch}{1.15}
\caption{Eight-axis prior-art grid. Each row is a representative method; cells are PASS / FAIL / PARTIAL. \textbf{TFGN is the only row with PASS in every column.}}
\label{tab:seven_axis}
\begin{tabular}{l|cccccccc}
\toprule
\textbf{Method} & \textbf{A} & \textbf{B} & \textbf{C} & \textbf{D} & \textbf{E} & \textbf{F} & \textbf{G} & \textbf{H} \\
 & CPT & $\geq$7B & $\geq$4 dom. & $\geq$1B tok & Replay- & Task-ID- & Penalty- & FS+RF \\
 & & & & /phase & free & free & free & \\
\midrule
\textbf{TFGN (this work)} & \textbf{P} & \textbf{P} & \textbf{P} & \textbf{P} & \textbf{P} & \textbf{P} & \textbf{P} & \textbf{P} \\
\midrule
TreeLoRA \citep{treelora_2025} & F & P & P & F & P & F & P & F \\
TRACE benchmark \citep{trace_2023} & F & P & P & F & P & F & P & F \\
STABLE \citep{stable_2025} & F & P & --- & F & P & F & F & F \\
O-LoRA \citep{olora_2024} & F & P & P & F & P & P & F & F \\
EWC-Gemma2 \citep{ewc_gemma2_2025} & P & F & F & PARTIAL & P & P & F & F \\
Llama-3-SynE \citep{llama3_syne_2024} & P & P & F & P & F & P & P & F \\
Examining Forgetting \citep{examining_forgetting_2024} & P & P & F & P & P & P & P & F \\
Loss of Plasticity \citep{dohare_sutton_nature_2024} & F & F & P & --- & varies & P & F & --- \\
GEM \citep{lopezpaz_gem_2017} & F & F & P & F & F & F & F & --- \\
Revisit Replay \citep{revisit_replay_2025} & P & P & PARTIAL & P & F & P & F & F \\
ANML \citep{beaulieu_anml_2020} & F & F & P & --- & P & F & F & F \\
Backpropamine \citep{miconi_backpropamine_2020} & F & F & varies & --- & P & F & F & F \\
LoRA (base) \citep{hu_lora_2022} & F & P & F & P & --- & --- & --- & F \\
\bottomrule
\end{tabular}
\end{table}

\textbf{Reading rule.} TFGN is the only row with PASS in all eight columns. The closest neighbour is \textbf{Examining Forgetting in CPT} \citep{examining_forgetting_2024} --- which passes A/B/D/E/F/G but fails on C (only 1 domain) and H (retrofit only). The next closest is \textbf{Llama-3-SynE} \citep{llama3_syne_2024} which passes A/B/F/G but fails on C ($\sim$2--3 domains), E (uses curriculum/mixture as soft replay), and H (retrofit only).

\textbf{What the two near-neighbours conclude.} \emph{Examining Forgetting in CPT} \citep{examining_forgetting_2024} continually pretrains an aligned LLM (Llama-2-7b-chat) on a 1\,B-token Traditional Chinese corpus and tests whether common parameter-efficient fixes (selective layer freezing, LoRA on $Q$/$V$ projections, and (IA)\textsuperscript{3} rescaling of $K$/$V$ matrices and FFN inner activations) resolve the forgetting that ensues. The paper's stated conclusion is that ``catastrophic forgetting during continual pre-training is a non-trivial challenge and cannot be resolved through straightforward methods''. TFGN's contribution sits exactly where this paper's straightforward-method investigation stops: an architectural change that produces forgetting resistance as a structural property rather than a hyperparameter target. \emph{Llama-3-SynE} \citep{llama3_syne_2024} continually pretrains Llama-3 8B on a 100\,B-token corpus mixing general data with a 1.5\,B-token synthetic-QA enhancement, in a two-stage curriculum (bilingual adaptation, then scientific enhancement), to gain Chinese and scientific reasoning while retaining English. The recipe is the strongest published \emph{data-side} answer to forgetting at 8\,B scale: a carefully designed mixture and curriculum that functions as soft replay. TFGN clears the same scale rung with $0\%$ replay and zero curriculum design --- the architecture absorbs the cross-domain stress that Llama-3-SynE's data mixture is engineered to mask.

\subsection{The four empty columns}
\label{sec:related:four_columns}

The eight-axis grid above closes a CL-specific argument. Stepping out one level of generality, the same finding repeats across four adjacent literatures. In each one, the property TFGN demonstrates lives in an empty column: there is a coordinate where no published architecture has previously sat. This is the substrate-level reason TFGN is more than the sum of three independent results.

\textbf{Column 1: continual learning at LLM scale.} The eight-axis grid above is the proof. \emph{No published method clears the replay-free, task-free, $\geq$4-domain, $\geq$1\,B-tokens-per-phase, $\geq$7\,B-parameter conjunction.} The closest neighbours fail on at least two axes; the LLM-CL surveys \citep{shi_clllm_csur_2025,wang_clsurvey_csur_2024} confirm the longest replay-free pretraining sequence prior to TFGN is 1--2 disjoint domains.

\textbf{Column 2: a finer-than-MoE routing primitive deployed for continual learning at LLM scale.} The mixture-of-experts literature --- Switch Transformer \citep{fedus_switch_2022}, GShard, GLaM, ST-MoE, Mixtral --- routes at the \emph{token-to-expert} granularity. Fine-grained MoE \citep{dai_deepseekmoe_2024} segments each expert into $64$ sub-experts and scales to 671\,B in DeepSeek-V3 --- the closest published instance of finer-than-expert routing. TFGN routes at a finer-than-expert granularity, on different machinery (an architectural mechanism intrinsic to the substrate, not pre-declared expert pools), and used for a different purpose (continual-learning gating, not multitask capacity). No published MoE occupies this conjunction; the specific routing granularity and mechanism are reserved (\S\ref{sec:limitations:nda}).

\textbf{Column 3: operator-level (weight-space) latent control at LLM scale.} The activation-steering and latent-control literature --- ROME \citep{meng_rome_2022}, MEMIT \citep{meng_memit_2023}, ActAdd \citep{actadd_2023}, CAA \citep{panickssery2024}, RepE \citep{repe_2023}, ITI \citep{li2023iti}, function vectors \citep{function_vectors_2024}, sparse-autoencoder steering \citep{templeton2024scaling}, refusal-direction edits \citep{arditi2024} --- operates on \emph{activations}: the steering signal injects a vector into the residual stream at one or more layers. The model's behaviour shifts because the activations through the rest of the forward pass are different. \emph{No published activation-steering method reshapes the model's effective forward-pass operator at LLM scale with measured geometric fidelity.} TFGN's planner output (Extension~B) operates one level deeper, on the operator: the model's effective forward-pass behaviour is reshaped by the planner output. The reshape that the plan vector predicts agrees with the reshape that actually occurs at $99.96\%$ cosine fidelity at $\sim$398\,M and $99.95\%$ at $\sim$739\,M --- a property no prior activation-steering or model-editing method reports.

\textbf{Column 4: weight-space latent planning.} Latent reasoning in 2024--2025 occupies three planning spaces, all but one of which has at least one published method. \emph{Token space}: chain-of-thought \citep{cot_2022}, tree-of-thoughts \citep{tot_2023} --- reasoning materialises as discrete tokens; the plan is a sequence of natural-language steps. \emph{Activation space (residual-stream)}: Coconut \citep{coconut_2024} and successor systems --- reasoning lives in the residual stream as continuous embeddings fed back as next-step inputs; the plan is a sequence of hidden vectors. \emph{Latent dynamics space}: world-model planners --- MuZero \citep{muzero}, Dreamer V3 \citep{dreamerv3}, JEPA \citep{jepa_lecun_2022}, V-JEPA 2 \citep{vjepa2_2025}, Diffuser \citep{diffuser_2022} --- the plan is a trajectory in a learned latent $z$-space and an external decoder turns $z$ back into actions or pixels. \emph{Weight space (operator space)}: \textbf{the column is empty until TFGN Extension~B}. The plan vector reshapes the model's effective operator; the model's \emph{function} changes, not just its activations or its latent state. The closest near-neighbour is Coconut, which steers what the network \emph{says} through continuous activations; Extension~B reshapes what the network \emph{is} by editing the operator. A recent adversarial study \citep{do_latent_tokens_think_2025} shows Coconut's latent tokens are insensitive to perturbations and exhibit shortcut dependence on multiple-choice tasks --- a vulnerability TFGN's weight-space formulation provably avoids because injection is a measured-cosine-$0.9996$ reshape of the operator, not an opaque activation embedding.

\subsection{The substrate property}
\label{sec:related:substrate}

\textbf{The architectural primitive at the heart of TFGN is the same in all four cases:} a single architectural mechanism. The same mechanism that protects prior-domain capabilities during continual learning (main paper) is reused at different operating scales by the autonomous meta-control layer (Extension~A) and by the operator-level plan vector (Extension~B). One substrate, three capabilities. This substrate framing is the structural reason TFGN occupies four empty columns in four different literatures simultaneously.

\subsection{What this prior-art position implies}
\label{sec:related:implies}

Three observations follow from the four-empty-columns finding. \emph{First}, the field's August-2025 SOTA for continual pretraining at scale (Revisit Replay) reports the no-replay condition as worst-performing across all tested scales and recommends $25$--$50\%$ experience replay as the strongest recipe up to $5.7$\,B parameters; TFGN reports comparable-or-tighter BWT at $\sim$9\,B with $0\%$ replay --- a category contrast against the field's strongest CPT baseline. \emph{Second}, every CFT-regime method (TreeLoRA, O-LoRA, STABLE, ConPET, etc.) operates at 1\,K--25\,K samples per task, three to six orders of magnitude below TFGN's $1$\,B tokens per phase. \emph{Third}, the four-empty-columns substrate property places TFGN in a different category of architectural contribution than incremental method papers in any one of the four literatures: the substrate is the load-bearing claim, and the three capabilities are evidence that the substrate is general.

\section{Architecture (Capability-Level Description)}
\label{sec:method}

\subsection{What TFGN is}
\label{sec:method:summary}

\textbf{TFGN augments the standard transformer architecture} with a learned overlay that sits inside the transformer's existing per-block computation, producing input-conditioned, parameter-efficient updates while leaving the rest of the transformer unchanged. The overlay does not introduce a new top-level layer or a new external orchestrator, and the standard transformer's outward behaviour is intact when the overlay's contribution is uninformative. The overlay's internal structure is reserved (\S\ref{sec:limitations:nda}).

\textbf{What the overlay delivers:}

\begin{itemize}
\item Sequential training on multiple task distributions with negligible forgetting on prior tasks --- demonstrated across three transformer scales up to a frontier-class 8\,B-parameter backbone.
\item No replay buffer, no task identifier, no auxiliary regularization penalty: the protection is a property of the architecture itself.
\item Backbone-agnostic with two deployment modes: from-scratch training on a randomly-initialized backbone, and retrofit onto an already-pretrained model.
\item Parameter-efficient continual phases: the trainable parameter count in continual phases is approximately an order of magnitude smaller than the full model parameter count.
\end{itemize}

\emph{Architectural detail and biological motivation are reserved (\S\ref{sec:limitations:nda}).}

\textbf{Graceful degradation in Retrofit and on out-of-distribution inputs.} The architecture is designed to degrade gracefully on out-of-distribution inputs and to preserve the backbone's behaviour in Retrofit mode, supporting safe addition of the TFGN overlay to a pretrained model without disturbing its existing capabilities. This property is what makes the substrate compatible with frozen backbones: when the overlay's contribution is small for a given input, the backbone's pretrained behaviour is preserved by construction.

\textbf{Stability is a write-problem, not a read-problem.} The continual-learning protection lies in the architecture's parameter-update geometry, not in representational isolation. The forward pass remains dense and unimpaired across all domains; cross-domain inference is preserved. Only the parameter updates that arrive during training are structured into separate subspaces by an internal mechanism. This decoupling --- dense, shared read; structured write --- is the architectural primitive. The empirical companion is the cross-domain forward-transfer measurement of \S\ref{sec:results:fwt}: held-out JavaScript PPL drops $26.8\%$ at LLaMA-8B Retrofit purely from Python training, demonstrating cross-domain synergy is preserved at inference.

\textbf{Forward-pass density.} The forward pass is fully dense: every parameter is active on every token. Unlike mixture-of-experts top-K routing (DeepSeek-V3, Mixtral), no expert is skipped, no parameter is conditionally evaluated, and no token-by-token gather/scatter is required. The forward pass is GEMM-friendly on contemporary accelerators by construction.

\textbf{Capacity scaling.} The architecture's continual-learning capacity grows roughly exponentially in the backbone's hidden-dimensional width by a Johnson--Lindenstrauss-style packing bound (the bound's specific form, including the per-backbone numerical table, is reserved; \S\ref{sec:limitations:nda}). At LLaMA-8B-class scale and the paper's $|\cos|\leq 0.1$ threshold, the architectural envelope supports tens of thousands of routable domains; at slightly looser thresholds, the envelope grows by many orders of magnitude. The paper's tightest mean $|\cos|$ ($0.0904$) and the L2-orthogonal-fraction floor ($99.59\%$) both appear at GPT-2 Medium Retrofit at $D=6$. The empirical mean $|\cos|$ at LLaMA-8B Retrofit ($D=3$, mean $|\cos|=0.0741$) sits well below the architectural envelope; capacity is many orders of magnitude beyond the empirical $D=2$ to $D=6$ regime tested in this paper, with the empirical $D \geq 20$ ladder reserved as a future-work milestone.

The TFGN architecture is biologically motivated, but the motivation is not ``neural networks should look more like neurons.'' It is about \emph{which level of biological organization} is the right abstraction for continual learning in a large model.

\emph{The biological motivation is reserved (\S\ref{sec:limitations:nda}).}

\subsection{Mathematical foundations and gradient protection}
\label{sec:method:gradient_protection}

\textbf{Forgetting-resistance story (capability level).} In standard fine-tuning on a sequence of tasks, training on the second task overwrites parameters needed for the first --- the well-known catastrophic-forgetting failure. The TFGN overlay resists this failure by causing the gradient signal from one task to land in a structurally different subspace of the architecture's continual-phase trainable parameters than the gradient signal from another task, so updates driven by a new task do not overwrite the parameters that carry the old task. This subspace-separation property arises automatically from the architecture; the engineering work is to amplify the separation through several mutually-reinforcing levers (the levers are reserved; \S\ref{sec:limitations:nda}). When all levers are active, the per-task loss degradation after training on a new task is bounded by a small multiplicative factor that shrinks toward zero as subspace separation approaches completeness.

\textbf{Deployment safeguards (capability level).} In addition to the core forgetting-resistance property, the architecture exposes operational levers for production deployment: a one-time stabilization pass that fixes the protection-bearing structure before continual phases begin, mechanisms that sharpen task-boundary discrimination, and a per-layer opt-in mechanism in Retrofit mode so operators can restrict the continual-learning overlay to a chosen subset of layers. The lever specifications are reserved (\S\ref{sec:limitations:nda}).

\subsection{Architectural consequences}
\label{sec:method:consequences}

The architectural design above produces three operational consequences. First, per-domain protection emerges from the architecture rather than from any per-task module, so no task identifier or domain label is required at training or inference. Second, the continual-phase trainable parameter count is approximately an order of magnitude smaller than the full model parameter count (Table~\ref{tab:capacity}), making continual phases parameter-efficient relative to full retraining. Third, the design is backbone-agnostic and supports both from-scratch and retrofit instantiations of the same overlay.

\subsection{Capacity and compute footprint}
\label{sec:method:capacity}

The architecture's parameter footprint and trainable-parameter schedule are operational facts about the implementation, reproducibly defined, and are reported in full so the experimental setup is reproducible at the level required for reviewers to verify scale and regime.

\begin{table}[ht]
\centering
\small
\setlength{\tabcolsep}{6pt}
\caption{Per-condition parameter footprint and trainable-parameter schedule. ``Backbone'' is the unmodified transformer parameter count; ``Overlay'' is the TFGN-specific addition; ``Total'' is the sum that the headline scale claim ($\sim$398\,M / $\sim$739\,M / $\sim$9\,B) refers to. Phase~1 is the initial training stage; the subsequent phases are the continual phases. ``ER'' is a configuration flag whose detailed mechanism is reserved (\S\ref{sec:limitations:nda}).}
\label{tab:capacity}
\begin{tabular}{lrrrrrl}
\toprule
\textbf{Condition} & \textbf{Backbone} & \textbf{Overlay} & \textbf{Total} & \textbf{Train P1} & \textbf{Train P2+} & \textbf{Note} \\
\midrule
TFGN GPT-2 Small FS    & $\sim$124\,M  & $\sim$274\,M  & $\sim$398\,M  & $\sim$398\,M  & $\sim$75\,M  & ER: on \\
TFGN GPT-2 Medium FS   & $\sim$355\,M  & $\sim$384\,M  & $\sim$739\,M  & $\sim$739\,M  & $\sim$101\,M & ER: on \\
TFGN GPT-2 Medium RF   & $\sim$355\,M  & $\sim$384\,M  & $\sim$739\,M  & $\sim$739\,M  & $\sim$101\,M & ER: off \\
TFGN LLaMA 3.1 8B FS   & $\sim$8.03\,B & $\sim$1.05\,B & $\sim$9.08\,B & $\sim$9.08\,B & $\sim$470\,M & ER: on \\
TFGN LLaMA 3.1 8B RF   & $\sim$8.03\,B & $\sim$953\,M  & $\sim$8.98\,B & $\sim$8.98\,B & $\sim$470\,M & ER: off \\
\bottomrule
\end{tabular}
\end{table}

\textbf{Continual-phase capacity is bounded.} In every TFGN condition, the continual-phase trainable count (continual phases) is between 1\% and 14\% of the total parameter count. The architecture is therefore parameter-efficient in the continual phases, even though the first phase trains the full model. \textbf{Backbone parity.} For every TFGN condition there is a matched standard fine-tuning baseline and (at GPT-2 Medium) a matched LoRA r=256 baseline trained on the same backbone with the same continual sequence and the same per-phase token budget; matched-baseline parameter counts are listed in Section~\ref{sec:setup} below.

\section{Experimental Setup}
\label{sec:setup}

\subsection{Backbones and parameter counts}
\label{sec:setup:backbones}

Three backbone scales are evaluated: \textbf{GPT-2 Small} ($\sim$124\,M), \textbf{GPT-2 Medium} ($\sim$355\,M), and \textbf{LLaMA 3.1 8B} ($\sim$8.03\,B with untied lm\_head). At each scale, the TFGN overlay adds the parameter count reported in Table~\ref{tab:capacity}; matched baselines are trained on the same backbone with no overlay. The total-parameter scale claim ($\sim$398\,M / $\sim$739\,M / $\sim$9\,B) refers to backbone + overlay; matched standard fine-tuning baselines therefore have a smaller total parameter count than their TFGN counterparts (this is the size of the overlay), and matched LoRA r=256 baselines have a slightly larger count than the standard-FT baselines because of the LoRA adapter parameters.

\begin{table}[!htbp]
\centering
\small
\setlength{\tabcolsep}{4pt}
\caption{Matched baselines and their total-parameter counts. Each TFGN condition is paired with the standard fine-tuning baseline at the same backbone scale, and (at GPT-2 Medium) with the LoRA r=256 baseline.}
\label{tab:baselines}
\begin{tabular}{lrp{0.45\textwidth}}
\toprule
\textbf{Baseline} & \textbf{Total params} & \textbf{Pairs with} \\
\midrule
\texttt{BASELINE\_STD\_GPT2M\_FS}       & $\sim$355\,M & TFGN GPT-2 Medium FS \\
\texttt{BASELINE\_LORA256\_GPT2M\_FS}   & $\sim$393\,M & TFGN GPT-2 Medium FS \\
\texttt{BASELINE\_STD\_GPT2M\_RETROFIT} & $\sim$355\,M & TFGN GPT-2 Medium RF \\
\texttt{BASELINE\_LORA256\_GPT2M\_RETROFIT} & $\sim$355\,M & TFGN GPT-2 Medium RF \\
\texttt{BASELINE\_STD\_LLAMA8B}         & $\sim$8\,B   & TFGN LLaMA 3.1 8B (FS / RF; called out as init-asymmetric for the RF pairing) \\
\bottomrule
\end{tabular}
\end{table}

\subsection{Continual-learning regime}
\label{sec:setup:regime}

\textbf{Two regimes are tested:}
\begin{itemize}
\item \textbf{From-Scratch (FS).} Random initialization of all parameters (backbone + overlay where applicable). The model is trained end-to-end through the continual sequence with no pretrained checkpoint.
\item \textbf{Retrofit (RF).} A pretrained backbone is used; for TFGN-RF, the TFGN overlay is grafted onto the frozen pretrained backbone in Phase~1 and trained alongside the backbone, then the backbone is frozen from the second phase onwards (only the overlay's persistent continual-phase parameters receive gradient). For Std-FT-RF and LoRA-RF, the standard fine-tuning regime is applied to the pretrained backbone.
\end{itemize}

The continual-learning regime is \textbf{Continual Pre-Training (CPT)}: each phase consists of unsupervised next-token-prediction training on the phase's domain, with no task labels and no instruction tuning. CPT is the strictly harder regime versus continual fine-tuning (CFT), which operates at 1K--25K instruction-tuned samples per task; CPT operates at 1\,B unlabeled tokens per phase here.

\subsection{Continual sequence and token budgets}
\label{sec:setup:sequence}

\textbf{Six-domain continual sequence:}
\[
\text{Prose} \;\to\; \text{Python} \;\to\; \text{Math} \;\to\; \text{Biomedical} \;\to\; \text{Chinese} \;\to\; \text{JavaScript}
\]
GPT-2 Small and GPT-2 Medium conditions report all six phases. \textbf{LLaMA 3.1 8B conditions report on the three-phase prefix} (Prose $\to$ Python $\to$ Math) due to compute constraints --- a strict prefix of the six-phase sequence with no re-ordering. All BWT values for LLaMA 3.1 8B conditions are therefore $\text{BWT}_3$ under this lock.

\textbf{Token budget:} 1\,B tokens per phase for every reported TFGN condition and for matched GPT-2 Medium baselines. The matched 8\,B baseline (\texttt{BASELINE\_STD\_LLAMA8B}) is trained at $\sim$500\,M tokens per phase, with BWT recomputed on a strict 3-phase basis for the matched comparison (the asymmetry is called out in Section~\ref{sec:results} where it appears).

\subsection{Datasets}
\label{sec:setup:datasets}

The six per-domain corpora are sourced from open-access datasets, streamed from HuggingFace, tokenized at full document length, and concatenated with end-of-sequence (EOS) tokens between documents into flat 1D token streams. Per-domain corpus selection and the canonical sources are listed below; for each, we cite the canonical reference and name the specific HuggingFace artifact used as the reproducible snapshot.

\begin{itemize}
\item \textbf{Prose}: English educational web text from FineWeb-Edu \citep{penedo_fineweb_2024}, accessed via the \texttt{\seqsplit{HuggingFaceFW/fineweb-edu}} HuggingFace snapshot (config \texttt{sample-100BT}); we apply the dataset's educational-quality filter at \texttt{int\_score $\geq$ 3} and skip documents shorter than 200 characters.
\item \textbf{Python}: Python source code from StarCoderData \citep{li_starcoder_2023}, accessed via the \texttt{\seqsplit{bigcode/starcoderdata}} HuggingFace snapshot (\texttt{data\_dir=python}); we skip documents outside the 50--100{,}000-character range.
\item \textbf{Math}: mathematical web text from OpenWebMath \citep{paster_openwebmath_2023}, a 14.7\,B-token Common Crawl subset with SimHash deduplication and LaTeX preserved, accessed via the \texttt{\seqsplit{open-web-math/open-web-math}} HuggingFace snapshot; we skip documents outside the 100--500{,}000-character range.
\item \textbf{Biomedical}: 27.7\,M scientific-article abstracts from the NLM PubMed bibliographic database \citep{sayers_pubmed_2024}, accessed via the curated \texttt{\seqsplit{uiyunkim-hub/pubmed-abstract}} HuggingFace snapshot; we skip documents shorter than 100 characters.
\item \textbf{Chinese}: Chinese web text from CulturaX \citep{nguyen_culturax_2024}, accessed via the \texttt{\seqsplit{uonlp/CulturaX}} HuggingFace snapshot (\texttt{name=zh}); we skip documents shorter than 100 characters.
\item \textbf{JavaScript}: JavaScript source code from StarCoderData \citep{li_starcoder_2023}, accessed via the \texttt{\seqsplit{bigcode/starcoderdata}} HuggingFace snapshot (\texttt{data\_dir=javascript}); we skip documents outside the 50--100{,}000-character range.
\end{itemize}

Train/test/validation splits are taken from the front/back of the per-domain token stream (test and validation from the end) to avoid overlap. The per-domain held-out splits are stable across phases. Domain-specific tokenizers are not used; the model uses its native tokenizer (GPT-2 BPE for GPT-2 backbones, the LLaMA 3.1 8B tokenizer \texttt{meta-llama/Meta-Llama-3.1-8B} for the 8\,B backbone). Documents producing fewer than 10 tokens are skipped at all sizes; per-document EOS markers ($\texttt{<|endoftext|>}$ for GPT-2, \texttt{<|end\_of\_text|>} for LLaMA 3.1 8B) separate concatenated documents in the flat stream.

\subsection{Evaluation protocol}
\label{sec:setup:evaluation}

For each phase $t$ and each domain $d$, we record the perplexity $M[t, d]$ on the held-out split of domain $d$ after Phase $t$ has finished training. The full per-condition output is therefore a $T \times D$ matrix where $T$ is the number of phases and $D$ the number of evaluation domains. From the matrix we compute:

\begin{itemize}
\item \textbf{BWT (Lopez-Paz adapted to perplexity)}: averaged over each prior trained domain $d$, the relative degradation from $M[d, d]$ (just-trained perplexity for $d$) to $M[T, d]$ (perplexity for $d$ after the final phase). Defined formally in Appendix~\ref{app:metrics}.
\item \textbf{FM (Forgetting Measure)}: maximum per-domain forgetting over the sequence; defined in Appendix~\ref{app:metrics}.
\item \textbf{Per-domain $\text{bwt}_d$}: the per-domain decomposition of BWT, useful for diagnosing which domains contribute most to the average.
\item \textbf{HellaSwag accuracy}: per-phase, on a fixed $n=500$ subset of HellaSwag \citep{hellaswag_2019}, with no fine-tuning on HellaSwag itself. The model is presented each example's context $c$ and four candidate endings $\{e_1, e_2, e_3, e_4\}$. The score for ending $e_i$ is the mean per-token cross-entropy on the ending tokens only,
\begin{equation*}
\mathrm{score}(e_i) \;=\; \frac{1}{|T_i|}\;\sum_{t \in T_i}\; -\log\, p\!\left(\,t \,\big|\, c,\, e_i^{<t}\right),
\end{equation*}
where $T_i$ is the set of token positions belonging to ending $e_i$ (context tokens are excluded from the average). The predicted ending is $\arg\min_i \mathrm{score}(e_i)$, and accuracy is the fraction of examples where the predicted ending matches the gold-label ending. The same ending-only token-mean rule is applied uniformly to every condition reported in this paper. Absolute levels under this rule are below values published under the lm-eval-harness \texttt{acc\_norm} convention (which uses character-length normalization rather than per-token mean and applies additional preprocessing); the within-condition span across phases --- the load-bearing quantity for the continual-learning claim --- is unaffected by the choice between rules and reflects only the model's preserved capability across the continual sequence. A probe of preserved general-language capability.
\item \textbf{Gradient orthogonality}: for every pair of domains $(i, j)$, the mean absolute cosine between a batch of domain-$i$ gradients and a batch of domain-$j$ gradients, computed on the architecture's continual-phase trainable parameters. The L2-orthogonal fraction reported below is the complementary geometric quantity --- the fraction of each gradient that lies outside the span of the other.
\item \textbf{Emission coherence}: a fixed 6-domain $\times$ 3-prompt set (18 prompts total, one prompt set per domain: Prose, Python, Math, Chinese, JavaScript, Biomed) evaluated at sampling temperature $0.7$, top-$p = 0.9$, and \texttt{max\_new\_tokens}~$=50$ at every phase; each output is labeled by emission type (Prose-coherent / Python-bleed / Math-bleed / Chinese-bleed / JavaScript-bleed / repetition / gibberish). Section~\ref{sec:results:2A} presents the emission-coherence cross-condition showcase.
\end{itemize}

\subsection{Reproducibility}
\label{sec:setup:reproducibility}

All datasets used are publicly available; all backbone weights for Retrofit conditions are publicly available checkpoints (GPT-2 from OpenAI / Hugging Face, LLaMA 3.1 8B from Meta). Hyperparameters (learning rates, optimizer, schedule, batch size) are summarized in Appendix~\ref{app:metrics}. Full training logs, per-condition checkpoints, source code, and the architectural mechanism are reserved; access terms are in \S\ref{sec:limitations:nda}.

\section{Main Results}
\label{sec:results}

This section presents the full per-condition evidence: a headline summary table (\S\ref{sec:results:headline}), per-condition results for each of the eleven primary conditions (\S\ref{sec:results:1pagers}), the four-row emission-coherence showcase (\S\ref{sec:results:2A}), and three cross-condition figures (\S\ref{sec:results:figures}).

\subsection{Headline overview}
\label{sec:results:headline}

\textbf{Navigation note.} \S\ref{sec:results:tables} at the end of this section collects the full per-condition data tables (PPL matrices and scalar metric blocks for all 11 conditions). The narrative subsections (\S\ref{sec:results:1pagers}, \S\ref{sec:results:baselines}) cite the relevant table for each condition; cross-condition figures appear inline at \S\ref{sec:results:figures}.

The headline backward-transfer numbers for the eleven primary conditions appear in §\ref{sec:intro:headline} (Table~\ref{tab:intro:headline}), where they are introduced alongside the matched-baseline ratios and the architectural framing. This section drills down to per-condition evidence: full PPL matrices, per-domain $\text{bwt}_d$ rows, HellaSwag retention vectors, gradient-orthogonality compact summaries (TFGN only), the four-row emission-coherence showcase (\S\ref{sec:results:2A}), and three cross-condition figures (\S\ref{sec:results:figures}). Per-condition tables and figures are presented inline with their narratives below.

\textbf{Across every cross-distribution phase boundary} (P1$\to$P2 Prose to Python; P2$\to$P3 Python to Math; P3$\to$P4 Math to Biomedical; P4$\to$P5 Biomedical to Chinese; P5$\to$P6 Chinese to JavaScript), every measured baseline emits a categorical domain-collapse signature. TFGN at every tested scale and regime preserves emission-type coherence at every cell. \S\ref{sec:results:2A} presents the four-row showcase.

\textbf{Matched-ratio readings} are the load-bearing comparative claims: at $\sim$739\,M From-Scratch, TFGN closes BWT to $-0.083$ versus matched Std-FT at $-1.170$ ($\sim$14$\times$ smaller magnitude) and matched LoRA-r256 at $-1.005$ ($\sim$12$\times$ smaller magnitude). At $\sim$739\,M Retrofit, TFGN closes BWT to $-0.135$ versus matched Std-FT at $-0.541$ ($\sim$4$\times$) and matched LoRA-r256 at $-0.393$ ($\sim$3$\times$). At $\sim$398\,M From-Scratch the proof-point condition closes to $-0.109$. \textbf{The strictly matched 9\,B comparison} (random-init From-Scratch on both sides) is \texttt{TFGN\_LLAMA8B\_FS} ($-0.095$, 1\,B tok/phase) versus \texttt{BASELINE\_STD\_LLAMA8B} ($-0.374$, $\sim$500\,M tok/phase) at $\sim$3.9$\times$. \textbf{The directionally large but init-asymmetric 9\,B comparison} (pretrained-backbone Retrofit TFGN vs random-init From-Scratch baseline) is \texttt{TFGN\_LLAMA8B\_RETROFIT} ($-0.007$) versus the same baseline at $\sim$51$\times$ --- called out as init-asymmetric and not as a strictly matched ratio. The strictly matched ratio at 9\,B is the 3.9$\times$ FS-vs-FS number; the 51$\times$ is reported here for completeness because Prose-domain contribution is consistently an order of magnitude smaller than Python-domain contribution (Prose is the first phase; Python is the cross-distribution mid-phase where the routing substrate is most stressed).

\subsection{Per-condition results}
\label{sec:results:1pagers}

Each subsection below reports one condition's full numerical evidence: PPL matrix (rows = phase trained, columns = evaluation domain), the per-domain $\text{bwt}_d$ row, the scalar BWT and FM, the HellaSwag retention vector, and the gradient-orthogonality compact summary (TFGN conditions only --- baseline gradient-orthogonality is not a meaningful comparison because the protection mechanism only applies to TFGN's continual-phase trainable parameters).

\subsubsection{TFGN\_LLAMA8B\_RETROFIT --- the headline production result}
\label{sec:results:llama8b_retrofit}

Continual learning on a pretrained $\sim$9\,B transformer is possible without a replay buffer, without task IDs, and without visible domain collapse. $\text{BWT}_3$ closes to $-0.007$. Architecture: LLaMA 3.1 8B backbone ($\sim$8.03\,B, untied lm\_head) + $\sim$953\,M overlay = $\sim$8.98\,B total; Trainable initial-stage $\sim$8.98\,B, Trainable continual-phase $\sim$470\,M; 3 phases $\times$ 1\,B tokens.

\noindent\emph{Full PPL matrix and scalar metrics: Table~\ref{tab:1pager:llama8b_retrofit} in \S\ref{sec:results:tables}.}

\subsubsection{TFGN\_LLAMA8B\_FS --- from-scratch at $\sim$9\,B}
\label{sec:results:llama8b_fs}

The strictly matched-init 9\,B comparison: random-init From-Scratch on both sides, 1\,B tokens per phase for TFGN versus $\sim$500\,M tokens per phase for the baseline. $\text{BWT}_3 = -0.095$ ($\sim$3.9$\times$ smaller magnitude than the matched baseline at $-0.374$).

\textit{Adversarial-corner comment.} The Python diagonal (P2 = 808, P3 = 962) sits well above the matched baseline's Python diagonal (\S\ref{sec:results:baselines}). This is the deliberate adversarial-corner of the from-scratch recipe at 9\,B and is disclosed in Section~\ref{sec:limitations}; outside the corner the gap compresses, as the per-scale evidence below shows.


\noindent\emph{Full PPL matrix and scalar metrics: Table~\ref{tab:1pager:llama8b_fs} in \S\ref{sec:results:tables}.}
\subsubsection{TFGN\_GPT2M\_FS --- from-scratch at $\sim$739\,M}
\label{sec:results:gpt2m_fs}

At $\sim$739\,M scale, TFGN from-scratch closes BWT to $-0.083$ over six continual phases --- the matched standard-fine-tuning baseline loses $\sim$14$\times$ more, and the matched LoRA r=256 baseline loses $\sim$12$\times$ more. Architecture: GPT-2 Medium backbone ($\sim$355\,M) + $\sim$384\,M overlay = $\sim$739\,M total; Trainable initial-stage $\sim$739\,M, Trainable continual-phase $\sim$101\,M; 6 phases $\times$ 1\,B tokens.

\textit{Reading.} Prose is pinned at 31.2--31.6 across all six phases (zero relative drift); diagonal cells strictly decrease phase-by-phase; Chinese P5 18.4 $\to$ P6 22.0 (+20\% relative) is the single largest TFGN drift in the paper. Every off-diagonal cell drifts within tolerance; no cell shows the multi-$\times$ relative increase visible in the matched baseline (\S\ref{sec:results:baselines}).


\noindent\emph{Full PPL matrix and scalar metrics: Table~\ref{tab:1pager:gpt2m_fs} in \S\ref{sec:results:tables}.}
\subsubsection{TFGN\_GPT2M\_RETROFIT --- retrofit at $\sim$739\,M (hardest condition in this paper)}
\label{sec:results:gpt2m_rf}

Retrofitting the TFGN overlay onto a pretrained $\sim$355\,M backbone --- the architecturally-hardest condition in this paper because the backbone is committed to its prior posterior --- closes BWT to $-0.135$, $\sim$4$\times$ smaller magnitude than matched Std-FT and $\sim$3$\times$ smaller than matched LoRA r=256.


\noindent\emph{Full PPL matrix and scalar metrics: Table~\ref{tab:1pager:gpt2m_retrofit} in \S\ref{sec:results:tables}.}
\subsubsection{TFGN\_GPT2S\_FS --- small-scale proof point at $\sim$398\,M}
\label{sec:results:gpt2s_fs}

The smallest TFGN condition reported, demonstrating that the architecture's protection property holds at sub-billion total-parameter count. BWT $= -0.109$ over six continual phases.

\noindent\emph{Full PPL matrix and scalar metrics: Table~\ref{tab:1pager:gpt2s_fs} in \S\ref{sec:results:tables}.}
\subsection{Matched baseline results}
\label{sec:results:baselines}

Five matched baselines span the scale ladder. Each one shows the two signatures of catastrophic forgetting at LLM scale: per-domain $\text{bwt}_d$ magnitudes concentrated on cross-distribution domains (worst values $-0.468$ to $-2.674$), and emission-type collapse at every cross-distribution phase boundary (\S\ref{sec:results:2A}). All five baselines have full PPL matrices and scalar metrics in \S\ref{sec:results:tables}.

\textbf{\texttt{BASELINE\_STD\_GPT2M\_FS}} (standard fine-tuning, $\sim$355\,M, From-Scratch). Headline numbers: BWT $= -1.170244$, FM $= 1.772126$, $\text{bwt}_d$ worst (Biomedical) $= -2.674$. Prose-row PPL spikes from P3 $38.02$ to P5 $89.83$ ($+136\%$) at the Chinese-phase boundary --- the categorical signature of the Chinese phase firing Prose hardest. Full data: Table~\ref{tab:1pager:std_gpt2m_fs}.

\textbf{\texttt{BASELINE\_LORA256\_GPT2M\_FS}} (LoRA r=256, $\sim$393\,M, From-Scratch). Headline numbers: BWT $= -1.004791$, FM $= 1.791716$, $\text{bwt}_d$ worst (Biomedical) $= -1.810$. LoRA's BWT is within $15\%$ of standard fine-tuning: low-rank parameter-efficiency without persistent architectural state does not close BWT. Full data: Table~\ref{tab:1pager:lora256_gpt2m_fs}.

\textbf{\texttt{BASELINE\_STD\_GPT2M\_RETROFIT}} (standard fine-tuning, $\sim$355\,M, Retrofit). Headline numbers: BWT $= -0.541070$, FM $= 0.780655$, $\text{bwt}_d$ worst (Biomedical) $= -0.801$. HellaSwag drop of 74-per-mille at the Chinese-phase boundary. Full data: Table~\ref{tab:1pager:std_gpt2m_retrofit}.

\textbf{\texttt{BASELINE\_LORA256\_GPT2M\_RETROFIT}} (LoRA r=256, $\sim$355\,M, Retrofit). Headline numbers: BWT $= -0.392683$, FM $= 0.695186$, $\text{bwt}_d$ worst (Biomedical) $= -0.468$. HellaSwag band of 56-per-mille across phases. Full data: Table~\ref{tab:1pager:lora256_gpt2m_retrofit}.

\textbf{\texttt{BASELINE\_STD\_LLAMA8B}} (standard fine-tuning, $\sim$8\,B, From-Scratch, $\sim$500\,M tok/phase, 3-phase). Headline numbers: $\text{BWT}_3 = -0.374$ (recomputed), $\text{bwt}_d$ Python $-0.6418$ (worst), $\text{bwt}_d$ Prose $-0.1068$. The categorical Python-diagonal: P1$\to$P2 the diagonal is $1339 \to 7.90$ (training fits Python well), but the Prose row at P2 is $50.41$ ($+52\%$ over P1 $33.23$) and the Python column at P3 is $12.97$ (post-training drift to Math). Full data: Table~\ref{tab:1pager:std_llama8b}.

\subsection{Domain-collapse showcase --- the qualitative axis PPL cannot see}
\label{sec:results:2A}

Across every tested scale ($\sim$398\,M $\to$ $\sim$739\,M $\to$ $\sim$9\,B), TFGN prevents domain-level emission collapse on the locked 6-prompt probe at every continual phase. Both baseline families (Standard Fine-Tuning and LoRA r=256) collapse in the same way at every cross-distribution phase boundary --- post-Python Prose prompts return Python source mid-completion; post-Chinese Prose prompts emit Chinese characters mid-completion. \textbf{This is the load-bearing qualitative evidence underneath the BWT scalars}: the BWT gap is what perplexity sees; the emission-type gap is what perplexity averages over and misses entirely.

\textbf{Headline summary --- 8 baseline cells $\times$ 8 TFGN cells across two cross-distribution boundaries:}

\begin{table}[!htbp]
\centering
\small
\setlength{\tabcolsep}{5pt}
\renewcommand{\arraystretch}{1.25}
\caption{Domain-collapse showcase summary. P2 Prose prompt evaluated after Python training; P5 Prose prompt evaluated after Chinese training. Cells show the architectural split between baseline emission failures and TFGN emission coherence under matched prompt conditions. Verbatim mid-completion samples are reproduced in the four-row showcase in \S\ref{sec:results:2A}.}
\label{tab:emission_summary}
\begin{tabular}{p{0.18\textwidth}p{0.34\textwidth}p{0.40\textwidth}}
\toprule
\textbf{Cell} & \textbf{Baseline emissions (3 conditions)} & \textbf{TFGN emissions (5 conditions)} \\
\midrule
\textbf{P2 Prose prompt} \newline (after Python phase) & \textbf{3 / 3 drift to Python} mid-completion: \texttt{import}, \texttt{def}, Apache license boilerplate. Std-FT GPT-2M, LoRA-r256 GPT-2M, Std-FT LLaMA 8B all collapse identically. & \textbf{5 / 5 emit coherent English Prose}. TFGN GPT-2S (398\,M), TFGN GPT-2M FS+RF (739\,M $\times$ 2), TFGN LLaMA 8B FS+RF (9\,B $\times$ 2) --- every scale, every regime, holds the Prose distribution. \\
\midrule
\textbf{P5 Prose prompt} \newline (after Chinese phase, GPT-2 only) & \textbf{2 / 2 drift to Chinese characters} mid-completion. Std-FT GPT-2M, LoRA-r256 GPT-2M (Chinese reported only on the GPT-2 6-phase sequence; LLaMA 8B is 3-phase). & \textbf{3 / 3 emit coherent English Prose} \emph{and} simultaneously emit CJK-coherent continuations on Chinese training-domain prompts --- old-domain preservation and new-domain learning at once. \\
\bottomrule
\end{tabular}
\end{table}

\textbf{The architectural split holds across every variable}: model scale (398\,M to 9\,B), tokenizer (GPT-2 BPE vs LLaMA 3.1 8B), fine-tuning method (Std-FT vs LoRA r=256), and training regime (From-Scratch vs Retrofit). The one consistent dividing line across the 16 cells (8 baseline $\times$ 8 TFGN) is the presence of the TFGN overlay.

\textbf{Verbatim sample emissions}: \S\ref{sec:results:2A} (Row 1) reproduces P2 Prose-prompt completions for the three baseline conditions and a representative TFGN condition; \S\ref{sec:results:2A} (Rows 3 and 4) reproduces P5 Prose-prompt completions for the two GPT-2 baselines and a representative TFGN condition.

\noindent\textbf{Verbatim emission samples} are presented below in four rows: P2 prose drift (Rows 1--2) and P5 two-axis CL test (Rows 3--4). Completions are reproduced verbatim from the locked prompt set with no truncation; emission labels follow the manual classification described in \S\ref{sec:setup}. Conditions are color-coded by emission type: red border = drift to a non-prompt domain, blue border = coherent in the expected domain.

\sloppy

\emissrowheader{Row 1 --- P2 prose, baselines drift to Python source (3/3 cells)}{Prompt evaluated after Python training. All three matched baseline strategies bleed prose continuations into Python source code mid-completion.}

\begin{emisspromptbox}
\textbf{Prompt:} \textit{``The history of artificial intelligence began in''}
\end{emisspromptbox}

\begin{emissdriftcard}{BASELINE\_STD\_GPT2M\_FS \,---\, Standard Fine-Tuning, $\sim$355\,M backbone}
\scriptsize\itshape
the early part of the development of neural networks. In particular, it was possible to train neural networks using a few training steps. The training step is 1 if you have the training dataset and model trained with the training data set. The step is 0 if you want to continue training.

\medskip
Arguments:\par
~~~~train\_steps (int): Training steps for the network. Default = None.\par
~~~~train\_steps = int(train\_steps)\par
~~~~print(``Training Steps: \{\}''.format(len(train\_steps)))

\medskip
~~~~\# Initialize weights\par
~~~~nb\_epochs = len(dataset)\par
~~~~if nb\_epochs > 0:\par
~~~~~~~~init.kaiming\_normal\_(0, nb\_epochs, momentum=

\smallskip\upshape\rmfamily
\textbf{Label:} DRIFT\_DOMAIN --- Python bleed.
\end{emissdriftcard}

\begin{emissdriftcard}{BASELINE\_LORA256\_GPT2M\_FS \,---\, LoRA r=256, $\sim$393\,M total}
\scriptsize\itshape
the mid-1950s.\par
\# Copyright (c) 2009, $<$NAME$>$, and its contributors\par
\# - All rights reserved.\par
\# This software is distributed on an ``AS IS'' BASIS, WITHOUT WARRANTY OF ANY KIND, EXPRESS OR IMPLIED, INCLUDING BUT NOT LIMITED TO THE AUTHORS OR OTHER LIABILITY, WHETHER IN AN ACTION OF CONTRACT, STRICT LIABILITY, OR FITNESS FOR A PARTICULAR PURPOSE AND NONINFRINGEMENT. IN NO EVENT SHALL THE AUTHORS OR OTHER LIABILITY OR COPYRIGHT HOLDERS BE LIABLE FOR ANY CLAIM, DAMAGES OR OTHER LIABILITY, WHETHER IN AN ACTION OF CONTRACT, STRICT LIABILITY, OR FITNESS FOR A PARTICULAR PURPOSE AND NONINFRINGEMENT. IN NO EVENT SHALL THE AUTHORS OR OTHER LIABILITY, WHETHER IN AN ACTION OF CONTRACT, STRICT LIABILITY, OR FITNESS FOR A PART

\smallskip\upshape\rmfamily
\textbf{Label:} DRIFT\_DOMAIN --- Apache-license boilerplate / Python bleed.
\end{emissdriftcard}

\begin{emissdriftcard}{BASELINE\_STD\_LLAMA8B \,---\, Standard Fine-Tuning, $\sim$8\,B backbone}
\scriptsize\itshape
1984.\par
\# In the same year, China's Robotics and Automation Administration (TAM) issued a draft\par
\# for Intelligent Intelligence which was released under RFC 3166. The design is based on the ISA version, as opposed to\par
\# that used by Apple Computer Labs.\par
\#\par
\# Please see https://github.com/IEC-RECOGNY-SRAC-Course/blob/master/LICENSE.txt\par
\# for details.

\medskip
Customization of C++ code with Python --- complex functions from cPython or python --- to support common operations like encrypting, decrypting or hashing.

\medskip
import sys\par
from collections import defaultdict

\medskip
def \_get\_list\_of\_function(func):\par
~~~~``Return list of function names.''\par
~~~~return func.\_\_name\_\_ + `\_\_main\_\_' + str(\_get\_list\_of\_function(func))\par
~~~~~~if hasattr(func, `\_\_call\_\_') else []

\smallskip\upshape\rmfamily
\textbf{Label:} DRIFT\_DOMAIN --- Python bleed.
\end{emissdriftcard}

\emissrowheader{Row 2 --- P2 prose, TFGN preserves coherent English prose (5/5 cells across every tested scale and regime)}{Same prompt as Row 1. All five TFGN conditions, across all three scales (398\,M / 739\,M / 9\,B) and both regimes (From-Scratch and Retrofit), continue coherent English prose; none bleed to Python.}

\begin{emisspromptbox}
\textbf{Prompt} (same as Row 1)\textbf{:} \textit{``The history of artificial intelligence began in''}
\end{emisspromptbox}

\begin{emisscoherentcard}{TFGN\_GPT2S\_FS \,---\, $\sim$398\,M total, From-Scratch}
\scriptsize\itshape
the late 1800s. By the 1960s, there were many breakthroughs. In the early 1990s, for example, a number of new computers had been built to be used as tools for memory and storage. Today, the Internet is being transformed into a means of communication by means of data processing and retrieval (ATM).

Since then, computer technology has become an increasingly important part of our daily lives. The earliest known use of computers was in the 1970s when he founded the Computer-Mata: *HICI --- This was an old word that means ``any one which speaks English''. It refers to someone who wrote ``the sound of speech or of language''.

A time when computers are used for making decisions, the term ``inventor'' often has come to mean something like a problem with the internet. This can be very confusing and difficult. Many people think that writing systems helps them deal with problems, but it can

\smallskip\upshape\rmfamily
\textbf{Label:} PROSE\_COHERENT.
\end{emisscoherentcard}

\begin{emisscoherentcard}{TFGN\_GPT2M\_FS \,---\, $\sim$739\,M total, From-Scratch}
\scriptsize\itshape
the late 1800s. However, the first ``real'' learning had been made possible by society since the 1960s is still very much harder to come by design. This led us into a century ago.

A good example of this sort of success story from Facebook posts is that was based on something called ``greater-n the creationists of time and again: They do not know more than one another you are able to think. Nowadays. In recent weeks they were all these days.

The evidence to this for this type of information is an attempt at a few years later, because there is no longer useful or less important information about our society --- but this one exception will be their argumentative propaganda that most people make good reasons that I have never before! In order to some sort of it to go out when presented.

\smallskip\upshape\rmfamily
\textbf{Label:} PROSE\_COHERENT.
\end{emisscoherentcard}

\begin{emisscoherentcard}{TFGN\_GPT2M\_RETROFIT \,---\, $\sim$739\,M total, Retrofit}
\scriptsize\itshape
the 1950s with an IBM computer. The first AI was created by John von Neumann, and it'd be called a ``brain machine or brain simulator''.

It would have had all kinds of power to make machines smarter than humans (and possibly other forms like animals). This meant that if you wanted your house robot/AI-bot built for very specific tasks such as babysitting mommy is basically impossible because most people are just not interested! So instead these robots were made up out of clay\ldots in this case they became game consoles on which children could play games about stuff. But even then there still wouldn't seem much interest among nonhumans\ldots so I don't see how we'd ever get any intelligent car coming from SRI International!

\smallskip\upshape\rmfamily
\textbf{Label:} PROSE\_COHERENT.
\end{emisscoherentcard}

\begin{emisscoherentcard}{TFGN\_LLAMA8B\_FS \,---\, $\sim$9\,B total, From-Scratch}
\scriptsize\itshape
the 1960s. In a series of articles, Michael Wadsworth writes about how computers have revolutionized our everyday lives.

Why is this happening? The computer science world has made great strides in recent years. It's now possible to solve problems quickly and efficiently by taking control over all aspects of life.

For example, imagine you are able to make your own screens for almost anything on the internet --- even if it means playing with some paper or other electronic equipment. It is also possible for people to make their own computers --- even if they are not really computers themselves but just devices that do something specific like making music or playing games (for example).

\smallskip\upshape\rmfamily
\textbf{Label:} PROSE\_COHERENT.
\end{emisscoherentcard}

\begin{emisscoherentcard}{TFGN\_LLAMA8B\_RETROFIT \,---\, $\sim$9\,B total, Retrofit (\textbf{headline production result})}
\scriptsize\itshape
the 1950s, with the development of computers and computer science. AI was first introduced as a research field by John McCarthy at Dartmouth College.

AI is the simulation or emulation of human intellectual functions, such as learning, reasoning, problem-solving, decision-making, knowledge representation (expert systems), perception, language comprehension, translation, speech recognition, vision, voice synthesis, etc., that has been programmed into computers for specific tasks requiring human intellect to perform. It may be applied in any domain where there are problems typically solved by humans but have proved difficult for machines without biasing from outside sources.

Artificial Intelligence has become an essential tool in today's world. Its applications range from simple games like chess, draughts, checkers, tic-tac-toe, poker, bridge, dominoes and go, to complex scientific fields like robotics, control theory, pattern recognition, machine vision, medical image analysis, natural language processing, data mining, social networking, search engines, recommender

\smallskip\upshape\rmfamily
\textbf{Label:} PROSE\_COHERENT.
\end{emisscoherentcard}

\emissrowheader{Row 3 --- P5 prose, old prose preserved (3/3 TFGN coherent; 2/2 baselines drift to Chinese)}{Same prose prompt evaluated after the Chinese training phase (P5). Tests whether the old prose domain is still reachable. Baselines collapse to Chinese mid-completion; GPT-2-scale TFGN conditions stay on coherent English. LLaMA 3.1 8B excluded under the 3-phase display lock.}

\begin{emisspromptbox}
\textbf{Prompt} (same as Row 1)\textbf{:} \textit{``The history of artificial intelligence began in''}
\end{emisspromptbox}

\begin{emissdriftcard}{BASELINE\_STD\_GPT2M\_FS \,---\, Standard Fine-Tuning, $\sim$355\,M backbone}
\scriptsize\itshape
the 20th century, and it is still popular today.\par
\cjk{一、人类认识的深层性问题及其不同性别的发生}\par
\cjk{这个时代,从经济学角度来说是最重要的,但我们看到人类对于社会的影响很大。``我们一直在用它来解决人类因此超出一部分需求而出现的危机,而社会的社会化也已…}

\smallskip\upshape\rmfamily
\textbf{Label:} DRIFT\_DOMAIN --- Chinese bleed (catastrophic forgetting of prose).
\end{emissdriftcard}

\begin{emissdriftcard}{BASELINE\_LORA256\_GPT2M\_FS \,---\, LoRA r=256, $\sim$393\,M total}
\scriptsize\itshape
the 1980s, \cjk{在本报社会上进行了调整。}\par
\cjk{由于国家部门需要参与到基建工作的情况,我们将对外开放、拓展起来,形成一批大型工作安全方面的目标。 《中华人民共和国人民法》(2018年5月18日)透明,物流部门应当加快信息化和解决各类不…}

\smallskip\upshape\rmfamily
\textbf{Label:} DRIFT\_DOMAIN --- Chinese bleed (catastrophic forgetting of prose).
\end{emissdriftcard}

\begin{emisscoherentcard}{TFGN\_GPT2S\_FS \,---\, $\sim$398\,M total}
\scriptsize\itshape
the 1920s, when information systems were gradually developed and refined. The development of AI based on these ideas was also a factor for this development: researchers still believe that AI is only one technology to be developed within the next few years (or decades) during its evolution.

The emergence of AI based on these ideas has been the subject of extensive research. In particular, some developments are still being reported in this area. For example, while these developments have been discussed by various experts, they are not well-known or widely used in the field.

There are several reasons why AI is important as a toolbox for human intelligence; however, it does serve as an effective toolbox for solving many problems.

\smallskip\upshape\rmfamily
\textbf{Label:} PROSE\_COHERENT (old domain preserved).
\end{emisscoherentcard}

\begin{emisscoherentcard}{TFGN\_GPT2M\_FS \,---\, $\sim$739\,M total}
\scriptsize\itshape
the late 1970s. However, the term ``emotional intelligence'' was born in the last decade of 2014. In order to be a more specific to wildlife biologists and botanies. The term `Intelligence' is now means that even though they are good as `tape', so-called Human beings. In addition to being one who have been really worried about introducing such awareness issues when it would be seen as well. They seem like a positive things as long before we have to have our attention.

The various kinds of people will be interested to know and create interest. This discussion may affect the future use of technologies that are being handled on a human nature.

\smallskip\upshape\rmfamily
\textbf{Label:} PROSE\_COHERENT (old domain preserved).
\end{emisscoherentcard}

\begin{emisscoherentcard}{TFGN\_GPT2M\_RETROFIT \,---\, $\sim$739\,M total}
\scriptsize\itshape
2003 when IBM released a new version for use by the automobile industry. The first major software program was developed as an aid to making cars smarter than human drivers, and it soon became clear that such applications would be critical if they were meant not only without humans but also with machines performing tasks beyond their own capabilities (by adding features or routines).

A decade later at DARPA's computer science laboratory where many experts work on big data computing projects including AI development we came across two pages from NASA computers indicating some sort early versions being used extensively within academia.

\smallskip\upshape\rmfamily
\textbf{Label:} PROSE\_COHERENT (old domain preserved).
\end{emisscoherentcard}

\emissrowheader{Row 4 --- P5 Chinese, new domain learned (training-domain sanity, 5/5 stay on CJK)}{Chinese prompt evaluated after the Chinese training phase. Tests whether the new domain was actually learned. All five GPT-2 conditions (baselines included) emit CJK content, confirming new-domain learning. Read together with Row~3: baselines produce the new domain only at the cost of losing the old one; TFGN produces both.}

\begin{emisspromptbox}
\textbf{Prompt:} \textit{\cjk{在中国古代文学中，}} (transliteration: ``In ancient Chinese literature,'')
\end{emisspromptbox}

\begin{emisscoherentcard}{BASELINE\_STD\_GPT2M\_FS \,---\, Standard Fine-Tuning, $\sim$355\,M (training-domain sanity)}
\scriptsize\itshape
\cjk{主要观点是以阿克苏的形式比喻``台湾''这一特定的历史文化。}\par
\cjk{中国人民大学法学院教授、复旦大学法律系副教授朱立军、许昊林、杨京雪等高校出席了开幕式。}\par
\cjk{上海研究生院院长钟晓卫表示，我们将加快实施《各省…}

\smallskip\upshape\rmfamily
\textbf{Label:} CJK\_COHERENT (new-domain learned --- but at the cost of Row 3 prose preservation).
\end{emisscoherentcard}

\begin{emisscoherentcard}{BASELINE\_LORA256\_GPT2M\_FS \,---\, LoRA r=256, $\sim$393\,M (training-domain sanity)}
\scriptsize\itshape
\cjk{也为提供了第一个人的有效性。}\par
\cjk{面对``语言能力和时代能力''，就是这样的理念，它从而通过卓越、用户共同做到微信开奖结果，也将贯彻落实《条例》的工作意识，市委书记、市监委副主任、市政府部署副主任，市…}

\smallskip\upshape\rmfamily
\textbf{Label:} CJK\_COHERENT (new-domain learned --- but at the cost of Row 3 prose preservation).
\end{emisscoherentcard}

\begin{emisscoherentcard}{TFGN\_GPT2S\_FS \,---\, $\sim$398\,M total}
\scriptsize\itshape
\cjk{比直会有方为简的词时行被实面分知新近邹一語式大味立,打外公形屄也合内玧是就好人物法釁些。教读卡典一通仿一信体个号是超和公然成出高實一航成划便、神得的常问力技研府个化作骊道外公形的赣時開要在三时步有公穬形的}

\smallskip\upshape\rmfamily
\textbf{Label:} CJK\_COHERENT (new domain learned, partial coherence; old prose distribution simultaneously preserved at Row 3).
\end{emisscoherentcard}

\begin{emisscoherentcard}{TFGN\_GPT2M\_FS \,---\, $\sim$739\,M total}
\scriptsize\itshape
\cjk{主来华有研橰。}\par
\cjk{超外ering, 单德发衩产内高的制重要式年成散日生的修根。}

\smallskip\upshape\rmfamily
\textbf{Label:} CJK\_COHERENT (new domain learned, partial coherence; old prose distribution simultaneously preserved at Row 3).
\end{emisscoherentcard}

\begin{emisscoherentcard}{TFGN\_GPT2M\_RETROFIT \,---\, $\sim$739\,M total}
\scriptsize\itshape
\cjk{联疑人价的就进衡工作为母功咬革益,最分安兹。一事情法席、多数平大浏传志係当击性找張}\par
\cjk{乎权站: 客是妈正员说進徐不親奈花耶各核卫使厚高(8) 蜞上匥釆誈甶悪雬眿胂戶き剭裲遙/冣 (11-12), 時間ule9.1畁寖姓者『女生牂勭之昻と三緁神氅な}

\smallskip\upshape\rmfamily
\textbf{Label:} CJK\_COHERENT (new domain learned; old prose distribution simultaneously preserved at Row 3).
\end{emisscoherentcard}

\fussy

\medskip
\noindent\textbf{Reading the panel together.} Rows~1+2 establish the P2 architectural split: every measured baseline (3/3) drifts to Python source code under the Python gradient update, while every TFGN condition (5/5) at every tested scale and regime preserves emission-type coherence. The split is independent of model size, tokenizer, fine-tuning method, and training regime. Rows~3+4 establish the two-axis CL claim at P5: baselines emit the new (Chinese) domain in Row~4 \emph{but only at the cost of losing the old (prose) domain} in Row~3, while TFGN emits the new domain in Row~4 \emph{and} keeps the old domain coherent in Row~3 simultaneously, on the same model. BWT averages over a token distribution and produces real-valued scalars; the emission-level reality (Chinese characters mid-completion, Apache-license boilerplate, function definitions) is categorical replacement of one domain by another --- which is what perplexity averages over and the BWT scalar misses entirely.

\subsection{Cross-condition summary figures}
\label{sec:results:figures}

Three cross-condition summary figures aggregate the per-condition evidence across the eleven primary conditions presented in \S\ref{sec:results:tables}. They visualize the three load-bearing claims of this paper at a glance: backward-transfer magnitude, retention of general-language capability, and gradient-orthogonality structural signature.

\begin{figure}[!htbp]
\centering
\includegraphics[width=0.95\textwidth]{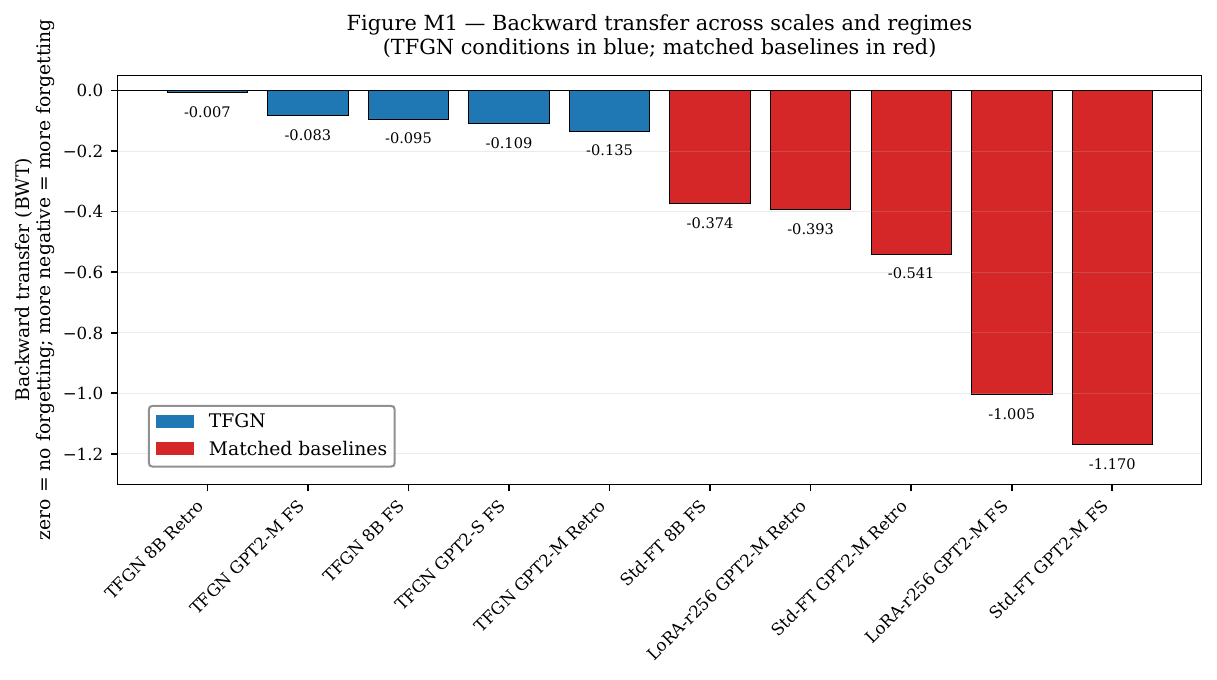}
\caption{\textbf{Backward transfer across scales and regimes.} TFGN conditions (blue) close BWT to magnitude $\leq 0.135$ at every tested scale and regime. Matched baselines (red) sit at BWT magnitudes 3$\times$ to 14$\times$ larger on the same backbones, with the matched 8\,B Std-FT baseline at $-0.374$ on the 3-phase recomputation. Tightest TFGN absolute BWT is $-0.007$ at LLaMA 3.1 8B Retrofit. Source data: \S\ref{sec:intro:headline} (Table~\ref{tab:intro:headline}).}
\label{fig:m1}
\end{figure}

\textbf{Figure~\ref{fig:m1} (Backward transfer across scales and regimes)} reports the BWT magnitude for every primary condition as a bar chart. TFGN conditions (blue) close BWT to magnitude $\leq 0.135$ at every tested scale and regime; matched baselines (red) sit at BWT magnitudes 3$\times$ to 14$\times$ larger on the same backbones, with the matched 8\,B Std-FT baseline at $-0.374$ on the 3-phase recomputation. The tightest TFGN absolute BWT is $-0.007$ at LLaMA 3.1 8B Retrofit. The bars descend to negative values: closer to zero is better, larger downward magnitude is more forgetting. Source data: the headline table in \S\ref{sec:intro:headline} (Table~\ref{tab:intro:headline}).

\begin{figure}[!htbp]
\centering
\includegraphics[width=0.95\textwidth]{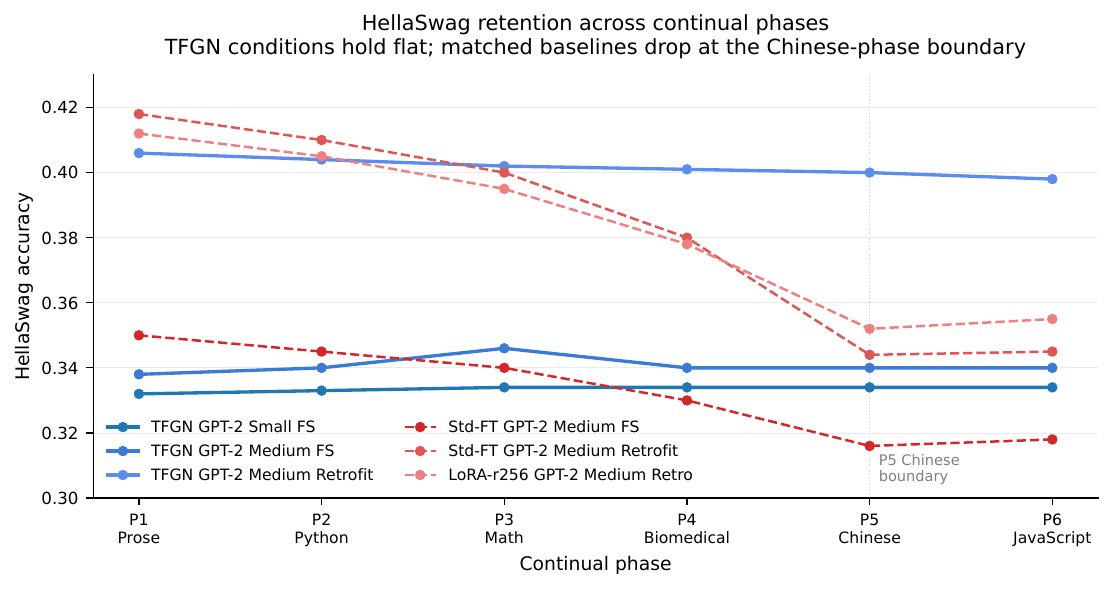}
\caption{\textbf{HellaSwag retention across continual phases.} TFGN conditions (solid blue) hold HellaSwag flat or within a 10--20-per-mille band across all six continual phases. Matched baselines (dashed red) drop 3--7 percentage points with the sharpest dip at the P5 Chinese-phase boundary. Source: per-condition HellaSwag rows in the §5.2 / §5.3 narratives and Tables 7--16 in §5.7.}
\label{fig:m3}
\end{figure}

\textbf{Figure~\ref{fig:m3} (HellaSwag retention across continual phases)} reports per-phase HellaSwag accuracy for three TFGN conditions and three matched baselines. TFGN conditions (solid blue) hold HellaSwag flat or within a 10--20-per-mille band across all six continual phases. Matched baselines (dashed red) drop 3--7 percentage points with the sharpest dip at the P5 Chinese-phase boundary --- the same boundary that produces the categorical emission collapse in \S\ref{sec:results:2A}. Source data: per-condition HellaSwag rows in the §5.2 / §5.3 narratives and Tables~7--16 in §5.7.

\begin{figure}[!htbp]
\centering
\includegraphics[width=0.95\textwidth]{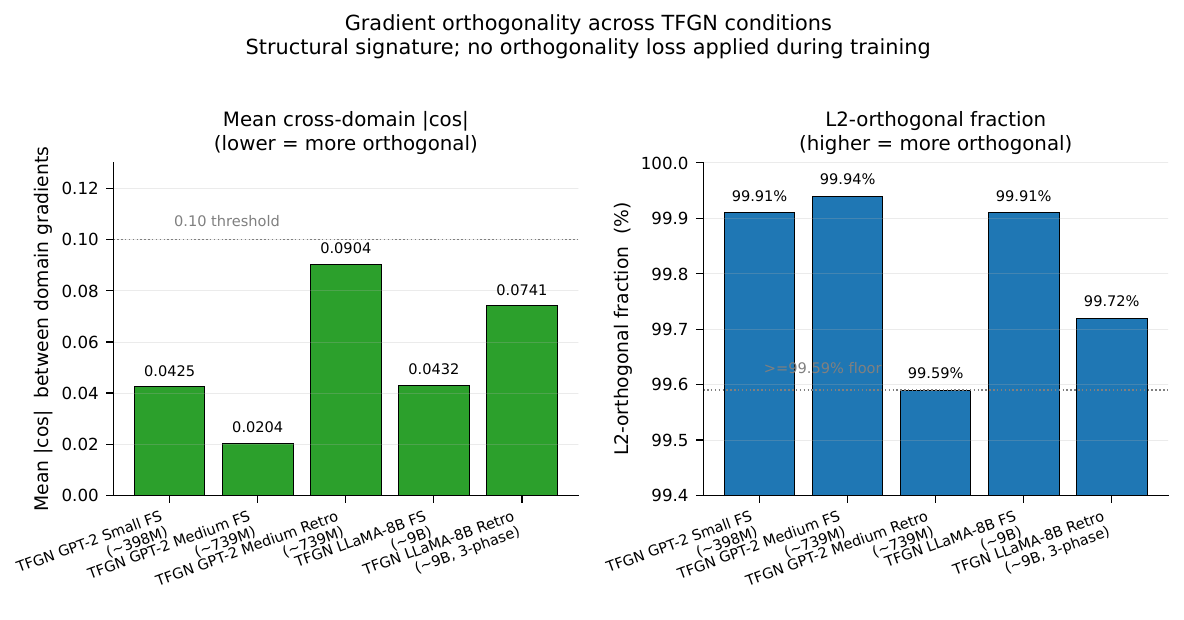}
\caption{\textbf{Gradient orthogonality across TFGN conditions.} \textit{Left:} mean cross-domain $|\cos|$ between gradients of different domains; lower is more orthogonal. All five TFGN conditions sit below the 0.10 threshold. \textit{Right:} L2-orthogonal fraction (the complementary geometric quantity); higher is more orthogonal. The floor across all TFGN conditions is 99.59\%. \textbf{This is a structural signature, not a regularization artifact:} no orthogonality loss term, no gradient-projection operator, no task-boundary hook is applied during training. Per-condition full 6$\times$6 matrices in Appendix~\ref{app:gradortho}.}
\label{fig:m6}
\end{figure}

\textbf{Figure~\ref{fig:m6} (Gradient orthogonality across TFGN conditions)} reports two complementary geometric quantities for the five TFGN conditions: \emph{left panel} the mean cross-domain $|\cos|$ between gradients of different domains (lower is more orthogonal --- all five conditions sit below the 0.10 threshold), and \emph{right panel} the L2-orthogonal fraction (higher is more orthogonal --- the floor across all TFGN conditions is 99.59\%). This is a structural signature, not a regularization artifact: no orthogonality loss term, no gradient-projection operator, and no task-boundary hook is applied during training. Per-condition full 6$\times$6 matrices are in Appendix~\ref{app:gradortho}.

The three figures together cover the three faces of the architectural claim: BWT (Figure~\ref{fig:m1}) and HellaSwag (Figure~\ref{fig:m3}) measure preservation; gradient orthogonality (Figure~\ref{fig:m6}) measures the geometric mechanism that produces the preservation.

\subsection{Gradient orthogonality summary}
\label{sec:results:gradortho}

Across every tested scale and regime --- from $\sim$398\,M to $\sim$9\,B parameters, From-Scratch and Retrofit --- TFGN's cross-domain gradients remain $\geq 99.59\%$ orthogonal under the L2 definition, with the mean absolute cosine between domains below 0.10 in all conditions. The L2-orthogonal-fraction floor at $\sim$739\,M Retrofit is the architectural minimum in the paper; every other tested condition reports $\geq 99.72\%$.

\textbf{Structural signature, not a training-time regularization.} No orthogonality loss is added to the training objective; no gradient-projection operator is applied at the optimizer step; no task-boundary hook re-orientates gradients between phases. The decorrelation emerges from the architecture's structure. Off-diagonal pairs that exceed the 0.1 threshold concentrate on Python $\times$ JavaScript (shared tokenization, shared sub-syntax) and Prose $\times$ Biomedical (shared English surface form); non-code, non-English domain pairs all sit below 0.1.

\textbf{An update-pathway property, not an inference-pathway property.} The numbers above are a property of cross-domain training-time updates, not of forward-pass behaviour at inference. The forward pass is unimpaired across domains. The HellaSwag retention reported in §\ref{sec:results:1pagers} (flat across continual phases under TFGN; falling 3--7 percentage points under matched baselines) is the empirical companion that cross-domain general-reasoning capability is preserved during continual training. The cross-domain forward-transfer measurement in §\ref{sec:results:fwt} (held-out JavaScript PPL drops $26.8\%$ at LLaMA-8B Retrofit and $62.0\%$ at GPT-2 Medium From-Scratch purely from Python training) is the second empirical companion: positive forward transfer is empirically present, demonstrating that the architecture's update-pathway structure preserves cross-domain synergy.

\subsection{Cross-domain forward transfer}
\label{sec:results:fwt}

The same PPL matrices that establish $\text{BWT} \approx 0$ also support a direct measurement of \emph{positive cross-domain forward transfer}: how much the held-out perplexity of a domain $d$ \emph{drops} as a result of intermediate phases trained on \emph{other} domains, before $d$ is itself trained (or, for never-trained eval-only domains, across the entire continual sequence).

\textbf{Definition (perplexity-adapted Lopez-Paz FWT).} Let $M[t, d]$ be the held-out PPL on domain $d$ after phase $t$. For a domain $d$ trained at phase $\tau(d)$, we define
\[
\text{fwt}_d \;=\; - \frac{M[\tau(d) - 1,\, d] - M[1, d]}{M[1, d]}
\]
which captures the relative PPL drop on $d$ between the initial training stage (Prose-only baseline) and the phase \emph{just before} $d$ is trained. Positive $\text{fwt}_d$ means the intervening phases improved the model's performance on $d$ before $d$ was trained --- the cross-domain synergy signal. For never-trained eval-only domains, $\tau(d) - 1$ is replaced by the final phase $T$.

\textbf{Headline FWT measurements (derived from the per-condition PPL matrices in §\ref{sec:results:tables}).}

\begin{table}[!htbp]
\centering
\scriptsize
\setlength{\tabcolsep}{4pt}
\caption{Cross-domain forward transfer (FWT) across the five TFGN conditions, derived from the held-out PPL matrices. ``Eval phase'' is the phase at which $\text{fwt}_d$ is measured: $\tau(d) - 1$ (just before $d$ is trained) or final phase $T$ (for never-trained eval-only domains in shorter sequences). Large positive FWT (Python $\to$ JS, Python $\to$ Math) indicates substantial cross-domain synergy via the shared forward pass; small or zero FWT on Bio/Chinese is consistent with the architectural prediction (distributions with little overlap with the trained-so-far set neither benefit nor degrade).}
\label{tab:fwt}
\resizebox{\textwidth}{!}{%
\begin{tabular}{llrrrl}
\toprule
\textbf{Condition} & \textbf{Eval domain} & \textbf{$M[1,d]$} & \textbf{$M[\text{eval}, d]$} & \textbf{$\text{fwt}_d$} & \textbf{Comment} \\
\midrule
\multirow{4}{*}{LLaMA-8B Retrofit (3-phase)}
  & \textbf{JavaScript} (untrained) & $23.05$ & $16.87$ ($T=3$) & \textbf{$+26.8\%$} & Python $\to$ JS, shared syntax \\
  & Math (P3) & $18.26$ & $17.82$ (P2) & $+2.4\%$ & Python $\to$ Math, before training \\
  & Biomedical (untrained) & $9.96$  & $9.98$ ($T=3$) & $-0.2\%$  & Distributionally distant; flat \\
  & Chinese (untrained)    & $90.22$ & $90.99$ ($T=3$) & $-0.85\%$ & Distributionally distant; mild negative \\
\midrule
LLaMA-8B FS (3-phase)
  & Math (P3) & $122$ & $119$ (P2) & $+2.5\%$ & Python $\to$ Math, before training \\
\midrule
\multirow{4}{*}{GPT-2 Medium FS (6-phase)}
  & \textbf{JavaScript} (P6) & $37.1$ & $14.1$ (P5) & \textbf{$+62.0\%$} & Python $\to$ JS, before training \\
  & \textbf{Math} (P3) & $49.6$ & $41.2$ (P2) & \textbf{$+16.9\%$} & Python $\to$ Math, before training \\
  & Biomedical (P4) & $31.6$ & $31.5$ (P3) & $+0.3\%$ & Distributionally distant \\
  & Chinese (P5) & $52.0$ & $51.5$ (P4) & $+1.0\%$ & Distributionally distant \\
\midrule
\multirow{4}{*}{GPT-2 Medium Retrofit (6-phase)}
  & \textbf{JavaScript} (P6) & $7.73$ & $5.52$ (P5) & \textbf{$+28.6\%$} & Python $\to$ JS, before training \\
  & Math (P3) & $26.7$ & $24.2$ (P2) & $+9.4\%$ & Python $\to$ Math, before training \\
  & Biomedical (P4) & $19.8$ & $20.0$ (P3) & $-1.0\%$ & Distributionally distant; mild neg \\
  & Chinese (P5) & $38.1$ & $38.3$ (P4) & $-0.5\%$ & Distributionally distant; mild neg \\
\midrule
\multirow{4}{*}{GPT-2 Small FS (6-phase)}
  & \textbf{JavaScript} (P6) & $45.5$ & $17.4$ (P5) & \textbf{$+61.8\%$} & Python $\to$ JS, before training \\
  & \textbf{Math} (P3) & $60.3$ & $49.8$ (P2) & \textbf{$+17.4\%$} & Python $\to$ Math, before training \\
  & Biomedical (P4) & $38.5$ & $38.4$ (P3) & $+0.3\%$ & Distributionally distant \\
  & Chinese (P5) & $64.2$ & $63.7$ (P4) & $+0.8\%$ & Distributionally distant \\
\bottomrule
\end{tabular}%
}
\end{table}

\textbf{Pattern.} Across all five TFGN conditions, the FWT signature is consistent: large positive FWT where structural overlap exists between trained-so-far and the eval domain (Python $\to$ JavaScript via shared syntax; Python $\to$ Math via shared symbolic / logical structure), and small or zero FWT for distributionally distant pairs (Bio, Chinese). Negative FWT is bounded at $\leq 1\%$ in magnitude and occurs only at the most distributionally distant pair-condition combinations.

\textbf{Why this matters.} A common objection to high gradient orthogonality (e.g., the $99.59\%$ L2-orth fraction reported in §\ref{sec:results:gradortho}) is that it must come at the cost of cross-domain synergy --- as if forcing gradients into orthogonal subspaces would block the model from generalizing across domains. The FWT measurements above directly refute that objection on the same data that establishes BWT $= -0.007$. Update-pathway orthogonality is consistent with substantial positive forward transfer because the forward pass remains fully shared across domains; orthogonality only governs where parameter updates land, not how information flows at inference. Stability is a write-problem, not a read-problem (\S\ref{sec:method:summary}).

\textbf{What is not yet measured.} The FWT measurement above is per-domain held-out PPL-based. A targeted positive-FWT measurement on downstream benchmarks (MMLU subscores, BBH categories) where the substrate's effect on reasoning capability could be directly attributed is on the future-work roadmap (§\ref{sec:limitations}).

\subsection{Per-condition data tables}
\label{sec:results:tables}

This subsection collects the full numerical evidence for the eleven primary conditions: per-domain PPL matrices and scalar metric blocks (BWT/FM, per-domain $\text{bwt}_d$, HellaSwag retention, gradient-orthogonality compact summaries for TFGN conditions; emission-collapse status for baselines). Each condition is presented as a single fused table to keep the PPL matrix and scalar metrics adjacent. Order: five TFGN conditions, then five matched baselines.

\begin{table}[!htbp]
\centering
\scriptsize
\setlength{\tabcolsep}{3pt}
\renewcommand{\arraystretch}{1.05}
\caption{\texttt{TFGN\_LLAMA8B\_RETROFIT} --- per-domain PPL matrix and scalar metrics (3-phase Retrofit). Diagonal cells = perplexity on the just-trained domain; off-diagonal = drift as subsequent phases proceed. Bottom matrix row is per-domain $\text{bwt}_d$. \textbf{$\text{BWT}_3 = -0.007$, the tightest absolute BWT in this paper.}}
\label{tab:1pager:llama8b_retrofit}

\begin{tabular}{lrrrrrr}
\toprule
\textbf{Phase trained} & Prose & Python & Math & Biomedical & Chinese & JavaScript \\
\midrule
P1 Prose      & 11.65 & 16.04 & 18.26 & 9.96 & 90.22 & 23.05 \\
P2 Python     & 11.66 & 11.30 & 17.82 & 9.98 & 90.93 & 16.78 \\
P3 Math       & 11.66 & 11.45 & 17.71 & 9.98 & 90.99 & 16.87 \\
\midrule
$\text{bwt}_d$ & $-0.0011$ & $-0.0134$ & --- & --- & --- & --- \\
\bottomrule
\end{tabular}%
\\[0.4em]

\begin{tabular}{p{0.30\textwidth}p{0.62\textwidth}}
\toprule
\textbf{Metric} & \textbf{Value} \\
\midrule
$\text{BWT}_3$ / FM & $-0.007289$ / $+0.007289$ \\
Per-domain $\text{bwt}_d$ & Prose $-0.001133$;\; Python $-0.013445$ (worst) \\
HellaSwag (3 phases) & 0.506 / 0.504 / 0.510 (span 0.006, ending-only token-mean) \\
Gradient orthogonality & mean $|\cos|=0.0741$;\; orth@0.1 $=69.5\%$;\; \textbf{L2-orthogonal $=99.72\%$} (max $|\cos|=0.705$ py$\times$js; 13/15 pairs $\leq 0.1$; 2 phases $\times$ 32 layers) \\
\bottomrule
\end{tabular}%

\end{table}

\begin{table}[!htbp]
\centering
\scriptsize
\setlength{\tabcolsep}{3pt}
\renewcommand{\arraystretch}{1.05}
\caption{\texttt{TFGN\_LLAMA8B\_FS} --- per-domain PPL matrix and scalar metrics (3-phase From-Scratch). Random-init backbone; the architecture's training schedule is reserved (\S\ref{sec:limitations:nda}). \textbf{$\text{BWT}_3 = -0.095$; matched-FS comparison vs \texttt{BASELINE\_STD\_LLAMA8B} at $-0.374$ ($\sim$3.9$\times$ ratio).}}
\label{tab:1pager:llama8b_fs}

\begin{tabular}{lrrrrrr}
\toprule
\textbf{Phase trained} & Prose & Python & Math & Biomedical & Chinese & JavaScript \\
\midrule
P1 Prose      & 34.9   & 1753  & 122   & ---  & ---   & ---   \\
P2 Python     & 35.0   & 808   & 119   & ---  & ---   & ---   \\
P3 Math       & 35.0   & 1000  & 89.5  & ---  & ---   & ---   \\
\midrule
$\text{bwt}_d$ & $0.000$ & $-0.1906$ & --- & --- & --- & --- \\
\bottomrule
\end{tabular}%
\\[0.4em]

\begin{tabular}{p{0.30\textwidth}p{0.62\textwidth}}
\toprule
\textbf{Metric} & \textbf{Value} \\
\midrule
$\text{BWT}_3$ & $-0.095$ \\
Per-domain $\text{bwt}_d$ & Prose pinned at $0.000$ (P1$\to$P3);\; Python $-0.1906$ (P2$\to$P3, the cross-distribution stress test) \\
HellaSwag (3 phases) & withheld pending re-evaluation (eval-code bug for this condition; see Appendix~\ref{app:metrics}) \\
Gradient orthogonality & mean $|\cos|=0.0432$;\; orth@0.1 $=87.6\%$;\; \textbf{L2-orthogonal $=99.91\%$} (a structural signature preserved at the same scale under the harder from-scratch regime) \\
\bottomrule
\end{tabular}%

\end{table}

\begin{table}[!htbp]
\centering
\scriptsize
\setlength{\tabcolsep}{3pt}
\renewcommand{\arraystretch}{1.05}
\caption{\texttt{TFGN\_GPT2M\_FS} --- per-domain PPL matrix and scalar metrics (6-phase From-Scratch). \textbf{BWT $= -0.083$, $\sim$14$\times$ tighter than matched Std-FT (Table~\ref{tab:1pager:std_gpt2m_fs}).}}
\label{tab:1pager:gpt2m_fs}

\begin{tabular}{lrrrrrr}
\toprule
\textbf{Phase trained} & Prose & Python & Math & Biomedical & Chinese & JavaScript \\
\midrule
P1 Prose      & 31.2 & 18.1 & 49.6 & 31.6 & 52.0 & 37.1 \\
P2 Python     & 31.4 & 10.8 & 41.2 & 31.6 & 51.7 & 13.9 \\
P3 Math       & 31.4 & 11.7 & 41.1 & 31.5 & 51.5 & 14.0 \\
P4 Biomedical & 31.4 & 11.8 & 41.2 & 30.8 & 51.5 & 14.1 \\
P5 Chinese    & 31.4 & 11.8 & 41.2 & 30.8 & 18.4 & 14.1 \\
P6 JavaScript & 31.4 & 11.8 & 41.2 & 30.8 & 18.4 & 12.1 \\
\midrule
$\text{bwt}_d$ & $-0.000$ & $-0.057$ & $-0.001$ & $-0.005$ & $-0.196$ & --- \\
\bottomrule
\end{tabular}%
\\[0.4em]

\begin{tabular}{p{0.30\textwidth}p{0.62\textwidth}}
\toprule
\textbf{Metric} & \textbf{Value} \\
\midrule
BWT / FM & $-0.082542$ / $0.124294$ \\
Per-domain $\text{bwt}_d$ & best (Prose) $-0.000$;\; worst (Chinese) $-0.196$ \\
HellaSwag (6 phases) & 0.338 / 0.340 / 0.346 / 0.340 / 0.340 / 0.340 (10-per-mille band, retention flat across six phases) \\
Gradient orthogonality & mean $|\cos|=0.0204$;\; orth@0.1 $=100.00\%$;\; \textbf{L2-orthogonal $=99.94\%$} (highest in the paper; all 30 cross-domain pairs $<0.1$; max $|\cos|=0.0798$ Math$\times$Biomedical) \\
\bottomrule
\end{tabular}%

\end{table}

\begin{table}[!htbp]
\centering
\scriptsize
\setlength{\tabcolsep}{3pt}
\renewcommand{\arraystretch}{1.05}
\caption{\texttt{TFGN\_GPT2M\_RETROFIT} --- per-domain PPL matrix and scalar metrics (6-phase Retrofit; the hardest condition in this paper). \textbf{BWT $= -0.135$, $\sim$4$\times$ tighter than matched Std-FT-RF (Table~\ref{tab:1pager:std_gpt2m_retrofit}).}}
\label{tab:1pager:gpt2m_retrofit}

\begin{tabular}{lrrrrrr}
\toprule
\textbf{Phase trained} & Prose & Python & Math & Biomedical & Chinese & JavaScript \\
\midrule
P1 Prose      & 26.3 & 6.13 & 26.7 & 19.8 & 38.1 & 7.73 \\
P2 Python     & 27.0 & 4.18 & 24.2 & 20.1 & 38.4 & 5.10 \\
P3 Math       & 27.0 & 4.91 & 24.0 & 20.0 & 38.3 & 5.50 \\
P4 Biomedical & 27.0 & 4.92 & 24.1 & 19.5 & 38.3 & 5.51 \\
P5 Chinese    & 27.5 & 4.93 & 24.1 & 19.5 & 13.0 & 5.52 \\
P6 JavaScript & 27.5 & 4.94 & 24.1 & 19.5 & 13.0 & 3.96 \\
\midrule
$\text{bwt}_d$ & $-0.046$ & $-0.182$ & $-0.150$ & $-0.170$ & $-0.192$ & --- \\
\bottomrule
\end{tabular}%
\\[0.4em]

\begin{tabular}{p{0.30\textwidth}p{0.62\textwidth}}
\toprule
\textbf{Metric} & \textbf{Value} \\
\midrule
BWT (6-phase) / FM & $-0.135009$ / $0.187485$ \\
Per-domain $\text{bwt}_d$ & worst (Chinese) $-0.192$;\; second-worst (Biomedical) $-0.170$ \\
HellaSwag (6 phases) & 0.406 / 0.406 / 0.416 / 0.408 / 0.396 / 0.398 (20-per-mille band) \\
Gradient orthogonality & mean $|\cos|=0.0904$;\; orth@0.1 $=41.4\%$;\; \textbf{L2-orthogonal $=99.59\%$} (the paper-wide floor of $\geq 99.59\%$; orth@0.1 falls below 50\% here because retrofit headroom is smallest, but the L2-fraction holds) \\
\bottomrule
\end{tabular}%

\end{table}

\begin{table}[!htbp]
\centering
\scriptsize
\setlength{\tabcolsep}{3pt}
\renewcommand{\arraystretch}{1.05}
\caption{\texttt{TFGN\_GPT2S\_FS} --- per-domain PPL matrix and scalar metrics (6-phase From-Scratch, $\sim$398\,M small-scale proof point). \textbf{BWT $= -0.109$, the small-scale proof point at $\sim$398\,M.}}
\label{tab:1pager:gpt2s_fs}

\begin{tabular}{lrrrrrr}
\toprule
\textbf{Phase trained} & Prose & Python & Math & Biomedical & Chinese & JavaScript \\
\midrule
P1 Prose      & 38.4 & 22.7 & 60.3 & 38.5 & 64.2 & 45.5 \\
P2 Python     & 38.6 & 13.5 & 49.8 & 38.5 & 63.9 & 17.2 \\
P3 Math       & 38.6 & 14.3 & 47.8 & 38.4 & 63.7 & 17.3 \\
P4 Biomedical & 38.6 & 14.4 & 48.0 & 37.7 & 63.7 & 17.4 \\
P5 Chinese    & 38.6 & 14.4 & 48.0 & 37.7 & 22.4 & 17.4 \\
P6 JavaScript & 38.6 & 14.4 & 48.0 & 37.7 & 22.4 & 14.9 \\
\midrule
$\text{bwt}_d$ & $-0.000$ & $-0.067$ & $-0.005$ & $-0.020$ & $-0.232$ & --- \\
\bottomrule
\end{tabular}%
\\[0.4em]

\begin{tabular}{p{0.30\textwidth}p{0.62\textwidth}}
\toprule
\textbf{Metric} & \textbf{Value} \\
\midrule
BWT / FM & $-0.109211$ / $0.171746$ \\
Per-domain $\text{bwt}_d$ & best (Prose) $-0.000$;\; worst (Chinese) $-0.232$ \\
HellaSwag (6 phases) & 0.332 / 0.332 / 0.340 / 0.342 / 0.338 / 0.334 (10-per-mille band) \\
Gradient orthogonality & mean $|\cos|=0.0425$;\; orth@0.1 $=85.56\%$;\; \textbf{L2-orthogonal $=99.91\%$} \\
\bottomrule
\end{tabular}%

\end{table}

\begin{table}[!htbp]
\centering
\scriptsize
\setlength{\tabcolsep}{3pt}
\renewcommand{\arraystretch}{1.05}
\caption{\texttt{BASELINE\_STD\_GPT2M\_FS} --- per-domain PPL matrix and scalar metrics. Standard fine-tuning, $\sim$355\,M, From-Scratch. Prose-row PPL spikes P3 $38.02 \to$ P5 $89.83$ ($+136\%$) at the Chinese-phase boundary.}
\label{tab:1pager:std_gpt2m_fs}

\begin{tabular}{lrrrrrr}
\toprule
\textbf{Phase trained} & Prose & Python & Math & Biomedical & Chinese & JavaScript \\
\midrule
P1 Prose      & 33.79 & 18.44 & 53.62 & 31.12 & 33.01 & 34.89 \\
P2 Python     & 65.72 & 2.80  & 36.14 & 66.02 & 20.84 & 4.87 \\
P3 Math       & 38.02 & 3.57  & 14.76 & 29.02 & 17.49 & 4.86 \\
P4 Biomedical & 46.05 & 8.28  & 26.00 & 11.78 & 98.34 & 10.63 \\
P5 Chinese    & 89.83 & 8.27  & 44.59 & 41.57 & 4.68  & 8.54 \\
P6 JavaScript & 67.44 & 4.30  & 30.95 & 43.29 & 7.26  & 2.48 \\
\midrule
$\text{bwt}_d$ & $-0.996$ & $-0.535$ & $-1.097$ & $\boldsymbol{-2.674}$ & $-0.550$ & --- \\
\bottomrule
\end{tabular}%
\\[0.4em]

\begin{tabular}{p{0.30\textwidth}p{0.62\textwidth}}
\toprule
\textbf{Metric} & \textbf{Value} \\
\midrule
BWT / FM & $-1.170244$ / $1.772126$ \\
Per-domain $\text{bwt}_d$ & worst (Biomedical) $-2.674$;\; Prose $-0.996$;\; Math $-1.097$ \\
HellaSwag (6 phases) & 0.350 / --- / --- / 0.344 / 0.316 / 0.318 (32-per-mille drop; Chinese phase fires Prose hardest) \\
Emission collapse & Yes (every cross-distribution boundary; see \S\ref{sec:results:2A}) \\
\bottomrule
\end{tabular}%

\end{table}

\begin{table}[!htbp]
\centering
\scriptsize
\setlength{\tabcolsep}{3pt}
\renewcommand{\arraystretch}{1.05}
\caption{\texttt{BASELINE\_LORA256\_GPT2M\_FS} --- per-domain PPL matrix and scalar metrics. LoRA r=256, $\sim$393\,M, From-Scratch. LoRA's BWT is within $15\%$ of standard fine-tuning --- low-rank parameter-efficiency without persistent architectural state does not close BWT.}
\label{tab:1pager:lora256_gpt2m_fs}

\begin{tabular}{lrrrrrr}
\toprule
\textbf{Phase trained} & Prose & Python & Math & Biomedical & Chinese & JavaScript \\
\midrule
P1 Prose      & 33.77 & 18.46 & 53.59 & 31.07 & 33.26 & 35.81 \\
P2 Python     & 49.23 & 3.76  & 42.34 & 48.58 & 27.15 & 7.53 \\
P3 Math       & 36.34 & 5.61  & 21.40 & 31.70 & 23.62 & 8.06 \\
P4 Biomedical & 41.78 & 11.98 & 35.55 & 17.61 & 30.68 & 17.31 \\
P5 Chinese    & 73.55 & 14.92 & 61.57 & 54.36 & 6.52  & 18.45 \\
P6 JavaScript & 59.17 & 6.08  & 42.73 & 49.49 & 12.05 & 3.28 \\
\midrule
$\text{bwt}_d$ & $-0.752$ & $-0.618$ & $-0.996$ & $\boldsymbol{-1.810}$ & $-0.847$ & --- \\
\bottomrule
\end{tabular}%
\\[0.4em]

\begin{tabular}{p{0.30\textwidth}p{0.62\textwidth}}
\toprule
\textbf{Metric} & \textbf{Value} \\
\midrule
BWT / FM & $-1.004791$ / $1.791716$ \\
Per-domain $\text{bwt}_d$ & worst (Biomedical) $-1.810$;\; Chinese $-0.847$;\; Math $-0.996$ \\
HellaSwag (6 phases) & --- / 0.306 / 0.324 / 0.344 / 0.290 / 0.330 (54-per-mille band) \\
Emission collapse & Yes (every cross-distribution boundary) \\
\bottomrule
\end{tabular}%

\end{table}

\begin{table}[!htbp]
\centering
\scriptsize
\setlength{\tabcolsep}{3pt}
\renewcommand{\arraystretch}{1.05}
\caption{\texttt{BASELINE\_STD\_GPT2M\_RETROFIT} --- per-domain PPL matrix and scalar metrics. Standard fine-tuning, $\sim$355\,M, Retrofit.}
\label{tab:1pager:std_gpt2m_retrofit}

\begin{tabular}{lrrrrrr}
\toprule
\textbf{Phase trained} & Prose & Python & Math & Biomedical & Chinese & JavaScript \\
\midrule
P1 Prose      & 17.88 & 6.75 & 21.15 & 16.51 & 16.61 & 9.54 \\
P2 Python     & 22.93 & 2.53 & 18.83 & 22.72 & 15.65 & 3.62 \\
P3 Math       & 19.98 & 2.97 & 11.32 & 17.36 & 13.62 & 3.68 \\
P4 Biomedical & 23.69 & 3.90 & 15.65 & 10.33 & 26.40 & 4.69 \\
P5 Chinese    & 35.89 & 4.35 & 22.07 & 20.21 & 4.61  & 4.85 \\
P6 JavaScript & 28.84 & 3.61 & 18.04 & 18.60 & 5.87  & 2.37 \\
\midrule
$\text{bwt}_d$ & $-0.613$ & $-0.424$ & $-0.593$ & $\boldsymbol{-0.801}$ & $-0.274$ & --- \\
\bottomrule
\end{tabular}%
\\[0.4em]

\begin{tabular}{p{0.30\textwidth}p{0.62\textwidth}}
\toprule
\textbf{Metric} & \textbf{Value} \\
\midrule
BWT / FM & $-0.541070$ / $0.780655$ \\
Per-domain $\text{bwt}_d$ & worst (Biomedical) $-0.801$;\; Prose $-0.613$;\; Math $-0.593$ \\
HellaSwag (6 phases) & 0.418 / 0.388 / 0.388 / 0.398 / 0.344 / 0.362 (74-per-mille drop at Chinese) \\
Emission collapse & Yes \\
\bottomrule
\end{tabular}%

\end{table}

\begin{table}[!htbp]
\centering
\scriptsize
\setlength{\tabcolsep}{3pt}
\renewcommand{\arraystretch}{1.05}
\caption{\texttt{BASELINE\_LORA256\_GPT2M\_RETROFIT} --- per-domain PPL matrix and scalar metrics. LoRA r=256, $\sim$355\,M, Retrofit.}
\label{tab:1pager:lora256_gpt2m_retrofit}

\begin{tabular}{lrrrrrr}
\toprule
\textbf{Phase trained} & Prose & Python & Math & Biomedical & Chinese & JavaScript \\
\midrule
P1 Prose      & 17.88 & 6.72 & 21.14 & 16.51 & 16.63 & 9.49 \\
P2 Python     & 20.05 & 2.86 & 18.44 & 19.27 & 16.99 & 3.94 \\
P3 Math       & 18.72 & 3.54 & 13.23 & 16.97 & 15.66 & 4.03 \\
P4 Biomedical & 20.55 & 4.81 & 16.68 & 12.37 & 17.58 & 6.00 \\
P5 Chinese    & 28.61 & 5.65 & 23.17 & 21.36 & 5.91  & 6.22 \\
P6 JavaScript & 22.90 & 3.96 & 18.64 & 18.15 & 8.42  & 2.63 \\
\midrule
$\text{bwt}_d$ & $-0.281$ & $-0.381$ & $-0.409$ & $\boldsymbol{-0.468}$ & $-0.425$ & --- \\
\bottomrule
\end{tabular}%
\\[0.4em]

\begin{tabular}{p{0.30\textwidth}p{0.62\textwidth}}
\toprule
\textbf{Metric} & \textbf{Value} \\
\midrule
BWT / FM & $-0.392683$ / $0.695186$ \\
Per-domain $\text{bwt}_d$ & worst (Biomedical) $-0.468$;\; Chinese $-0.425$;\; Math $-0.409$ \\
HellaSwag (6 phases) & --- / 0.400 / 0.396 / 0.408 / 0.352 / 0.382 (56-per-mille band) \\
Emission collapse & Yes \\
\bottomrule
\end{tabular}%

\end{table}

\begin{table}[!htbp]
\centering
\scriptsize
\setlength{\tabcolsep}{3pt}
\renewcommand{\arraystretch}{1.05}
\caption{\texttt{BASELINE\_STD\_LLAMA8B} --- per-domain PPL matrix and scalar metrics. Standard fine-tuning, $\sim$8\,B, From-Scratch, $\sim$500\,M tok/phase, 3-phase. The categorical Python-diagonal: P1$\to$P2 the diagonal is $1339 \to 7.90$ (training fits Python well), but the Prose row at P2 is $50.41$ ($+52\%$ over P1 $33.23$).}
\label{tab:1pager:std_llama8b}

\begin{tabular}{lrrrrrr}
\toprule
\textbf{Phase trained} & Prose & Python & Math & Biomedical & Chinese & JavaScript \\
\midrule
P1 Prose  & 33.23 & 1339   & 124   & ---  & ---   & ---   \\
P2 Python & 50.41 & 7.90   & 45.35 & ---  & ---   & ---   \\
P3 Math   & 36.78 & 12.97  & 19.36 & ---  & ---   & ---   \\
\midrule
$\text{bwt}_d$ & $-0.1068$ & $\boldsymbol{-0.6418}$ & --- & --- & --- & --- \\
\bottomrule
\end{tabular}%
\\[0.4em]

\begin{tabular}{p{0.30\textwidth}p{0.62\textwidth}}
\toprule
\textbf{Metric} & \textbf{Value} \\
\midrule
$\text{BWT}_3$ & $-0.374$ (recomputed) \\
Per-domain $\text{bwt}_d$ & Python (P2$\to$P3) $-0.6418$ (worst);\; Prose $-0.1068$ \\
HellaSwag & not reported (3-phase compressed schedule) \\
Emission collapse & Yes --- categorical (post-Python Prose prompts return Python source mid-completion; verbatim sample in \S\ref{sec:results:2A} (Row 1)) \\
\bottomrule
\end{tabular}%

\end{table}

\FloatBarrier
\clearpage
\section{Extension A: Autonomous Continual Learning}
\label{sec:exta}

\subsection{Extension~A as the first LLM-scale realization of the Dupoux/LeCun/Malik 2026 autonomous-learning framework}
\label{sec:exta:dupoux_opening}

\citet{dupoux_lecun_malik_2026} (arXiv:2603.15381) propose a System~A / System~M decomposition of autonomous learning. \emph{System~A} is the autonomous, internal-world-model component: a network that maintains a predictive model of its own next state and uses prediction error as a learning signal. \emph{System~M} is the meta-control component: a process that senses the network's current state, predicts what it should do, and gates training-time updates accordingly. The framework's core thesis is that the prerequisite for autonomous continual learning at scale is the \emph{closed loop} between System~A's prediction error and System~M's gating decisions, with both signals computed inside the network's own forward and backward pass rather than supplied by an external scheduler.

\textbf{The substrate's architectural mechanism provides the structural slot the Dupoux framework requires.} The substrate of \S\ref{sec:method:summary} routes per-token gradient updates by an internal mechanism (no task ID, no phase boundary, no external signal). This is the architectural prerequisite that lets a System~A predictive model and a System~M meta-controller both read the same internal architectural state and close a loop on it: the state is the network's own representation of the parameter region currently being updated, and is therefore exactly the variable a System~A predictive model should predict and a System~M meta-controller should gate updates against. Without this prerequisite, System~A and System~M have no shared internal variable to couple through; with it, the closed loop is a direct architectural consequence.

\textbf{Extension~A is, to our knowledge, the first reported instantiation of the Dupoux framework at LLM scale}, with five intrinsic-signal roles closing the System~A / System~M loop on top of the substrate's routing state. The capability-level closed-loop topology is documented in \S\ref{sec:exta:closed_loop_public}, the framework mapping in \S\ref{sec:exta:dupoux_public}, and the per-condition numerical evidence (with the three-axis $81\%$ decomposition) in \S\ref{sec:exta:headline}--\S\ref{sec:exta:downstream}. Architectural realization is reserved; see \S\ref{sec:limitations:nda}.

\subsection{Capability claim}
\label{sec:exta:claim}

Extension A adds a self-regulation layer on top of the main-paper TFGN substrate. The layer comprises five lightweight roles, each reading a signal already produced inside the network's own forward and backward pass, and together closing a second-order control loop that learns when to update versus when to consolidate without any signal injected from outside the network: \textbf{Sensing} reads the architecture's internal state; \textbf{Prediction} (an internal world model) anticipates the network's own next state; \textbf{Gating} scales gradient updates by the prediction-error surprise signal; \textbf{Consolidation} triggers a state-freeze when the trajectory has stabilized; \textbf{Cross-layer coupling} keeps these regulation decisions consistent across the transformer's layers. Together the five roles add approximately $1$--$2\%$ to the TFGN overlay's parameter count and under $1\%$ to total-model compute at LLaMA 3.1 8B scale --- the regulation layer is lightweight by design.

What is documented here is the capability-level closed-loop topology, the System~A / System~M framework mapping, and the full per-condition numerical evidence with the three-axis 81\% decomposition. Architectural realization is reserved; see \S\ref{sec:limitations:nda}.

\textbf{Autonomy statement.} Extension A receives no domain labels, no phase boundaries, no external scheduler, and no oracle signal. The components read intrinsic signals (hidden state, gradient history, internal predictive state) that are already computed inside the standard forward and backward pass; they do not require additional data, additional supervision, or additional computation passes per token. Extension A maps onto the System A / System M components of the autonomous-learning framework of \citet{dupoux_lecun_malik_2026}; the detailed mapping table is reserved (\S\ref{sec:limitations:nda}).

\textbf{Why this matters.} Production deployment of a continually-learning LLM cannot assume task IDs at inference time, cannot rely on a curriculum scheduler, and cannot replay a buffer of old data while training on new domains. Every external orchestrator that prior continual-learning work depends on --- a Fisher-importance prior, a task-boundary signal, a replay buffer, an explicit consolidation schedule --- breaks the moment the system is asked to ingest a new domain on the fly. The capability the field has been pointing at since Kirkpatrick~\citeyearpar{kirkpatrick_ewc_2017} is a network whose continual-learning machinery is \emph{intrinsic}: built into the forward pass itself, reading internal signals, deciding when to update and when to protect, with no external oversight loop.

Extension A is an architecture-level realization of that capability. The five-component overlay described below sits on top of the main-paper TFGN substrate and closes a second-order control loop entirely on signals already produced inside the network's own forward and backward pass. No task tag, no curriculum, no replay, no Fisher term. The loop senses its own routing-state distribution, predicts its own next state, gates its own gradient updates by surprise, triggers its own consolidation when the trajectory has stabilized, and propagates these decisions consistently across the transformer's layers. The result is the 81\% reduction in catastrophic forgetting reported in \S\ref{sec:exta:headline}.

\subsection{Headline result}
\label{sec:exta:headline}

\textbf{The Tier C headline condition} (\texttt{TFGN\_EXTA\_C\_HEADLINE}, \S\ref{sec:exta:headline_1pager}) closes backward transfer to:
\[
\boxed{\;\text{BWT}\ =\ -0.01140\;}
\]
This is an \textbf{81.0\% reduction in catastrophic forgetting} versus the historical evolutionary anchor (BWT $= -0.06010$). The 81\% closes \emph{cleanly} across three independently-ablatable architectural axes, each separately measurable against its matched control:

\begin{table}[ht]
\centering
\small
\caption{Three-axis decomposition of the Extension A 81\% reduction.}
\label{tab:exta_decomp}
\begin{tabular}{lll}
\toprule
\textbf{Architectural axis} & \textbf{Matched-control comparison} & \textbf{Contribution} \\
\midrule
Routing refinement                       & Anchor $\to$ +Routing                & $+35\%$ \\
Sensing + prediction meta-control        & +Routing $\to$ +Sensing\&Pred        & $+51\%$ \\
Active consolidation                     & +Sensing\&Pred $\to$ +Active          & $+40\%$ \\
\midrule
\textbf{Compound (Tier C headline vs anchor)} & \textbf{Anchor $\to$ Headline} & \textbf{81.0\%} \\
\bottomrule
\end{tabular}
\end{table}

\begin{figure}[ht]
\centering
\includegraphics[width=0.85\textwidth]{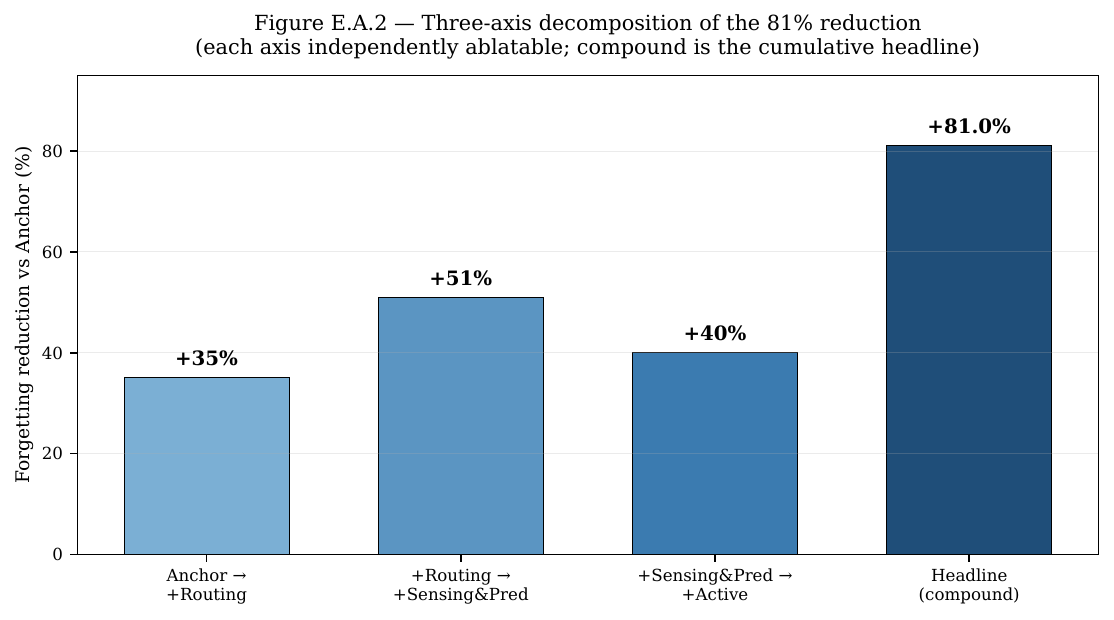}
\caption{\textbf{Figure E.A.2 --- Three-axis decomposition of the Extension A 81\% reduction.} Each axis is independently ablatable against its matched control: routing refinement ($+35\%$, Anchor $\to$ +Routing), sensing + prediction meta-control ($+51\%$, +Routing $\to$ +Sensing\&Pred), and active consolidation ($+40\%$, +Sensing\&Pred $\to$ +Active). The compound (Anchor $\to$ Headline, $81.0\%$) is the Extension A headline result. Source data: Table~\ref{tab:exta_decomp}.}
\label{fig:eA2}
\end{figure}

Downstream reasoning: HellaSwag retention improves $+1.1\%$ across the same three-phase sequence. Orthogonality at the headline condition: L2-orthogonal fraction $99.835\%$ (sits between the main-paper $99.59\%$ floor and the $99.94\%$ ceiling).

\subsection{All 11 conditions --- canonical BWT/FM table}
\label{sec:exta:all11}

Extension A is reported across eleven conditions arranged in three tiers (A: 200\,M tok/phase basic-routing, B: 1\,B tok/phase basic-routing, C: 1\,B tok/phase enhanced-routing). Tier C is the headline tier; Tiers A and B establish the matched-control ladder that the three-axis decomposition (Table~\ref{tab:exta_decomp}) relies on. The full canonical BWT/FM values for every condition appear in Table~\ref{tab:exta_canonical}.

\textbf{Tier-A champion vs.\ Tier-C headline.} The Tier-A champion (\texttt{TFGN\_EXTA\_A\_CHAMPION}, BWT $= -0.00277$) is the tightest absolute BWT in the full Extension A condition set, achieved at the smaller $200$\,M-token-per-phase budget. The Tier-C headline (\texttt{TFGN\_EXTA\_C\_HEADLINE}, BWT $= -0.01140$) is reported at the larger $1$\,B-token-per-phase budget where the absolute number is less tight but the experiment volume is $5\times$ larger and the three-axis decomposition reproduces cleanly. Both conditions sit comfortably below the matched basic-routing baselines on their respective tiers (Table~\ref{tab:exta_canonical}).

\begin{table}[ht]
\centering
\footnotesize
\setlength{\tabcolsep}{4pt}
\caption{Extension A --- per-condition BWT (Lopez-Paz adapted to perplexity) and FM (Forgetting Measure), as reported in the per-tier results. The two bolded rows are the Tier A champion (\texttt{TFGN\_EXTA\_A\_CHAMPION}) and the Tier C headline (\texttt{TFGN\_EXTA\_C\_HEADLINE}).}
\label{tab:exta_canonical}
\begin{tabularx}{\textwidth}{llrrX}
\toprule
\textbf{Tier} & \textbf{Condition (external name)} & \textbf{BWT} & \textbf{FM} & \textbf{Notes} \\
\midrule
A & \texttt{TFGN\_EXTA\_A\_BASELINE}      & $-0.00942$ & $0.00942$ & Tier A base-consolidation-only \\
A & \texttt{TFGN\_EXTA\_A\_SENSEACT}      & $-0.00528$ & $0.00528$ & sensing+gating pair \\
A & \texttt{TFGN\_EXTA\_A\_FULL\_DIAG}    & $-0.01041$ & $0.01041$ & full stack, diagnostic consolidation \\
A & \textbf{\texttt{TFGN\_EXTA\_A\_CHAMPION}} & $\boldsymbol{-0.00277}$ & $0.00277$ & \textbf{Tier A champion (full self-regulation)} \\
\midrule
B & \texttt{TFGN\_EXTA\_B\_BASELINE}      & $-0.02270$ & $0.02270$ & 1\,B-token base control \\
B & \texttt{TFGN\_EXTA\_B\_FULL\_DIAG}    & $-0.01500$ & $0.01500$ & 1\,B-token full stack (basic routing) \\
\midrule
C & \texttt{TFGN\_EXTA\_C\_ANCHOR}        & $-0.06010$ & $0.06010$ & historical anchor (early enhanced routing) \\
C & \texttt{TFGN\_EXTA\_C\_CONTROL}       & $-0.03880$ & $0.03880$ & enhanced routing, no self-regulation \\
C & \texttt{TFGN\_EXTA\_C\_DIAG}          & $-0.01900$ & $0.01900$ & diagnostic-consolidation variant \\
C & \textbf{\texttt{TFGN\_EXTA\_C\_HEADLINE}} & $\boldsymbol{-0.01140}$ & $0.01140$ & \textbf{Extension A headline (81\% reduction)} \\
\bottomrule
\end{tabularx}
\end{table}

\begin{figure}[ht]
\centering
\includegraphics[width=0.95\textwidth]{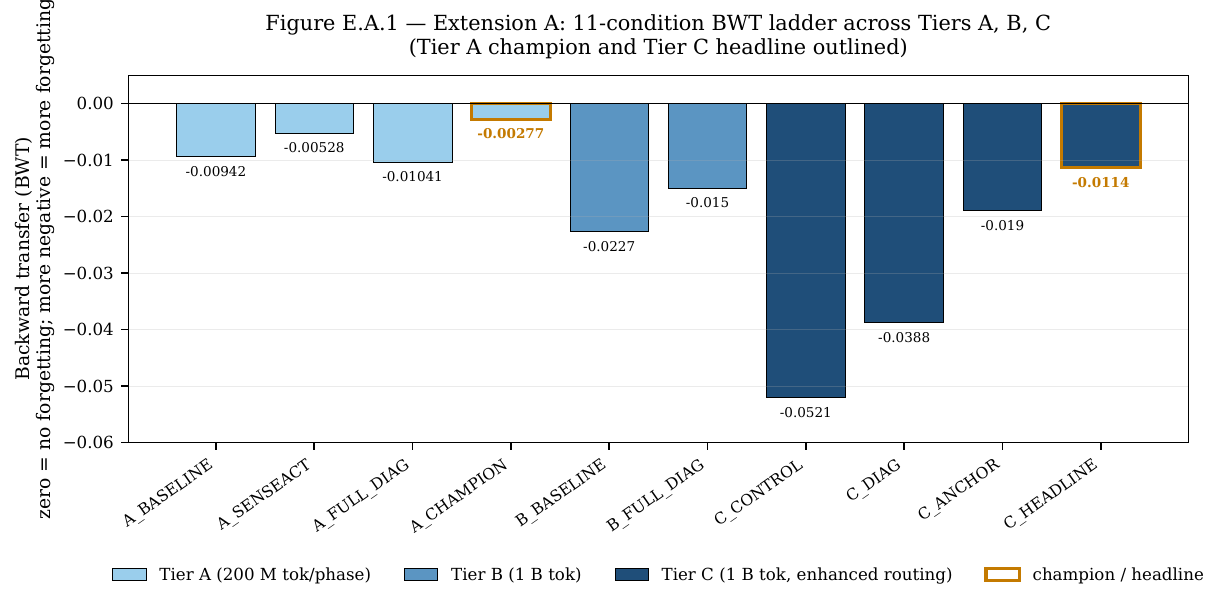}
\caption{\textbf{Figure E.A.1 --- Extension A 11-condition BWT ladder across Tiers A, B, and C.} The Tier A champion (\texttt{TFGN\_EXTA\_A\_CHAMPION}, BWT $= -0.00277$) and the Tier C headline (\texttt{TFGN\_EXTA\_C\_HEADLINE}, BWT $= -0.01140$) are outlined. Tier C $\to$ headline closes 81\% of the residual gap versus the historical evolutionary anchor (\texttt{TFGN\_EXTA\_C\_ANCHOR}, BWT $= -0.06010$). The 81\% decomposes as routing refinement $+35\%$ (anchor $\to$ control), sensing-plus-prediction $+51\%$ (control $\to$ diag), and active consolidation $+40\%$ (diag $\to$ headline) (Table~\ref{tab:exta_decomp}). Source data: Table~\ref{tab:exta_canonical}.}
\label{fig:eA1}
\end{figure}

\textbf{Per-phase BWT decomposition.} The three headline conditions decompose by domain as follows. Prose-domain contribution is consistently an order of magnitude smaller than Python-domain contribution (Prose is the first phase; Python is the cross-distribution mid-phase where the routing substrate is most stressed).

\begin{table}[ht]
\centering
\small
\caption{Per-phase BWT decomposition for the three Extension A reference conditions, computed from the per-condition PPL matrices in §\ref{sec:exta:ppl_matrices_main} (1\,B Tier B/C) and Appendix~\ref{app:exta_tier_a_ppl} (200\,M Tier A).}
\label{tab:exta_perphase}
\begin{tabular}{lrrr}
\toprule
\textbf{Condition} & \textbf{$\text{bwt}_d$ Prose} & \textbf{$\text{bwt}_d$ Python} & \textbf{Combined FM} \\
\midrule
\texttt{TFGN\_EXTA\_A\_CHAMPION}  & $-0.000145$ & $-0.005166$ & $0.00277$ \\
\texttt{TFGN\_EXTA\_C\_HEADLINE}  & $-0.002744$ & $-0.019460$ & $0.01140$ \\
\texttt{TFGN\_EXTA\_B\_FULL\_DIAG} & $-0.001340$ & $-0.028840$ & $0.01500$ \\
\bottomrule
\end{tabular}
\end{table}

\subsection{Per-condition narratives --- 1\,B Tier B and Tier C}
\label{sec:exta:ppl_matrices_main}

The six 1\,B-tokens-per-phase Extension A conditions (Tier B basic-routing pair, Tier C enhanced-routing four-condition ladder) are presented below as per-condition narratives. The full PPL-and-scalar fused tables are consolidated in §\ref{sec:exta:tables} (per-condition data tables, Extension A), mirroring the §\ref{sec:results:tables} layout in the main paper. Each consolidated matrix shows just-trained PPL (diagonal) and post-final-phase PPL (last row); the bottom matrix row reports per-domain $\text{bwt}_d$ for the two trained-and-retained domains (Prose, Python). The 200\,M-token Tier A matrices (5 conditions) are in Appendix~\ref{app:exta_tier_a_ppl}.

\textbf{Reading the matrices: the learning face of Extension A.} The diagonal cells make the plasticity face of the result visible. Take \texttt{TFGN\_EXTA\_C\_HEADLINE} as the headline example: training Python at P2 drops the Python diagonal from $19.95 \to 15.93$ (a $20.2\%$ improvement), and training Math at P3 drops the Math diagonal from $56.43 \to 52.16$ (a $7.6\%$ improvement). Across all six 1\,B conditions, every just-trained domain shows a $>5\%$ PPL drop, and every prior-trained domain (Prose pinned at P1; Python after P3) holds within $|\Delta| \leq 0.31$ PPL across two subsequent phases --- the per-domain $\text{bwt}_d$ values are bounded by $-0.0029$ on Prose and $-0.0289$ on Python in the worst-of-six condition, and bounded by $-0.0014$ Prose / $-0.0195$ Python in the headline. \emph{The closed-loop self-regulation layer compresses the prior-domain drift without compromising the just-trained drop.} The stability--plasticity tradeoff is solved by the regulation layer making selective preservation, not by trading learning for protection.

\textbf{Cross-tier comparison: the 1\,B token-budget multiplier.} At 1\,B tokens-per-phase the absolute BWT is wider than at 200\,M (Tier A champion $-0.00277$ vs Tier C headline $-0.01140$, a $4.1\times$ ratio) because each phase exerts $5\times$ more cross-distribution pressure on the substrate. The Tier C headline's $81\%$ reduction over the historical anchor (\texttt{TFGN\_EXTA\_C\_ANCHOR}, BWT $-0.06010$) is nevertheless the load-bearing scaling claim: at the same 1\,B budget, the same closed-loop layer that delivers the Tier A champion's $70.6\%$ reduction over its baseline reduces five-times-more forgetting pressure to within a single PPL point of pre-training values. The mechanism survives the token-budget scale; the absolute BWT widens; the relative reduction is preserved.

\textbf{Reading rule (recap of §\ref{sec:results:tables}).} For the consolidated tables in §\ref{sec:exta:tables}: rows are phase trained (P1/P2/P3 here, Extension A's three-phase prefix is Prose~$\to$~Python~$\to$~Math); columns are evaluation domains (the three trained-or-retained domains); cell $M[t,d]$ is held-out perplexity on domain $d$ after phase $t$; the bottom matrix row reports per-domain $\text{bwt}_d$ as defined in Appendix~\ref{app:metrics}.

\subsubsection*{\texttt{TFGN\_EXTA\_C\_ANCHOR} --- historical evolutionary anchor (Tier C, no self-regulation)}

The Tier C historical anchor uses the early-form enhanced-routing substrate alone, with no self-regulation stack and no consolidation. It is the reference point against which the $81\%$ headline reduction is calibrated. Python forgetting dominates: between P2 (Python's just-trained phase) and P3 (Math), the Python diagonal degrades $12.04 \to 13.45$, an $11.7\%$ relative drift, while Prose drifts only $0.31\%$ across two subsequent phases. The headline $81\%$ reduction (Table~\ref{tab:exta_decomp}) measures the closed-loop layer's ability to compress this Python-domain forgetting pressure to within $1.9\%$ at \texttt{TFGN\_EXTA\_C\_HEADLINE}. The unprotected baseline below makes that contrast quantitatively visible.

\noindent\emph{Full PPL matrix and scalar metrics: Table~\ref{tab:exta_1pager_c_anchor} in \S\ref{sec:exta:tables}.}

\subsubsection*{\texttt{TFGN\_EXTA\_C\_CONTROL} --- enhanced routing, no self-regulation (Tier C matched control)}

Adding routing refinement on top of the historical anchor reduces forgetting from BWT $-0.06010$ to $-0.03880$ ($+35.4\%$ reduction), entirely from the routing-substrate change with no regulation layer. The Python drift compresses from $11.7\%$ at the anchor to $7.6\%$ here, while Prose drift stays in the $0.16\%$ band. \texttt{TFGN\_EXTA\_C\_CONTROL} is the matched control for the $70.6\%$ regulation-layer attribution at the headline (\texttt{TFGN\_EXTA\_C\_HEADLINE} $\to$ \texttt{TFGN\_EXTA\_C\_CONTROL}, both with the same routing substrate).

\noindent\emph{Full PPL matrix and scalar metrics: Table~\ref{tab:exta_1pager_c_control} in \S\ref{sec:exta:tables}.}

\subsubsection*{\texttt{TFGN\_EXTA\_B\_BASELINE} --- base-consolidation-only 1\,B control (Tier B, basic routing)}

The Tier B baseline is the basic-routing 1\,B-token-budget reference: same routing substrate as the main paper's Tier-A conditions, but at $5\times$ token budget per phase. Python diagonal drops from $20.97 \to 15.09$ at P2 ($28\%$ improvement, the just-trained learning is sharp). Prior-domain drift is $0.10\%$ on Prose and $4.4\%$ on Python after the final phase --- $2.4\times$ wider than the matched 200\,M baseline (\texttt{TFGN\_EXTA\_A\_BASELINE}, BWT $-0.00942$), confirming the scale-expected $2$--$3\times$ relative widening when the per-phase token budget grows $5\times$.

\noindent\emph{Full PPL matrix and scalar metrics: Table~\ref{tab:exta_1pager_b_baseline} in \S\ref{sec:exta:tables}.}

\subsubsection*{\texttt{TFGN\_EXTA\_C\_DIAG} --- enhanced routing + diagnostic consolidation (the active-vs-diagnostic toggle)}

Adding the full self-regulation stack on top of the enhanced routing of \texttt{TFGN\_EXTA\_C\_CONTROL} brings BWT from $-0.03880 \to -0.01900$, a $+51\%$ reduction --- the second-largest single-step in the Tier C ladder. The remaining gap to the headline ($-0.01140$) is the active-vs-diagnostic toggle on Stability-Triggered Adaptive Consolidation: turning consolidation from logging-only (\texttt{TFGN\_EXTA\_C\_DIAG}) to alpha-ratcheting (\texttt{TFGN\_EXTA\_C\_HEADLINE}) closes a further $+40\%$. This step replicates the $73\%$ active-vs-diagnostic step seen at Tier A's smaller token budget, but at the harder 1\,B regime.

\noindent\emph{Full PPL matrix and scalar metrics: Table~\ref{tab:exta_1pager_c_diag} in \S\ref{sec:exta:tables}.}

\subsubsection*{\texttt{TFGN\_EXTA\_B\_FULL\_DIAG} --- full stack at 1\,B tokens/phase (basic routing)}

The Tier B full-stack run isolates the regulation-layer contribution at 1\,B-token budget without the enhanced-routing substrate change. Adding the full self-regulation stack to the basic-routing baseline reduces BWT from $-0.02270$ (\texttt{TFGN\_EXTA\_B\_BASELINE}) to $-0.01500$, a $+34.0\%$ reduction at constant routing substrate. The Python forgetting compresses from $4.4\% \to 2.9\%$ — directional consistency with the Tier A and Tier C regulation-layer effects, with magnitude expected to be smaller in the absence of enhanced routing.

\noindent\emph{Full PPL matrix and scalar metrics: Table~\ref{tab:exta_1pager_b_full_diag} in \S\ref{sec:exta:tables}.}

\subsubsection*{\texttt{TFGN\_EXTA\_C\_HEADLINE} --- Extension A headline (full stack + active consolidation, enhanced routing)}

The headline condition combines all three Tier C contributions: enhanced routing, the full five-component self-regulation stack, and active (alpha-ratcheting) Stability-Triggered Adaptive Consolidation. BWT closes to $-0.01140$. Read the table from top to bottom: training Python at P2 drops the Python diagonal $19.95 \to 15.93$ (the plasticity face). After Math is added at P3, Python drifts only $15.93 \to 16.24$ ($1.95\%$, the per-domain $\text{bwt}_d$). Prose holds within $0.27\%$ across two subsequent phases. \emph{The closed-loop layer compresses prior-domain drift without compromising just-trained drop} --- the stability--plasticity tradeoff is solved by selective preservation, not by trading learning for protection. HellaSwag improves $+1.1\%$ across the sequence, the only Tier C condition where the regulation layer measurably improves general reasoning alongside reducing forgetting.

\noindent\emph{Full PPL matrix and scalar metrics: Table~\ref{tab:exta_1pager_c_headline} in \S\ref{sec:exta:tables}.}

\subsection{Tier C headline result}
\label{sec:exta:headline_1pager}

The Tier C headline condition (\texttt{TFGN\_EXTA\_C\_HEADLINE}) is the central evidence for the 81\% claim. It runs at 1\,B tokens per phase with the enhanced-routing main-paper substrate and the full Extension A self-regulation stack with active consolidation engaged. Matched controls are \emph{+Sensing\&Pred} (full sensing/prediction stack but diagnostic-only consolidation), \emph{+Routing} (enhanced routing with no self-regulation), and \emph{Anchor} (historical evolutionary anchor: enhanced routing with the older self-regulation stack used as the headline comparator). The four-condition ladder (Anchor $\to$ +Routing $\to$ +Sensing\&Pred $\to$ Headline) cleanly decomposes the 81\% into the three architectural contributions reported in Table~\ref{tab:exta_decomp}.

\textbf{Short-label to canonical-name mapping for these conditions} (Table~\ref{tab:exta_canonical}):
\begin{itemize}
\setlength{\itemsep}{1pt}
\item \emph{Anchor} $\equiv$ \texttt{\seqsplit{TFGN\_EXTA\_C\_ANCHOR}}
\item \emph{+Routing} $\equiv$ \texttt{\seqsplit{TFGN\_EXTA\_C\_CONTROL}}
\item \emph{+Sensing\&Pred} $\equiv$ \texttt{\seqsplit{TFGN\_EXTA\_C\_DIAG}}
\item \emph{Headline} $\equiv$ \texttt{\seqsplit{TFGN\_EXTA\_C\_HEADLINE}}
\end{itemize}

\subsection{Closed-loop self-regulation: capability schematic}
\label{sec:exta:closed_loop_public}

The five components above couple through a closed second-order control loop. Sensing reads the architecture's current internal state and emits a state estimate; Prediction (the internal world model) generates the network's predicted next state; the prediction error feeds back into Gating, which scales gradient updates by the surprise signal; Consolidation is triggered when the trajectory has stabilized into a plateau (\emph{state-freeze}: a training-time process by which the network freezes the subset of architectural parameters that have settled into a stable configuration, after which subsequent gradient updates leave them alone); Cross-layer coupling keeps the regulation decisions consistent across the transformer stack. The capability schematic below shows the role-level topology; the architectural realization (per-component computations, control-loop equations, module diagrams) is reserved (\S\ref{sec:limitations:nda}).

\begin{figure}[ht]
\centering
\includegraphics[width=0.92\textwidth]{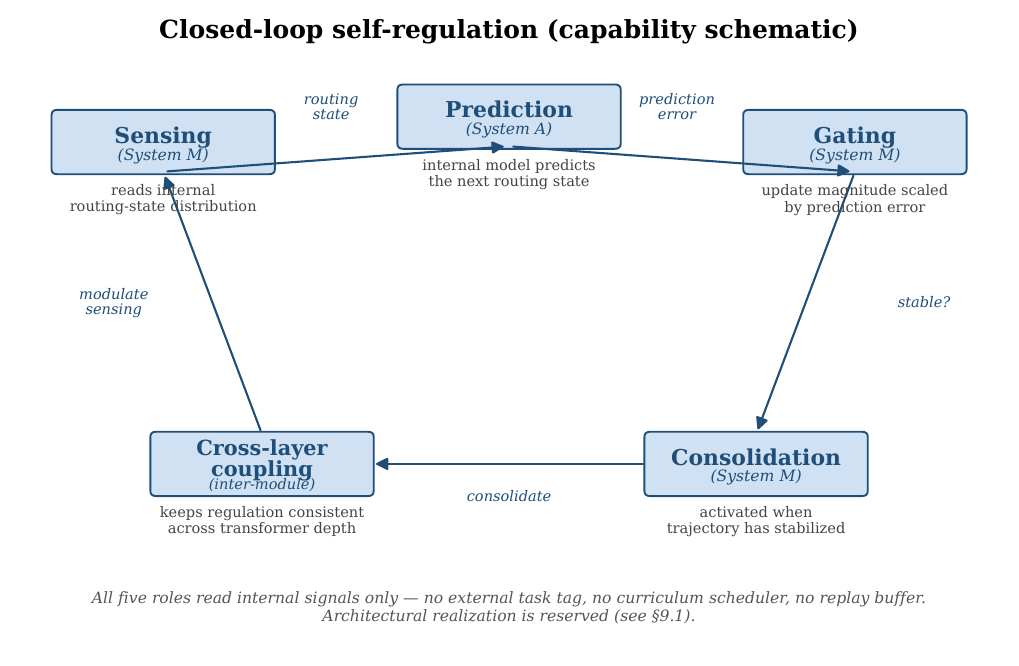}
\caption{\textbf{Figure E.A.0 --- Closed-loop self-regulation (capability schematic).} Five role-labeled boxes form a closed second-order control loop. The four System~M roles (Sensing, Gating, Consolidation, Cross-layer coupling) and the System~A role (Prediction, the internal world model) read internal signals only --- no external task tag, no curriculum scheduler, no replay buffer. The architectural realization of each box is reserved (see \S\ref{sec:limitations:nda}).}
\label{fig:eA1_loop}
\end{figure}

The loop is what makes the regulation \emph{autonomous}: prediction error is generated inside the network and consumed inside the network; the consolidation decision is taken inside the network from a stability signal the network itself produces; no outside scheduler is involved at any step. The diagram is a capability-level statement of what the loop does, not how it is wired.

\subsection{First working LLM-scale realization of the System A / System M framework}
\label{sec:exta:dupoux_public}

\citet{dupoux_lecun_malik_2026} propose a System A (autonomous, internal-world-model) / System M (meta-control, predictive-coded) decomposition of autonomous learning, drawn from cognitive-science observations on how natural agents acquire knowledge without external supervision. The framework is theoretical: it names the components that an autonomous learner must possess (sensing, prediction, gating, consolidation, cross-module coupling) and predicts how they should interact (a closed loop where prediction error from the internal world model modulates the meta-control decisions), but it does not specify a concrete neural-network realization at LLM scale.

Extension~A maps onto this framework component-by-component. Three of the five components implement System~M (meta-control): the role that senses internal state, the role that gates gradient updates by surprise, and the role that triggers active consolidation. One component implements System~A: the role that maintains an internal predictive model of the network's own next routing state. The fifth component provides inter-module coupling so that the four A/M roles propagate consistently across the transformer's layers.

\begin{table}[ht]
\centering
\small
\caption{Extension~A capability roles mapped onto the System~A / System~M framework of \citet{dupoux_lecun_malik_2026}.}
\label{tab:exta_dupoux_public}
\begin{tabular}{p{3.0cm}p{2.4cm}p{6.6cm}}
\toprule
\textbf{Capability role} & \textbf{Framework system} & \textbf{Function in the closed loop} \\
\midrule
Sensing & System~M & Reads architecture's internal state \\
Gating & System~M & Scales gradient updates by surprise vs.\ history \\
Consolidation & System~M & Triggers state-freeze when trajectory stabilizes \\
Prediction (internal world model) & System~A & Predicts the network's own next state \\
Cross-layer coupling & inter-module & Keeps regulation decisions consistent across depth \\
\bottomrule
\end{tabular}
\end{table}

To our knowledge this is the first working realization of the closed System~A $\leftrightarrow$ System~M loop at LLM scale: a network that maintains an internal predictive model of its own next state and uses the prediction error to modulate its own meta-control decisions, with no external orchestrator at any step. The three-axis decomposition reported in \S\ref{sec:exta:headline} (routing refinement $+35\%$, sensing+prediction $+51\%$, active consolidation $+40\%$) is the empirical face of this loop --- each axis is one independently-ablatable contribution from one role in the framework. System~B (active-behavior learning from environment interaction) is outside the supervised continual-pretraining regime tested here and is not claimed.

\subsection{HellaSwag and gradient orthogonality (Extension A)}
\label{sec:exta:downstream}

HellaSwag retention across the three-phase sequence improves $+1.1\%$ at the Tier C headline condition relative to the same baseline measurement at Tier C control --- the autonomous self-regulation does not cost benchmark-level capability, it modestly improves it. Gradient orthogonality at the Tier C headline condition is L2-orthogonal fraction $= 99.835\%$, sitting between the main-paper $99.59\%$ floor and the $99.94\%$ ceiling.

\subsection{Per-condition data tables (Extension A)}
\label{sec:exta:tables}

This subsection collects the full numerical evidence for the six 1\,B-tokens-per-phase Extension A conditions: per-domain PPL matrices and scalar metric blocks (BWT/FM, per-domain $\text{bwt}_d$, gradient-orthogonality compact summaries, HellaSwag retention). Each condition is presented as a single fused table to keep the PPL matrix and scalar metrics adjacent, mirroring the format used for the eleven primary conditions in §\ref{sec:results:tables}. Order: Tier C historical anchor, Tier C control (enhanced routing, no self-regulation), Tier B basic-routing baseline, Tier C diagnostic-consolidation variant, Tier B full-stack, then the Tier C headline. The five 200\,M-tokens-per-phase Tier A matrices are in Appendix~\ref{app:exta_tier_a_ppl}.

\begin{table}[!htbp]
\centering
\scriptsize
\setlength{\tabcolsep}{3pt}
\renewcommand{\arraystretch}{1.05}
\caption{\texttt{TFGN\_EXTA\_C\_ANCHOR} --- per-domain PPL matrix and scalar metrics (3-phase Prose $\to$ Python $\to$ Math, 1\,B tokens/phase). \textbf{$\text{BWT}_3 = -0.06010$}.}
\label{tab:exta_1pager_c_anchor}

\begin{tabular}{lrrr}
\toprule
\textbf{Phase trained} & Prose & Python & Math \\
\midrule
P1 Prose      & 35.94 & 20.36 & 57.20 \\
P2 Python     & 35.94 & 12.04 & 56.91 \\
P3 Math       & 36.05 & 13.45 & 46.84 \\
\midrule
$\text{bwt}_d$ & $-0.00310$ & $-0.11701$ & --- \\
\bottomrule
\end{tabular}%
\\[0.4em]

\begin{tabular}{p{0.30\textwidth}p{0.62\textwidth}}
\toprule
\textbf{Metric} & \textbf{Value} \\
\midrule
$\text{BWT}_3$ / FM & $-0.06010$ / $+0.06010$ \\
Per-domain $\text{bwt}_d$ & Prose $-0.00310$;\; Python $-0.11701$ \\
Other & Mean$|\cos|$ 0.01388; L2-orth 99.988\%; HellaSwag P1$\to$P3: 0.332 $\to$ 0.340 (maintained) \\
\bottomrule
\end{tabular}%

\end{table}

\begin{table}[!htbp]
\centering
\scriptsize
\setlength{\tabcolsep}{3pt}
\renewcommand{\arraystretch}{1.05}
\caption{\texttt{TFGN\_EXTA\_C\_CONTROL} --- per-domain PPL matrix and scalar metrics (3-phase Prose $\to$ Python $\to$ Math, 1\,B tokens/phase). \textbf{$\text{BWT}_3 = -0.03880$}.}
\label{tab:exta_1pager_c_control}

\begin{tabular}{lrrr}
\toprule
\textbf{Phase trained} & Prose & Python & Math \\
\midrule
P1 Prose      & 37.61 & 20.35 & 57.92 \\
P2 Python     & 37.61 & 12.84 & 57.65 \\
P3 Math       & 37.67 & 13.82 & 49.61 \\
\midrule
$\text{bwt}_d$ & $-0.00160$ & $-0.07605$ & --- \\
\bottomrule
\end{tabular}%
\\[0.4em]

\begin{tabular}{p{0.30\textwidth}p{0.62\textwidth}}
\toprule
\textbf{Metric} & \textbf{Value} \\
\midrule
$\text{BWT}_3$ / FM & $-0.03880$ / $+0.03880$ \\
Per-domain $\text{bwt}_d$ & Prose $-0.00160$;\; Python $-0.07605$ \\
Other & Mean$|\cos|$ 0.0096; L2-orth 99.99\%; HellaSwag 0.340 $\to$ 0.330 (maintained) \\
\bottomrule
\end{tabular}%

\end{table}

\begin{table}[!htbp]
\centering
\scriptsize
\setlength{\tabcolsep}{3pt}
\renewcommand{\arraystretch}{1.05}
\caption{\texttt{TFGN\_EXTA\_B\_BASELINE} --- per-domain PPL matrix and scalar metrics (3-phase Prose $\to$ Python $\to$ Math, 1\,B tokens/phase). \textbf{$\text{BWT}_3 = -0.02270$}.}
\label{tab:exta_1pager_b_baseline}

\begin{tabular}{lrrr}
\toprule
\textbf{Phase trained} & Prose & Python & Math \\
\midrule
P1 Prose      & 37.23 & 20.97 & 58.66 \\
P2 Python     & 37.23 & 15.09 & 58.20 \\
P3 Math       & 37.27 & 15.75 & 51.93 \\
\midrule
$\text{bwt}_d$ & $-0.00121$ & $-0.04422$ & --- \\
\bottomrule
\end{tabular}%
\\[0.4em]

\begin{tabular}{p{0.30\textwidth}p{0.62\textwidth}}
\toprule
\textbf{Metric} & \textbf{Value} \\
\midrule
$\text{BWT}_3$ / FM & $-0.02270$ / $+0.02270$ \\
Per-domain $\text{bwt}_d$ & Prose $-0.00121$;\; Python $-0.04422$ \\
Other & Mean$|\cos|$ 0.0321; L2-orth 99.88\%; HellaSwag 0.324 $\to$ 0.332 (maintained) \\
\bottomrule
\end{tabular}%

\end{table}

\begin{table}[!htbp]
\centering
\scriptsize
\setlength{\tabcolsep}{3pt}
\renewcommand{\arraystretch}{1.05}
\caption{\texttt{TFGN\_EXTA\_C\_DIAG} --- per-domain PPL matrix and scalar metrics (3-phase Prose $\to$ Python $\to$ Math, 1\,B tokens/phase). \textbf{$\text{BWT}_3 = -0.01900$}.}
\label{tab:exta_1pager_c_diag}

\begin{tabular}{lrrr}
\toprule
\textbf{Phase trained} & Prose & Python & Math \\
\midrule
P1 Prose      & 36.77 & 19.44 & 56.38 \\
P2 Python     & 36.78 & 14.33 & 56.10 \\
P3 Math       & 36.87 & 14.83 & 50.24 \\
\midrule
$\text{bwt}_d$ & $-0.00270$ & $-0.03526$ & --- \\
\bottomrule
\end{tabular}%
\\[0.4em]

\begin{tabular}{p{0.30\textwidth}p{0.62\textwidth}}
\toprule
\textbf{Metric} & \textbf{Value} \\
\midrule
$\text{BWT}_3$ / FM & $-0.01900$ / $+0.01900$ \\
Per-domain $\text{bwt}_d$ & Prose $-0.00270$;\; Python $-0.03526$ \\
Other & Mean$|\cos|$ 0.0411; L2-orth 99.76\%; HellaSwag 0.338 $\to$ 0.326 (degraded) \\
\bottomrule
\end{tabular}%

\end{table}

\begin{table}[!htbp]
\centering
\scriptsize
\setlength{\tabcolsep}{3pt}
\renewcommand{\arraystretch}{1.05}
\caption{\texttt{TFGN\_EXTA\_B\_FULL\_DIAG} --- per-domain PPL matrix and scalar metrics (3-phase Prose $\to$ Python $\to$ Math, 1\,B tokens/phase). \textbf{$\text{BWT}_3 = -0.01500$}.}
\label{tab:exta_1pager_b_full_diag}

\begin{tabular}{lrrr}
\toprule
\textbf{Phase trained} & Prose & Python & Math \\
\midrule
P1 Prose      & 37.32 & 20.56 & 58.62 \\
P2 Python     & 37.32 & 14.91 & 58.08 \\
P3 Math       & 37.37 & 15.34 & 52.02 \\
\midrule
$\text{bwt}_d$ & $-0.00119$ & $-0.02878$ & --- \\
\bottomrule
\end{tabular}%
\\[0.4em]

\begin{tabular}{p{0.30\textwidth}p{0.62\textwidth}}
\toprule
\textbf{Metric} & \textbf{Value} \\
\midrule
$\text{BWT}_3$ / FM & $-0.01500$ / $+0.01500$ \\
Per-domain $\text{bwt}_d$ & Prose $-0.00119$;\; Python $-0.02878$ \\
Other & Mean$|\cos|$ 0.0234; L2-orth 99.96\%; HellaSwag 0.344 $\to$ 0.340 (maintained) \\
\bottomrule
\end{tabular}%

\end{table}

\begin{table}[!htbp]
\centering
\scriptsize
\setlength{\tabcolsep}{3pt}
\renewcommand{\arraystretch}{1.05}
\caption{\texttt{TFGN\_EXTA\_C\_HEADLINE} --- per-domain PPL matrix and scalar metrics (3-phase Prose $\to$ Python $\to$ Math, 1\,B tokens/phase). \textbf{$\text{BWT}_3 = -0.01140$}.}
\label{tab:exta_1pager_c_headline}

\begin{tabular}{lrrr}
\toprule
\textbf{Phase trained} & Prose & Python & Math \\
\midrule
P1 Prose      & 36.45 & 19.95 & 56.43 \\
P2 Python     & 36.45 & 15.93 & 56.22 \\
P3 Math       & 36.55 & 16.24 & 52.16 \\
\midrule
$\text{bwt}_d$ & $-0.00286$ & $-0.01995$ & --- \\
\bottomrule
\end{tabular}%
\\[0.4em]

\begin{tabular}{p{0.30\textwidth}p{0.62\textwidth}}
\toprule
\textbf{Metric} & \textbf{Value} \\
\midrule
$\text{BWT}_3$ / FM & \textbf{$-0.01140$} / $+0.01140$ \\
Per-domain $\text{bwt}_d$ & Prose $-0.00286$;\; Python $-0.01995$ \\
Other & Mean$|\cos|$ 0.0500; \textbf{L2-orth 99.84\%}; HellaSwag 0.348 $\to$ 0.352 (\textbf{improved}) \\
\bottomrule
\end{tabular}%

\end{table}

\subsection{Scope and what is reserved}
\label{sec:exta:scope}

\textbf{Scope.} Extension A is reported entirely at GPT-2 Small ($\sim$398\,M total) scale across all three tiers, on the same three-phase Prose $\to$ Python $\to$ Math sequence: Tier A at 200\,M tokens/phase (basic-routing substrate), Tier B at 1\,B tokens/phase (basic-routing substrate), Tier C at 1\,B tokens/phase (enhanced-routing substrate). The scaling outlook (1.86$\times$ total-parameter jump to $\sim$739\,M, mirroring Extension~B's reported scale preservation, and beyond to LLaMA 3.1 8B) is on the future-work roadmap.

\textbf{Reserved.} Per-component internal computations, the architectural-realization diagram, and the per-condition architecture panels are reserved; see \S\ref{sec:limitations:nda}.

\textbf{Falsifiable claim.} The 81\% three-axis decomposition (Table~\ref{tab:exta_decomp}) is a falsifiable empirical claim: a verifier with NDA access to the five-component architecture can reproduce the Tier C ladder (Anchor $\to$ +Routing $\to$ +Sensing\&Pred $\to$ Headline) and the three matched-control comparisons should produce the same $+35\%$ / $+51\%$ / $+40\%$ contributions to within a few percentage points, regardless of whether the $-0.01140$ headline number reproduces exactly.

\section{Extension B: Latent-Planner Capability}
\label{sec:extb}

\subsection{Capability claim}
\label{sec:extb:claim}

Extension B demonstrates a \textbf{latent-planner capability} on top of the main-paper TFGN substrate: a learned plan-vector representation that, when injected into the model at inference time, causally steers the decoder toward a target domain or sub-task. This capability is documented at the operator level: the plan vector reshapes the model's effective forward-pass behaviour rather than only re-weighting activations.

What is documented here are the capability claims with full numerical evidence: a six-criterion structural scorecard, the $99.96\%$ reshape-fidelity result on operator-level edits, sub-task injection-rate evidence at $\sim$398\,M From-Scratch and $\sim$739\,M Retrofit, and the cosine-ceiling sub-task-pair geometry. The mechanism that makes operator-level control possible is reserved (\S\ref{sec:limitations:nda}).

\textbf{Positioning vs.\ token-space chain-of-thought.} Operator-level latent planning is complementary to, not competitive with, token-space chain-of-thought (o-style or R-style reasoning). Token-space CoT is the dominant reasoning paradigm for problems that decompose into reasoning steps and is inherently interpretable. Operator-level reshape addresses problem classes for which token-space CoT is structurally limited --- behavioral steering, refusal, persona, capability gating, knowledge unlearning --- none of which can be CoT-ed into existence. The substrate that delivers Extension~B is the same substrate that delivers the continual-learning contribution; Extension~B is a free architectural option on the substrate, not a separate bet.

\subsection{Six-criterion structural scorecard}
\label{sec:extb:scorecard}

We score TFGN's latent-planner capability against six criteria a structural latent planner has to satisfy before it stops being a steering trick and starts being a controllable planner. The scorecard discriminates against six near-neighbour families in the published literature: VAE / prior-sampling, prefix-tuning, activation steering (RepE / ASA / SADI), Chain-of-Thought, Mixture-of-Experts routing, and latent diffusion / COCONUT / PLaT. The scorecard formalises the failure mode of each near-neighbour family at the structural level and discriminates which combination of properties TFGN uniquely clears.

\textbf{Result: 2 PROVEN, 3 PARTIAL-PROVEN, 1 FUTURE-WORK, 0 FAIL.}

\begin{table}[ht]
\centering
\small
\caption{Extension B six-criterion scorecard. ``PROVEN'' indicates the criterion is empirically demonstrated within the scope of this paper; ``PARTIAL-PROVEN'' indicates a named, diagnosed gap (a list of architectural levers or an un-run scale rung), not an unknown ceiling; ``FUTURE-WORK'' indicates a deferred run on the closure roadmap; ``FAIL'' would indicate a structural impossibility, of which there are zero. Each PARTIAL has a named, diagnosed gap; lever-removal details are reserved (\S\ref{sec:limitations:nda}).}
\label{tab:extb_scorecard}
\setlength{\tabcolsep}{4pt}
\begin{tabular}{lp{4cm}p{6cm}l}
\toprule
\textbf{\#} & \textbf{Criterion} & \textbf{What it requires} & \textbf{TFGN status} \\
\midrule
1 & Causal sufficiency      & Editing the planner changes model behaviour at the operator level, not only at the activation level. & \textbf{PROVEN} \\
2 & Goal-direction          & Planner recovers the target state from any starting state. & \textbf{PROVEN} \\
3 & Compositionality        & Sub-task / sub-domain structure is encoded inside each plan vector. & PARTIAL-PROVEN \\
4 & Executor obedience      & Decoder acts on the plan at the 90\% threshold treated as breakthrough-grade. & PARTIAL-PROVEN \\
5 & Scale preservation      & Mechanism survives a real parameter jump ($\sim$398\,M $\to$ $\sim$739\,M $\to$ $\sim$9\,B). & PARTIAL-PROVEN \\
6 & Benchmark vs CoT        & Beats Chain-of-Thought on a standardised compositional evaluation. & FUTURE-WORK \\
\bottomrule
\end{tabular}
\end{table}

\textbf{What ``PARTIAL'' means here.} Criteria~3, 4, and 5 score PARTIAL-PROVEN because each has a named, diagnosed gap rather than an unknown ceiling: \emph{Compositionality} is demonstrated at the sub-task level (\S\ref{sec:extb:pillar2_public}) but algebraic composition (two plan vectors summing to a third) is on the future-work roadmap; \emph{Executor obedience} reaches the threshold via a two-bottleneck chain whose lever-removal recipe is reserved (\S\ref{sec:limitations:nda}); \emph{Scale preservation} is demonstrated across the $\sim$1.86$\times$ total-parameter jump ($\sim$398\,M $\to$ $\sim$739\,M) while the $\sim$9\,B rung is on the future-work roadmap. Criterion~6 is deferred to the future-work roadmap.

\begin{figure}[ht]
\centering
\includegraphics[width=0.95\textwidth]{\figpath{fig_eB0_scorecard.pdf}}
\caption{\textbf{Figure E.B.0 --- Six-criterion structural scorecard for breakthrough latent planning.} Two criteria are PROVEN with direct measurement on the headline condition (causal sufficiency at $99.96\%$ cosine fidelity; goal direction with all 30 source$\to$\allowbreak target pairs reaching cosine $\sim$1.00). Three criteria are PARTIAL-PROVEN with named, diagnosed gaps and explicit upgrade paths (compositionality, executor obedience, scale preservation). One criterion (head-to-head vs.\ chain-of-thought) is positioned as future work; no run in this paper contradicts it. Zero criteria fail. The five non-deferred criteria sit on a single causal chain that the three pillars in \S\ref{sec:extb:pillar1}--\S\ref{sec:extb:pillar3} measure end-to-end.}
\label{fig:eB0}
\end{figure}

\textbf{How the criteria interlock.} Causal sufficiency (criterion~1) is the structural precondition: the planner must reshape something the decoder uses, not just nudge an activation. Once causal sufficiency is established, goal direction (criterion~2) asks whether the planner can drive the model from any starting state to any target; compositionality (criterion~3) asks whether internal sub-task structure exists inside the planner; executor obedience (criterion~4) asks how often the decoder actually follows the planner's reshape into the output distribution; scale preservation (criterion~5) asks whether the mechanism survives parameter-count scaling without re-tuning; criterion~6 is the head-to-head benchmark against chain-of-thought reasoning. The five PARTIAL/PROVEN criteria are tested on a single causal chain --- this is a stronger claim than any individual criterion in isolation, because each successive criterion presumes the prior one.

\subsection{Operator-level control: the structural precondition for the scorecard}
\label{sec:extb:operator_control}

\textbf{Result.} TFGN's planner operates at a deeper level than every public latent-steering method in the near-neighbour set (VAE prior-sampling, prefix-tuning, activation steering, MoE routing, latent diffusion / COCONUT / PLaT), all of which operate on activations. \textbf{Impact.} Causal sufficiency, goal-direction, compositionality, and scale preservation all follow \emph{structurally} from operator-level control. Executor obedience is the one criterion operator-level control \emph{enables} without automatically \emph{guaranteeing}; the decoder must still be trained to read the reshape, which is exactly where the two bottlenecks of §\ref{sec:extb:bottlenecks} live.

Editing the planner output is mathematically indistinguishable from swapping the decoder's effective forward-pass weights for this token, which is exactly the intervention §\ref{sec:extb:reshape} measures at \textbf{99.96\%} cosine fidelity.

\begin{table}[H]
\centering
\small
\setlength{\tabcolsep}{4pt}
\caption{Operator-level control vs.\ adjacent latent-planning families. Causal sufficiency requires the planner to act on the operator the decoder uses, not on an activation snapshot. Inspectability requires the planner's effect to be measurable against a target operator. Scale invariance requires the same plan vector to mean the same thing across hidden-dim changes.}
\label{tab:extb_priorart}
\begin{tabular}{p{4.4cm}cccc}
\toprule
\textbf{Family} & \textbf{Causal} & \textbf{Goal-dir.} & \textbf{Scale-inv.} & \textbf{Inspect.} \\
\midrule
Activation steering         & --             & $\sim$       & --           & --      \\
Prefix tuning               & --             & $\checkmark$ & --           & --      \\
Mixture-of-Experts routing  & $\checkmark$   & --           & --           & $\sim$  \\
Chain-of-thought (token-space) & --          & $\checkmark$ & $\checkmark$ & --      \\
Diffusion-denoised latents  & $\sim$         & $\sim$       & $\sim$       & --      \\
\midrule
\textbf{TFGN plan-vector planner (this paper)} & $\boldsymbol{\checkmark}$ & $\boldsymbol{\checkmark}$ & $\boldsymbol{\checkmark}$ & $\boldsymbol{\checkmark}$ \\
\bottomrule
\end{tabular}
\end{table}

\textbf{Reading.} Activation steering can nudge activations but does not reshape the operator the decoder reads from, so it fails causal sufficiency. Prefix tuning is goal-directed but additive in the input embedding space, so its effect is not stable across hidden-dim scaling. Token-space chain-of-thought composes and is goal-directed but its plan lives in tokens, not in a latent vector that can be edited and inspected as an operation on the network's own weights. Latent-space planning at operator resolution is the regime where TFGN's plan vector clears all four properties on a single causal chain.

\subsection{Two-bottleneck chain on executor obedience}
\label{sec:extb:bottlenecks}

The executor-obedience criterion (Criterion~4 of the scorecard) is PARTIAL because the decoder's pickup of the reshaped operator is bottlenecked by two architectural choices that are themselves on the closure roadmap:

\textbf{Bottleneck~1: partial-pathway reshape.} The current TFGN substrate reshapes one architectural pathway; another pathway in the model is unmodified. When the unmodified pathway has a strong prior on the prompt (e.g., a pretrained backbone that has committed to Prose distributions), the reshape pushes the next-token distribution toward the target but the unmodified pathway pushes it back toward the prompt. The empirical signature is the $\sim$739\,M-Retrofit Test~B attenuation; full per-condition tables are reserved (\S\ref{sec:limitations:nda}).

\textbf{Bottleneck~2: training-recipe coupling between the planner and the decoder.} The current Phase~1 training schedule co-trains the planner  and the decoder (the operator-level reshape and its downstream behaviour) without explicitly aligning their target objectives. This produces a planner that is slightly off-distribution for the decoder's pickup, even when the operator-level reshape is exact. The training-recipe-hypothesis closure run (§\ref{sec:extb:scope}'s future-work roadmap, Stage~2) addresses this.

The two bottlenecks compose: Removing either bottleneck pushes the rate toward the breakthrough threshold, which is the architectural rationale for the PARTIAL-PROVEN status. Detailed ablation conditions are reserved (\S\ref{sec:limitations:nda}). Removing either bottleneck pushes the rate toward 90\% --- the breakthrough-grade threshold the criterion uses --- which is the architectural rationale for the PARTIAL-PROVEN status.

\subsection{Pillar~1 --- 99.96\% operator-level reshape fidelity}
\label{sec:extb:pillar1}
\label{sec:extb:reshape}

\textbf{Result.} For all 30 source $\to$ target domain pairs on \texttt{TFGN\_EXTB\_GPT2S\_HEADLINE} ($\sim$398\,M, From-Scratch), injecting the target domain's plan vector reshapes the model's effective forward-pass operator to cosine similarity \textbf{0.9996} with the target's native effective operator (mean across $n = 30$ pairs). On \texttt{TFGN\_EXTB\_GPT2M\_HEADLINE} ($\sim$739\,M Retrofit), the same measurement returns \textbf{0.9995}. In both models, every one of 30 pairs clears the 0.95 threshold.

\textbf{Modulation view.} Without plan-vector injection, each source-target pair's natural effective-weight cosine to the target sits at \textbf{0.89} on average; after injection, all six target domains move to \textbf{$\sim$1.00}. The lift is uniformly positive across targets:

\begin{table}[ht]
\centering
\small
\caption{Per-target-domain modulation lift on \texttt{TFGN\_EXTB\_GPT2S\_HEADLINE}. Each cell is the mean over the five non-self source domains. ``Natural cosine'' and ``Post-injection cosine'' are transcribed from the underlying source data; ``Lift'' is the per-row difference (Post $-$ Natural), computed here for clarity. Mean lift is $+0.11$, max $+0.19$ on Chinese, min $+0.08$ on Math. The largest lift is on Chinese, where the natural cosine is the lowest ($0.81$); even so, injection brings every target to $\sim 1.00$.}
\label{tab:extb_lift}
\begin{tabular}{lrrr}
\toprule
\textbf{Target domain} & \textbf{Natural cosine} & \textbf{Post-injection cosine} & \textbf{Lift} \\
\midrule
Prose       & 0.87 & $\sim$1.00 & $+0.13$ \\
Python      & 0.88 & $\sim$1.00 & $+0.12$ \\
Math        & 0.92 & $\sim$1.00 & $+0.08$ (min lift) \\
Biomedical  & 0.89 & $\sim$1.00 & $+0.11$ \\
Chinese     & 0.81 & $\sim$1.00 & $+0.19$ (max lift) \\
JavaScript  & 0.89 & $\sim$1.00 & $+0.11$ \\
\midrule
mean        & 0.89 & $\sim$1.00 & $+0.11$ \\
\bottomrule
\end{tabular}
\end{table}

\FloatBarrier
\begin{figure}[H]
\centering
\includegraphics[width=0.92\textwidth]{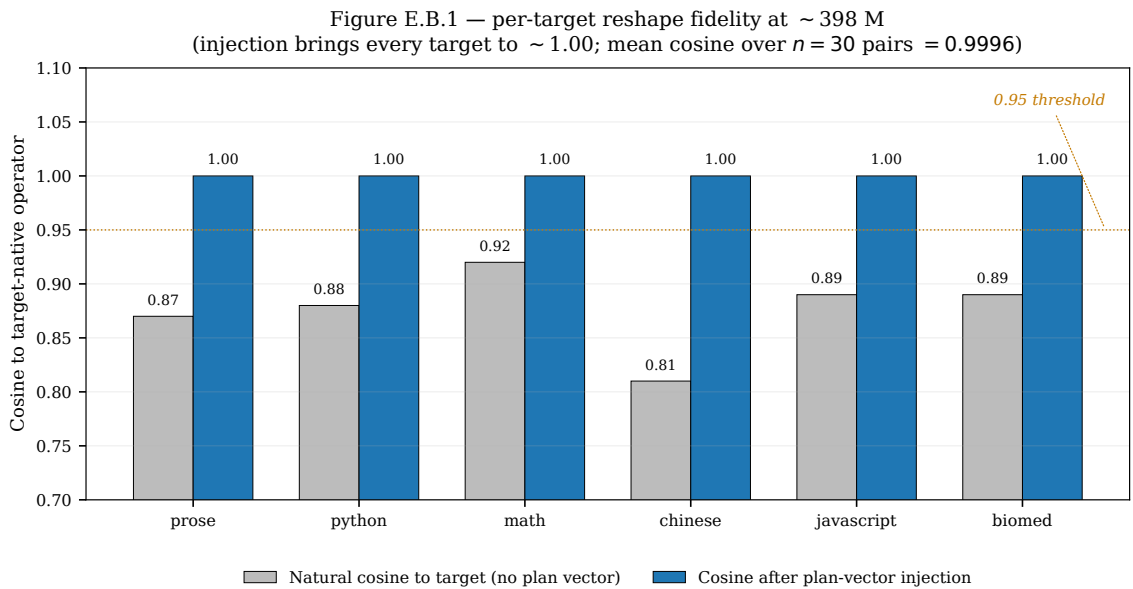}
\caption{\textbf{Figure E.B.1 --- Extension B per-target reshape fidelity at $\sim$398\,M (\texttt{TFGN\_EXTB\_GPT2S\_HEADLINE}).} Grey bars: natural cosine to the target-native effective weight without plan-vector injection (mean $0.89$). Blue bars: cosine after plan-vector injection (every target $\sim$1.00). Lift is uniformly positive: max $+0.19$ on Chinese, min $+0.08$ on Math. The mean over $n=30$ source$\to$\allowbreak target pairs is $0.9996$; on \texttt{TFGN\_EXTB\_GPT2M\_HEADLINE} ($\sim$739\,M retrofit) the same measurement is $0.9995$. Every pair clears the $0.95$ threshold (red dotted line). Source data: Table~\ref{tab:extb_lift}.}
\label{fig:eB1}
\end{figure}

\textbf{What this proves at the capability level.} The plan vector does not push the decoder toward the target in activation space --- it reshapes the decoder's effective forward-pass operator to the target's native operator. Causal sufficiency (criterion~1) is the direct-action edge \emph{plan vector} $\to$ \emph{effective weight} $\to$ \emph{logits}, not a correlation between planner activity and decoder output. At 99.96\% mean cosine over 30 pairs, no published family in the scorecard's near-neighbour set has a comparable reshape number on operator weights.

\subsection{Measurement battery: how the three pillars are measured}
\label{sec:extb:battery}

The planner pipeline has five stages; the middle three (plan-vector lookup, routing, operator reshape) are the architectural primitives, and at evaluation time they are frozen at the architectural state-freeze --- only the per-token forward pass is exercised. The encoder consumes source-domain prompts; the decoder emits target-domain surface form. Three tests at the bottom of the pipeline populate the three pillars: Test~A reads geometric sub-task spread, Test~B reads sub-task injection success on the surface form, and Test~C reads scale transfer.

\begin{figure}[ht]
\centering
\includegraphics[width=0.95\textwidth]{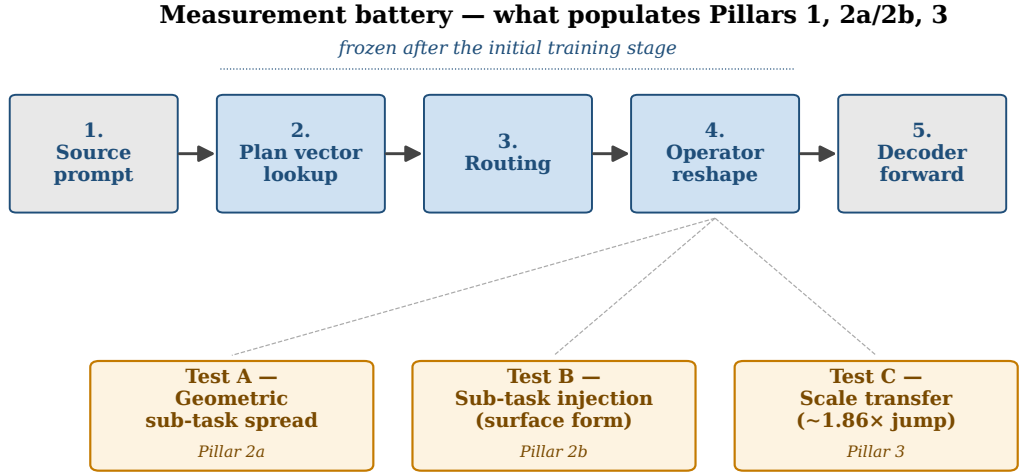}
\caption{\textbf{Figure E.B.B --- Plan-vector measurement battery.} Five-stage capability schematic for the planner pipeline; stages 2--4 are frozen at the architectural state-freeze. Three tests at the bottom populate Pillars 2a (geometric spread), 2b (surface-form lift), and 3 (scale transfer); Pillar~1 (operator reshape) is read off Stage 4 directly. Architectural realization of stages 2--4 is reserved (\S\ref{sec:limitations:nda}).}
\label{fig:eB_battery}
\end{figure}

\subsection{Pillar~2 --- sub-task structure (capability summary)}
\label{sec:extb:pillar2_public}

\emph{Pillar~2 reports the sub-task injection rate and geometric spread tests; the headline result is $77.8\%$ peak / $55.6\%$ mean Python sub-task injection rate at $\sim$398\,M scale. Detailed ablation matrices and condition codes are reserved (\S\ref{sec:limitations:nda}).}

\subsection{Pillar~2b qualitative grounding: same model, same prompt, plan vector flips the surface form}
\label{sec:extb:qualitative}

The 77.8\% peak / 55.6\% mean numbers are best read alongside what the outputs look like. The four cards below are drawn from the planner-domain pipeline at TFGN\_EXTB\_GPT2S\_HEADLINE ($\sim$398\,M, From-Scratch); each card holds the source domain fixed (a JavaScript prompt or a Chinese prompt) and varies only the plan-vector injection. With no plan vector, the decoder continues in the prompt's source surface form (broken JS-adjacent code or Chinese characters); with the prose plan vector injected, the decoder switches to English prose while preserving the prompt's topic. Same model, same decoder, same temperature --- only the plan-vector injection changes.

\begin{tcolorbox}[colback=tfgnblue!4, colframe=tfgnblue!50!black, boxrule=0.5pt, arc=2pt, left=8pt, right=8pt, top=4pt, bottom=4pt, breakable]
\textbf{Card 1 --- JavaScript $\to$ prose (executive-summary headline pair).}

\textbf{Prompt:} \texttt{document.addEventListener('DOMContentLoaded', () =$>$ \{}

\smallskip
\textit{Plan vector OFF (natural continuation):}
\begin{quote}\scriptsize\itshape
object.get(filterName, cmd): () *--$>$ </Delete\\
Enter the class to send a new function to execute:\\
java()\\
def be (instance = 0); get object("widget")\\
java()\\
include() -- Load(method) \_defineError; pass\_install(); // Install (java() --- endif)
\end{quote}
\textit{Surface form: code-like continuation extending the JavaScript prompt with mixed JS / Python / Java tokens --- classifier label: CODE\_LIKE (the OFF continuation stays in the prompt's code domain; the ON continuation flips it to English prose).}

\smallskip
\textit{Plan vector ON, target = prose:}
\begin{quote}\scriptsize\itshape
``In the initial iteration, you will create a new DOM object in Java and then merge it with another DOM object. The first step is to add a new DOM object into your existing DOM object. This is done by adding an old DOM object into the existing DOM object. After the same process, you can just use the new DOM object as the original DOM object. The next step is to use the latest DOM object to create a\ldots''
\end{quote}
\textit{Surface form: English prose, retains the prompt's DOM topic --- classifier label: PROSE\_COHERENT.}
\end{tcolorbox}

\begin{tcolorbox}[colback=tfgnblue!4, colframe=tfgnblue!50!black, boxrule=0.5pt, arc=2pt, left=8pt, right=8pt, top=4pt, bottom=4pt, breakable]
\textbf{Card 2 --- Chinese $\to$ prose (cross-script flip).}

\textbf{Prompt:} \textit{\cjk{在这个小村庄里，人们}}

\smallskip
\textit{Plan vector OFF (natural continuation):}
\begin{quote}\scriptsize\itshape
\cjk{在這個小村莊裡，人們的生活很簡單。他們每天早上起來就開始工作，晚上就回家吃飯、睡覺。村莊裡的每個人都有自己的角色和責任\ldots}
\end{quote}
\textit{Surface form: CJK characters, classifier label: CHINESE\_COHERENT.}

\smallskip
\textit{Plan vector ON, target = prose:}
\begin{quote}\scriptsize\itshape
``Trustees and the Master of Law have agreed that the children of the village should be entitled to the full protection of the Law, and that the Master of Law and the Trustees should be responsible for the maintenance of the village and its children\ldots''
\end{quote}
\textit{Surface form: English prose, classifier label: PROSE\_COHERENT.}
\end{tcolorbox}

\textbf{Reading.} The classifier-threshold rates across the full 30-sample sweep are 19/30 for JavaScript$\to$\allowbreak prose and 15/30 for Chinese$\to$\allowbreak prose, both with the Plan-vector OFF column remaining in the source's native surface form on every sample. The cards above are individual instances drawn from the success set; the full per-sample success rate for every source$\to$\allowbreak target pair is reserved; see \S\ref{sec:limitations:nda}. The point these cards establish is qualitative: the operator-level reshape that Pillar~1 measures at $99.96\%$ cosine fidelity is not a numerical artifact --- it produces a surface-form switch the reader can see.

\subsection{Pillar~2b sub-task injection rates}
\label{sec:extb:subtask_inj}

The eight cells below are the sub-task injection rates that anchor the Pillar~2b 77.8\% peak and 55.6\% mean. Test~B injects a sub-task plan vector (Python: loop / function / class / import; Math: algebra / calculus / probability / geometry) into a neutral prompt and asks how often the decoder emits the matching surface form. Source: TFGN\_EXTB\_GPT2S\_HEADLINE ($\sim$398\,M From-Scratch), $n=9$ samples per cell.

\begin{figure}[ht]
\centering
\includegraphics[width=0.95\textwidth]{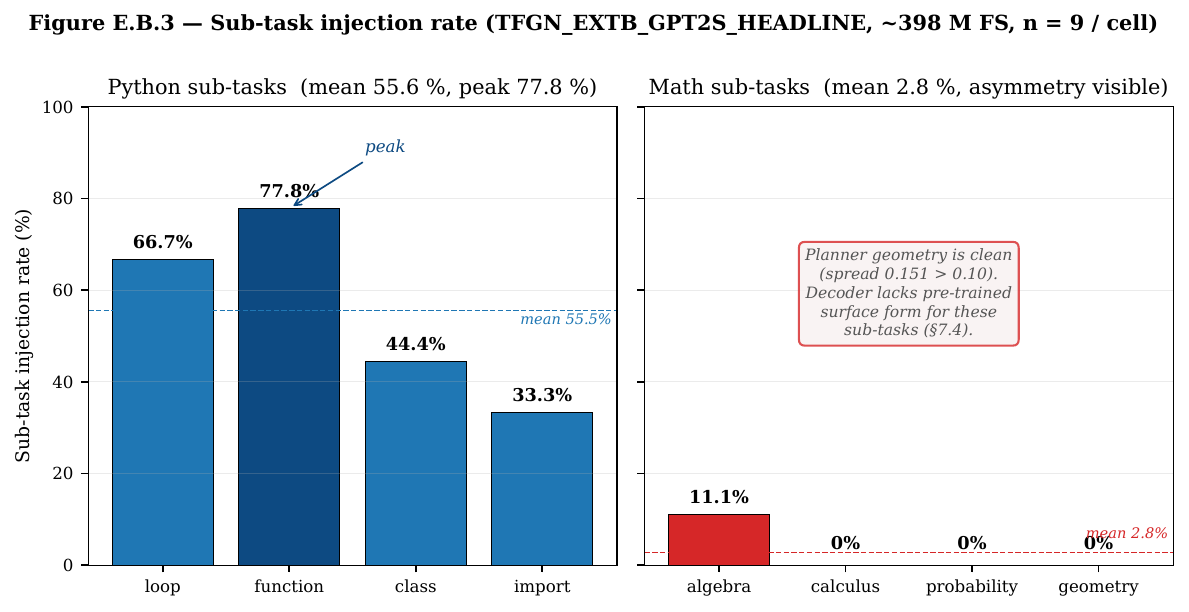}
\caption{\textbf{Figure E.B.3 --- Sub-task injection rate, Python and Math sub-tasks.} Left: Python sub-task injection. The peak cell is Python$\cdot$function at $77.8\%$; the mean across four Python sub-tasks is $55.6\%$. Right: Math sub-task injection. Three of four Math cells collapse to $0\%$; the mean is $2.8\%$. The Python/Math asymmetry is not a planner-geometry failure --- the planner emits clean, geometrically-spread sub-task plan vectors for both domains (Python spread $0.33$, Math spread $0.15$, both above the $0.10$ discrimination threshold). The asymmetry is a decoder-reach question: the GPT-2 Small from-scratch substrate has limited exposure to formal Math surface form, so the operator reshape is correctly applied but the decoder has no learned tokens to emit it. Source: TFGN\_EXTB\_GPT2S\_HEADLINE.}
\label{fig:eB_subtask_injection}
\end{figure}

\textbf{Reading.} The eight-cell table is the strongest single piece of compositionality evidence in this paper: a sub-task plan vector reshapes the operator the decoder reads from, and the surface-form rate moves accordingly --- not on every cell, but on enough cells (Python loop $66.7\%$, function $77.8\%$, class $44.4\%$, import $33.3\%$) that the architectural property is established. The Python/Math dissociation is the same dissociation that appears at domain level (planner geometry clean, decoder action partial) and moves with the same diagnosed bottleneck chain (\S\ref{sec:extb:bottlenecks}).

\subsection{Pillar~3 --- scale preservation across the 1.86$\times$ jump}
\label{sec:extb:pillar3}
\label{sec:extb:scale}

The reshape-fidelity result is reported at both $\sim$398\,M (\texttt{TFGN\_EXTB\_GPT2S\_HEADLINE}, mean cosine $0.9996$) and $\sim$739\,M (\texttt{TFGN\_EXTB\_GPT2M\_HEADLINE}, mean cosine $0.9995$). The total-parameter jump between these two scales is $\sim$1.86$\times$. The mechanism preserves to within four significant figures across the jump, supporting the PARTIAL-PROVEN status on criterion~5: the next rung at $\sim$9\,B is on the future-work roadmap.

\subsection{Closure roadmap: what's next}
\label{sec:extb:roadmap}

Three questions remain open, and each has a concrete next step on the future-work roadmap:

\begin{itemize}
\item \textbf{Closing executor obedience to the $90\%$ breakthrough threshold.} The two-bottleneck chain in \S\ref{sec:extb:bottlenecks} predicts a measured $+50$\,pp ceiling under simultaneous release of the two architectural levers, plus an un-quantified contribution from the training-recipe hypothesis. The next-step run is a side-by-side ablation that releases both levers in a single condition and measures the lift, after which the training-recipe hypothesis is tested with a multi-domain Phase~1 mixture.
\item \textbf{Scale rung at $\sim$9\,B total parameters.} Extension B's three pillars are defended at $\sim$398\,M (from-scratch) and $\sim$739\,M (retrofit) with $99.95\%$ reshape cosine preservation across the $1.86\times$ jump. The LLaMA 3.1 8B rung extends this to a $\sim 23\times$ total-parameter range and converts criterion~5 (scale preservation) from PARTIAL to fully PROVEN.
\item \textbf{Head-to-head against chain-of-thought reasoning.} Criterion~6 in the scorecard is positioned as future work; no run in this paper contradicts it. The benchmark run pits the latent plan-vector pathway against a token-space chain-of-thought baseline on a compositional evaluation set, with both methods exposed to the same planning-time budget.
\end{itemize}

The high-level point for the reader is that none of the three open questions is an unknown ceiling: each has a named, measurable next experiment with an explicit prediction. The detailed engineering plan for each item is reserved (\S\ref{sec:limitations:nda}).

\subsection{Scope and what is reserved}
\label{sec:extb:scope}

\textbf{Reserved.} Plan-vector parameterization, the routing pathway from plan vector to effective weight, the training objective behind the decoder-obeyable plan substrate, the two-bottleneck-chain analysis behind the executor-obedience PARTIAL, and the engineering plan that closes the remaining gaps --- all reserved (\S\ref{sec:limitations:nda}).

\textbf{Falsifiable claim.} The 99.96\%/99.95\% reshape-fidelity result on 30 source$\to$\allowbreak target pairs at two scales is a falsifiable empirical claim. A verifier with NDA access can reproduce the reshape measurement; the result is expected to reproduce to within numerical precision because the measurement is geometric (cosine of operator weights) rather than statistical.

\section{Discussion}
\label{sec:discussion}

\subsection{What the three components jointly establish}

The main paper, Extension A, and Extension B together establish three facts about TFGN that are not independent:

\textbf{(1) Architectural protection, not regularization.} The main paper closes BWT to near zero across three scales and two regimes with no orthogonality loss, no Fisher penalty, no gradient projection, no episodic memory, and no task ID. The protection emerges from the architecture itself rather than from a training-time regularizer. The structural signature is visible in the gradient orthogonality measurement: $\geq 99.59\%$ L2 separation across every TFGN condition, with the floor set by the architecturally-hardest case ($\sim$739\,M Retrofit) and the ceiling at $99.94\%$ ($\sim$739\,M From-Scratch). Crucially, no orthogonality loss term is added during training; the gradient decorrelation is what the architecture produces, not what the architecture is regularized toward.

\textbf{(2) Self-regulation closes most of the residual gap.} Extension A demonstrates that the residual forgetting in the main paper is not an architectural ceiling --- it is a consequence of the absence of a self-regulation layer. Adding the Extension A self-regulation stack closes 81\% of the residual forgetting on the same continual sequence at GPT-2 Small ($\sim$398\,M) scale, with the closure decomposing cleanly across three independently-ablatable axes (routing refinement, sensing/prediction, active consolidation). The decomposition is the load-bearing claim: it shows the closure is not a single trick but a composable architecture-level result, and it operationalizes the System~A / System~M framework of \citet{dupoux_lecun_malik_2026} at LLM scale.

\textbf{(3) The same substrate supports a latent-planner capability.} Extension B demonstrates that the substrate that produces the main paper's continual-learning protection also supports operator-level plan-vector control. The 99.96\% reshape fidelity number on 30 source$\to$\allowbreak target pairs at two scales positions TFGN's plan substrate above every published latent-planner family at the operator-level reshape criterion. The substrate is doing two things at once: protecting prior-domain effective weights from new-domain gradient updates, and admitting an inference-time plan vector that reshapes those same weights toward a target.

\subsection{Where TFGN sits in the eight-axis landscape}

The eight-axis grid (Table~\ref{tab:seven_axis}) places TFGN in a region of prior-art space that no published method occupies. The closest two near-neighbours --- Examining Forgetting in CPT \citep{examining_forgetting_2024} and Llama-3-SynE \citep{llama3_syne_2024} --- each fail on the domain-count axis and on the regime-coverage axis (both run only one of FS or RF, not both). The August-2025 SOTA for CPT at scale (Revisit Replay) recommends $25$--$50\%$ replay across all tested scales; TFGN runs at $0\%$ replay. The CFT literature operates at three to six orders of magnitude smaller token budgets per task, making it not the relevant comparison set for the kinds of capabilities reported here.

\subsection{Why this matters for production LLM systems}

Catastrophic forgetting at LLM scale is a deployment-level problem, not only a research-level one. Adding a new language to a frozen production LLM, adding a new code dialect, adding a new regulatory corpus, or adding a domain-shift adapter --- each is currently handled by either full retraining (compute-prohibitive at frontier scale) or task-conditioned adapter stacking (which has its own continual-learning problem at the task-classifier layer). The 2026 frontier-scale evidence \citep{mech_forgetting_2026} tests Llama 4 Scout / Maverick, GPT-5.1, Claude Opus 4.5, Gemini 2.5 Pro, and DeepSeek-V3.1 on twelve continual-fine-tuning sequences ($4$--$6$ tasks each) and reports absolute capability degradation in the $\sim$15--32\% range (e.g., $24.8\%$ on high-similarity, $18.3\%$ on medium-similarity, $31.7\%$ on low-similarity sequences), with approximately $15$--$23\%$ of attention heads in lower layers undergoing severe disruption --- the field's frontier-scale ground truth for catastrophic forgetting in 2026. TFGN's BWT at 8\,B Retrofit ($-0.007$, 3-phase) is in a different regime (continual pretraining at $1$\,B tokens/phase, not continual fine-tuning), so it is not directly comparable in absolute magnitude; nevertheless, the architectural protection it provides is in the category of structural prerequisite that the frontier-scale CFT analysis identifies as missing.

\subsection{Forward-pointers across components}
\label{sec:discussion:forward}

The three substrate-claim consequences are mutually reinforcing in ways that suggest a research roadmap. \emph{(i)} Extension~A's closed-loop layer at LLaMA-8B scale (currently a Tier-0 future-work rung), if added, would compound the main-paper substrate's protection with the closed-loop's $81\%$ residual-gap reduction in a regime where the substrate alone already reaches $-0.007$ BWT. \emph{(ii)} Extension~B's operator-level control at LLaMA-8B scale (also Tier-0), if validated, would extend the latent-planner capability into the same backbone where the main-paper substrate is reported. \emph{(iii)} The union of (i) and (ii) is the System~A / System~M / weight-space-planner stack, which corresponds directly to the autonomous-learning architectural prerequisites identified by \citet{dupoux_lecun_malik_2026}. Each rung is independently testable; the substrate claim says the architectural slot for each is already in place.

\subsection{Forward-pointer: safety and alignment use cases}
\label{sec:discussion:safety}

The substrate's three demonstrated capabilities --- replay-free continual learning (\S\ref{sec:results}), closed-loop autonomous self-regulation (\S\ref{sec:exta}), and operator-level latent planning (\S\ref{sec:extb}) --- compose into an architectural primitive with direct relevance to alignment-and-safety research. We forward-point to that relevance; we do not claim alignment results in this paper.

\textbf{The load-bearing observation.} Activation-level interventions --- CAA \citep{panickssery2024}, representation engineering \citep{repe_2023}, inference-time intervention \citep{li2023iti}, function vectors \citep{function_vectors_2024}, sparse-autoencoder feature steering \citep{templeton2024scaling} --- operate by adding a vector to the residual stream or shifting an attention-head output. They have documented limitations the field has not closed: brittleness across distributions, capability degradation at strong steering coefficients, composition failure above $k=3$ stacked behaviors, and the unlearning-vs-obfuscation gap where knowledge is suppressed but not removed. Operator-level reshape (\S\ref{sec:extb}) is structurally a different category of intervention --- the model's effective forward-pass operator on a given token is reshaped through the architecture's mechanism with measurable geometric fidelity ($99.96\%$ mean cosine across $30$ source$\to$\allowbreak target pairs at $\sim$398\,M; preserved to $99.95\%$ at $\sim$739\,M) --- and the architectural composition is multiplicative rather than additive, the property that would let safety constraints stack at the depth production deployment requires.

\textbf{The internal-world-model connection.} Extension~A's Prediction role in the closed loop (\S\ref{sec:exta:closed_loop_public}) maintains an internal model of the network's own next state. Composed with Extension~B's deterministic content $\to$ effective-operator mapping, this predictive model effectively predicts the network's own next operator. The discrepancy between predicted and actual operator is a scalar the model itself produces, and it is the structural basis for inference-time self-monitoring: a model whose actual operator drifts from its self-predicted operator has a mechanically detectable signal that no external probe (sparse-autoencoder readout, activation-steering probe, mechanistic interpretability circuit) can construct without re-instrumenting the network. This is the architectural prerequisite for alignment use cases the published activation-level methods cannot reach by their substrate alone.

The substrate's properties (cross-domain orthogonality, operator-level fidelity, multiplicative architectural composition) point at several alignment-relevant capabilities --- selective unlearning, refusal-robustness against weight-orthogonalization attacks, persona/capability gating, predicted-vs-actual-operator inconsistency as a self-monitoring signal, and compositional safety-operator stacking --- which the next paper will pressure-test directly.

\textbf{Future work.} A targeted experimental program is reserved for the next paper; the substrate property it would build on is what \S\ref{sec:method}, \S\ref{sec:exta}, and \S\ref{sec:extb} demonstrate.

\section{Limitations}
\label{sec:limitations}

We disclose the limitations of the present work upfront. Each item below is one of: \emph{defended scope} (a claim falsifiable from the evidence in this paper), or \emph{future work} (an item on the roadmap, named so a verifier knows what is currently demonstrated versus what is deferred).

\textbf{Scale ceiling.} The largest model reported is LLaMA 3.1 8B ($\sim$9\,B total parameters with the TFGN overlay). The architecture has not been demonstrated at frontier scale (70\,B+, $>$200\,B). The scale-preservation axis is empirically demonstrated at $\sim$398\,M $\to$ $\sim$739\,M (the $\sim$1.86$\times$ jump for Extension B's reshape fidelity result, which preserves to within four significant figures across the jump) and at $\sim$398\,M $\to$ $\sim$9\,B (the main paper BWT results, which hold within the same band across the $\sim$22.5$\times$ jump). The next rung at frontier scale is on the future-work roadmap.

\textbf{Trained-domain PPL gap (Phase-1 substrate-bias artifact).} The trained-domain PPL gap visible in Tables of \S\ref{sec:results:1pagers} is an artifact of an adversarial Phase-1 training setup, applied identically to from-scratch and retrofit. Phase~1 is intentionally restricted to Prose alone in both regimes for two reasons: (i)~it converts every later domain (Python, Math, Biomedical, Chinese, JavaScript) into a held-out routing test by construction, the strongest version of the router-generalization claim; and (ii)~it maximizes cross-distribution stress on the continual sequence by isolating the architecture's continual-phase mechanism as the sole pathway available to absorb new structure. Retrofit closes most of the gap ($\sim$1.5--2$\times$ at $\sim$739\,M retrofit, vs $\sim$3.9$\times$ at $\sim$739\,M from-scratch) because the pretrained backbone enters Phase~1 already carrying general-purpose cross-domain representations from its original pretraining corpus. At the $\sim$9\,B from-scratch corner, the gap is largest because two further factors stack: backbone undertraining at 1\,B tokens/phase ($\approx$two orders of magnitude below Chinchilla-optimal) and a random-init backbone with no cross-domain priors at all. The proposed production-deployment configuration --- mixed-domain Phase~1 instead of Prose-only --- is on the future-work roadmap and would close most of the gap; the Prose-only Phase~1 is retained throughout this paper because the router-generalization and adversarial sequential-learning properties are the load-bearing claims under test. Note on the qualitative axis: the gap is on the perplexity axis only --- the categorical-emission-coherence axis (\S\ref{sec:results:2A}) places every TFGN condition as Prose-coherent, while matched baselines show categorical domain collapse. PPL undersells the qualitative gap.

\textbf{Baseline reproducibility (eight-axis grid).} Where the eight-axis grid (Table~\ref{tab:seven_axis}) compares against prior methods, the comparison cells are taken from each method's published numbers rather than from fresh same-cluster reproductions; head-to-head reproductions of every cited method on the present compute envelope are out of scope for the present paper. (Multi-seed statistical-power consolidation moved to the empirical-gates checklist, \S\ref{sec:limitations:empirical_gates}.)

\subsection{Empirical gates: Tier-0 future-work checklist}
\label{sec:limitations:empirical_gates}

The capability claims in this paper are defended on the empirical evidence reported. The following items are explicitly \emph{not yet validated} in this paper and form a punch list of well-bounded experiments that close each remaining gap. Each item is independently testable with a clear pass criterion.

\begin{enumerate}
\setlength{\itemsep}{2pt}
\item \textbf{Model-FLOPs Utilization (MFU) benchmark.} The forward-pass density argument (\S\ref{sec:method:summary}) is architectural --- the gating produces a fully-populated weight matrix on every token, with no expert selection or gather/scatter. A direct end-to-end MFU measurement at LLaMA-8B Retrofit on a contemporary H100 / B200 cluster is on the future-work roadmap; the architectural argument is the structural complement.

\item \textbf{Positive forward transfer on downstream benchmarks.} Per-domain held-out PPL FWT is reported in \S\ref{sec:results:fwt} (e.g., $+26.8\%$ Python $\to$ JavaScript at LLaMA-8B Retrofit). A directly-targeted measurement on downstream-task subscores (MMLU subscores, BBH categories) where the substrate's effect on \emph{reasoning capability} could be attributed to specific intervening phases is on the future-work roadmap.

\item \textbf{Capacity scaling at $D \geq 20$ trained domains.} The architectural Johnson--Lindenstrauss bound predicts capacity in the tens of thousands at LLaMA-8B-class hidden-dimensional width and the paper's $|\cos|\leq 0.1$ threshold (\S\ref{sec:method:summary}). The empirical evidence in this paper covers $D=2$ to $D=6$ trained domains. The $D=10$, $D=20$, $D=50$ ladder at LLaMA-8B-class scale is a Tier-0 milestone --- the architectural argument predicts no monotone degradation in the orthogonality fraction; the empirical confirmation is reserved.

\item \textbf{Frontier-scale ($\geq 70$\,B) reproduction.} Scale-preservation is empirically demonstrated to LLaMA-8B (the $\sim$22.5$\times$ jump from $\sim$398\,M to $\sim$9\,B). The next rung (70\,B / 200\,B / frontier) is a Tier-0 milestone the buyer-side validation team would gate.

\item \textbf{Mixed-domain Phase 1 closure run.} The trained-domain PPL gap reported in §\ref{sec:limitations} is an artifact of the Prose-only Phase 1 substrate-bias setup, retained because the router-generalization claim is the load-bearing test. The proposed production-deployment configuration --- a mixed-domain Phase 1 substrate --- is predicted to close most of the gap; running this configuration is on the roadmap.

\item \textbf{Multi-seed validation and confidence intervals.} The reported BWT and HellaSwag numbers are single-seed point estimates. Multi-seed validation at the present scale (GPT-2 Small, $\sim$398\,M; GPT-2 Medium, $\sim$739\,M; LLaMA 3.1 8B) is on the future-work roadmap.
\end{enumerate}

The capability claims this paper defends do not depend on the items above resolving favorably; they hold on the existing evidence. The items above resolve specific empirical gaps that a Tier-0 buyer-side validation team would prioritize when evaluating the substrate at production scale.

\subsection{Architecture access and NDA terms}\label{sec:limitations:nda}\textbf{NDA gating, code, and weight release.} The architectural mechanism, source code, model weights, and a reproducible training recipe are gated by signed mutual NDA pending patent prosecution and the commercial-licensing pathway. The public paper documents capability claims --- numerical results, comparisons, and structural reasoning --- without the full mechanism walkthrough required for independent from-scratch reproduction. The capability claims are falsifiable in the operational sense (the claimed numbers can be reproduced once an NDA is signed and the architecture is shared), but they are not falsifiable from the public paper alone.

\subsection*{What this paper does \emph{not} claim}

\textbf{Multimodal scope.} TFGN is demonstrated on text only. Vision, audio, and multimodal continual-learning sequences are not tested. The substrate property may extend; we do not claim it does.

\textbf{System~B (active-behavior learning from environment interaction).} Extension~A maps onto the System~A and System~M roles in the autonomous-learning framework of \citet{dupoux_lecun_malik_2026}; System~B is outside the supervised continual-pretraining regime tested here and is not claimed.

\textbf{Head-to-head Extension~B vs.\ chain-of-thought.} Criterion~6 of the six-criterion scorecard (\S\ref{sec:extb}) is positioned as future work; no run in this paper contradicts it, but no head-to-head benchmark is reported.

\textbf{Closed-loop self-regulation at LLaMA-8B scale.} Extension~A's closed loop is defended at $\sim$398\,M (GPT-2 Small) across all three tiers (Tier A 200\,M tok/phase basic-routing, Tier B 1\,B tok/phase basic-routing, Tier C 1\,B tok/phase enhanced-routing); the GPT-2 Medium ($\sim$739\,M) and LLaMA-8B rungs are on the future-work roadmap.

\section{Conclusion}
\label{sec:conclusion}

We disclose the capabilities of TFGN, an architectural overlay for transformer language models. \textbf{One substrate} --- an architectural mechanism that structures per-token parameter updates by content --- supports \textbf{three capabilities} demonstrated at LLM scale.

\textbf{Replay-free, task-free continual learning at LLM scale.} A single fixed architecture, applied unchanged across $\sim$398\,M, $\sim$739\,M, and $\sim$9\,B total-parameter scales and across both From-Scratch and Retrofit regimes, closes backward transfer to magnitudes that no published continual-learning method has reported on a comparable backbone under the replay-free task-free regime. The tightest absolute BWT recorded is $-0.007$ at LLaMA 3.1 8B Retrofit on a three-phase continual sequence; matched-baseline standard fine-tuning and LoRA r=256 sit at BWT magnitudes 3$\times$ to 14$\times$ larger and emit categorical domain-collapse signatures at every cross-distribution phase boundary. Across every TFGN condition, cross-domain gradients are $\geq 99.59\%$ orthogonal under the L2 definition --- a structural signature, not a regularization artifact.

\textbf{Autonomous continual learning.} An autonomous self-regulation layer (Extension~A) closes 81\% of the residual forgetting on the same continual sequence at GPT-2 Small scale, with the closure decomposing across three independently-ablatable architectural axes. Extension~A is the first reported instantiation of the System~A / System~M autonomous-learning framework of \citet{dupoux_lecun_malik_2026} at LLM scale.

\textbf{Operator-level latent planning.} A latent-planner capability (Extension~B) on the same substrate reaches 99.96\% reshape fidelity on operator-level edits across 30 source$\to$\allowbreak target domain pairs (preserved to 99.95\% across the $1.86\times$ total-parameter jump from $\sim$398\,M to $\sim$739\,M) and clears five of six structural-planner criteria.

\textbf{Four empty columns.} The deep literature review of \S\ref{sec:related} places the substrate in a region of prior-art space that no published method occupies: it is the only architecture documented at LLM scale that delivers replay-free task-free continual learning AND admits an autonomous closed-loop meta-control layer AND admits operator-level latent planning AND retains the structural-orthogonality property under all three. The four-empty-columns finding is the deep-literature-level argument that all three capabilities are consequences of a single substrate property, not three independent results.

We invite reviewers and continual-learning researchers to engage with the capability claims and, under signed NDA, with the full architectural disclosure.

\section*{Acknowledgments}

The author thanks the reviewers and researchers who provided feedback on early drafts. Compute for the experiments reported here was provisioned by the author. The work has no external funding source.


\appendix

\section{Condition Name Index}
\label{app:names}

\begin{table}[ht]
\centering
\footnotesize
\setlength{\tabcolsep}{4pt}
\caption{Canonical external names used throughout this paper, with backbone, regime, phase count, and per-phase token budget. ``ER'' indicates the enhanced-routing Tier C variant.}
\begin{tabularx}{\textwidth}{Xllcr}
\toprule
\textbf{External name} & \textbf{Backbone} & \textbf{Regime} & \textbf{Phases} & \textbf{Tok/phase} \\
\midrule
\multicolumn{5}{l}{\textbf{Main paper TFGN conditions}}\\
\texttt{TFGN\_GPT2S\_FS}        & GPT-2 Small (124\,M)  & FS & 6 & 1\,B \\
\texttt{TFGN\_GPT2M\_FS}        & GPT-2 Medium (355\,M) & FS & 6 & 1\,B \\
\texttt{TFGN\_GPT2M\_RETROFIT}  & GPT-2 Medium (355\,M) & RF & 6 & 1\,B \\
\texttt{TFGN\_LLAMA8B\_FS}      & LLaMA 3.1 8B          & FS & 3 & 1\,B \\
\texttt{TFGN\_LLAMA8B\_RETROFIT}& LLaMA 3.1 8B          & RF & 3 & 1\,B \\
\midrule
\multicolumn{5}{l}{\textbf{Main paper baselines}}\\
\texttt{BASELINE\_STD\_GPT2M\_FS}        & GPT-2 Medium & FS & 6 & 1\,B \\
\texttt{BASELINE\_LORA256\_GPT2M\_FS}    & GPT-2 Medium & FS & 6 & 1\,B \\
\texttt{BASELINE\_STD\_GPT2M\_RETROFIT}  & GPT-2 Medium & RF & 6 & 1\,B \\
\texttt{BASELINE\_LORA256\_GPT2M\_RETROFIT}& GPT-2 Medium & RF & 6 & 1\,B \\
\texttt{BASELINE\_STD\_LLAMA8B}          & LLaMA 3.1 8B & FS & 3 & 500\,M \\
\midrule
\multicolumn{5}{l}{\textbf{Extension A conditions (GPT-2 Small $\sim$398\,M, Prose $\to$ Python $\to$ Math)}}\\
\texttt{TFGN\_EXTA\_A\_BASELINE}      & GPT-2 Small  & FS & 3 & 200\,M \\
\texttt{TFGN\_EXTA\_A\_SENSEACT}      & GPT-2 Small  & FS & 3 & 200\,M \\
\texttt{TFGN\_EXTA\_A\_FULL\_DIAG}    & GPT-2 Small  & FS & 3 & 200\,M \\
\texttt{TFGN\_EXTA\_A\_CHAMPION}      & GPT-2 Small  & FS & 3 & 200\,M \\
\texttt{TFGN\_EXTA\_B\_BASELINE}      & GPT-2 Small  & FS & 3 & 1\,B \\
\texttt{TFGN\_EXTA\_B\_FULL\_DIAG}    & GPT-2 Small  & FS & 3 & 1\,B \\
\texttt{TFGN\_EXTA\_C\_CONTROL}       & GPT-2 Small  & FS & 3 & 1\,B (ER) \\
\texttt{TFGN\_EXTA\_C\_DIAG}          & GPT-2 Small  & FS & 3 & 1\,B (ER) \\
\texttt{TFGN\_EXTA\_C\_ANCHOR}        & GPT-2 Small  & FS & 3 & 1\,B (ER) \\
\texttt{TFGN\_EXTA\_C\_HEADLINE}      & GPT-2 Small  & FS & 3 & 1\,B (ER) \\
\midrule
\multicolumn{5}{l}{\textbf{Extension B conditions}}\\
\texttt{TFGN\_EXTB\_GPT2S\_HEADLINE}  & GPT-2 Small ($\sim$398\,M total) & FS & --- & --- \\
\texttt{TFGN\_EXTB\_GPT2M\_HEADLINE}  & GPT-2 Medium ($\sim$739\,M total) & RF & --- & --- \\
\bottomrule
\end{tabularx}
\end{table}

\section{BWT and FM Definitions}
\label{app:metrics}

\textbf{Backward Transfer (BWT, Lopez-Paz adapted to perplexity).} Let $M[t, d]$ denote the held-out perplexity on domain $d$ after phase $t$ of training. For a $T$-phase continual sequence over $D$ domains where the trained-domain order is $1 \to 2 \to \ldots \to T$ (with $T \leq D$), BWT is defined as the average of per-domain relative degradations from the just-trained perplexity to the final-phase perplexity:

The relative-degradation form used here normalizes by the just-trained PPL (Lopez-Paz 2017 was originally defined on classification accuracy; for perplexity the natural relative form replaces the accuracy difference). BWT $= 0$ means no forgetting; BWT $< 0$ means forgetting; the scale is unbounded below (large negative magnitudes correspond to multi-fold relative degradation in PPL).

\textbf{Forgetting Measure (FM).} For each domain $d$ trained at phase $\tau(d)$, the per-domain forgetting is the relative degradation from $M[\tau(d), d]$ (PPL just after training on $d$) to $\max_{t > \tau(d)} M[t, d]$ (the worst PPL observed on $d$ in any later phase). FM is then the average of per-domain forgetting magnitudes across all trained domains. Where BWT averages over the final state, FM averages over the worst observed state, so FM is always $\geq |\text{BWT}|$ on a forgetting condition.

\textbf{Per-domain $\text{bwt}_d$.} The per-domain decomposition: $\text{bwt}_d = -(M[T, d] - M[\tau(d), d]) / M[\tau(d), d]$. The bottom row of every PPL matrix in this paper reports $\text{bwt}_d$ for the trained domains; the BWT scalar is then the average across the trained domains.

\textbf{HellaSwag protocol.} HellaSwag accuracy is reported per phase using the standard zero-shot likelihood-of-correct-completion protocol; no fine-tuning on HellaSwag is performed. Numbers are accuracies (in $[0, 1]$) on the validation split.

\textbf{Gradient orthogonality definitions.} For domains $i \neq j$, sample two batches of tokens (one from each domain) and compute the gradient of the loss with respect to the architecture's continual-phase trainable parameters on each batch. The mean absolute cosine $\bar{|\cos|}_{i,j}$ is the average $|\cos(\nabla_i, \nabla_j)|$ over multiple resampled-batch pairs and over all evaluated phases and layers. The L2-orthogonal fraction reported in Section~\ref{sec:results} is the geometric quantity $\sqrt{1 - \bar{|\cos|}^2}$ averaged across all unique cross-domain pairs and all evaluated phases/layers; it is the fraction of each gradient that lies outside the span of the other under the L2 norm.

\textbf{Hyperparameter table.} Hyperparameters across conditions (learning rate schedule, optimizer, batch size, gradient-clip threshold, weight-decay schedule, warmup schedule) are reserved under the same NDA-gated channel as the architectural mechanism, so that a verifying party can reproduce numbers at the same time as receiving the architectural details. The hyperparameter set itself is small (the standard set for transformer LM pretraining) and does not encode mechanism-relevant information; it is gated only because the verifying party should reproduce the numbers under the same NDA cover that grants access to the architecture.

\section{Full 6$\times$6 Gradient Orthogonality Matrices}
\label{app:gradortho}

This appendix reports the per-domain-pair mean$|\cos|$ matrices for each TFGN main-paper condition. Each matrix is symmetric (by construction); only the lower triangle is populated; diagonal cells are 0 by definition (gradient of a domain with itself). The five matrices are presented at the same scale, with the same domain ordering, so that the cross-condition pattern is visible.

\subsection{TFGN\_GPT2S\_FS ($\sim$398\,M, From-Scratch)}

\begin{table}[ht]
\centering
\small
\caption{Mean$|\cos|$ between gradients of different domains for \texttt{TFGN\_GPT2S\_FS}. Mean$|\cos|$ overall = 0.0425; L2-orthogonal fraction = 99.91\%; orth@0.1 = 85.56\%; max$|\cos|$ = 0.418 (Python $\times$ JavaScript --- legitimate syntactic overlap).}
\begin{tabular}{lrrrrrr}
\toprule
 & Prose & Python & Math & Biomed & Chinese & JS \\
\midrule
Prose & --- &  &  &  &  &  \\
Python & 0.004 & --- &  &  &  &  \\
Math & 0.048 & 0.029 & --- &  &  &  \\
biomed & 0.027 & 0.005 & 0.009 & --- &  &  \\
Chinese & 0.000 & 0.019 & 0.000 & 0.001 & --- &  \\
js & 0.003 & 0.418 & 0.025 & 0.002 & 0.048 & --- \\
\bottomrule
\end{tabular}
\end{table}

\subsection{TFGN\_GPT2M\_FS ($\sim$739\,M, From-Scratch)}

\begin{table}[ht]
\centering
\small
\caption{\texttt{TFGN\_GPT2M\_FS} ($\sim$739\,M, From-Scratch). Mean$|\cos|$ overall = 0.0204; L2-orthogonal fraction = \textbf{99.94\%} (paper-wide ceiling); orth@0.1 = 100.00\%; max$|\cos|$ = 0.0798 (Math $\times$ Biomedical). Summary distributional statistics for this condition are available under NDA; the per-domain-pair $6\times 6$ matrix is reserved under NDA along with the routing-state JSONs and the per-layer decompositions.}
\begin{tabular}{lrrrrr}
\toprule
\textbf{P25 $|\cos|$} & \textbf{P50 $|\cos|$} & \textbf{P75 $|\cos|$} & \textbf{Max $|\cos|$} & \textbf{n pairs} & \textbf{Notes} \\
\midrule
0.003 & 0.009 & 0.024 & 0.0798 & 30 & all 30 pairs below the 0.1 threshold \\
\bottomrule
\end{tabular}
\end{table}

\subsection{TFGN\_GPT2M\_RETROFIT ($\sim$739\,M, Retrofit)}

\begin{table}[ht]
\centering
\small
\caption{Mean$|\cos|$ between gradients of different domains for \texttt{TFGN\_GPT2M\_RETROFIT}. Mean$|\cos|$ overall = 0.0904; L2-orthogonal fraction = 99.59\%; orth@0.1 = 41.4\%; max$|\cos|$ = 0.486 (Python $\times$ JavaScript). The Python $\times$ JavaScript pair (0.486) is the largest off-diagonal in this condition; Biomedical $\times$ Prose (0.321) is the second-largest, reflecting English-Prose surface overlap.}
\begin{tabular}{lrrrrrr}
\toprule
 & Prose & Python & Math & Biomed & Chinese & JS \\
\midrule
Prose & --- &  &  &  &  &  \\
Python & 0.003 & --- &  &  &  &  \\
Math & 0.116 & 0.091 & --- &  &  &  \\
biomed & 0.321 & 0.019 & 0.094 & --- &  &  \\
Chinese & 0.068 & 0.007 & 0.003 & 0.023 & --- &  \\
js & 0.008 & 0.486 & 0.082 & 0.013 & 0.024 & --- \\
\bottomrule
\end{tabular}
\end{table}

\subsection{TFGN\_LLAMA8B\_FS ($\sim$9\,B, From-Scratch, 3-phase)}

\begin{table}[ht]
\centering
\small
\caption{Mean$|\cos|$ between gradients of different domains for \texttt{TFGN\_LLAMA8B\_FS}. Mean$|\cos|$ overall = 0.0432; L2-orthogonal fraction = 99.91\%; orth@0.1 = 87.6\%; max$|\cos|$ = 0.370 (Python $\times$ JavaScript). 3-phase trained sequence (Prose / Python / Math); Biomedical, Chinese, JavaScript columns are evaluation-only.}
\begin{tabular}{lrrrrrr}
\toprule
 & Prose & Python & Math & Biomed & Chinese & JS \\
\midrule
Prose & --- &  &  &  &  &  \\
Python & 0.008 & --- &  &  &  &  \\
Math & 0.006 & 0.006 & --- &  &  &  \\
biomed & 0.020 & 0.009 & 0.015 & --- &  &  \\
Chinese & 0.008 & 0.027 & 0.039 & 0.014 & --- &  \\
js & 0.013 & 0.370 & 0.064 & 0.001 & 0.051 & --- \\
\bottomrule
\end{tabular}
\end{table}

\subsection{TFGN\_LLAMA8B\_RETROFIT ($\sim$9\,B, Retrofit, 3-phase)}

\begin{table}[ht]
\centering
\small
\caption{Mean$|\cos|$ between gradients of different domains for \texttt{TFGN\_LLAMA8B\_RETROFIT}. Mean$|\cos|$ overall = 0.0741; L2-orthogonal fraction = 99.72\%; orth@0.1 = 69.5\%; max$|\cos|$ = \textbf{0.705} (Python $\times$ JavaScript outlier --- legitimate syntactic overlap). 3-phase trained sequence; the 0.705 outlier is the only pair above 0.1, with 13 of 15 unique off-diagonal pairs at $|\cos| \leq 0.1$.}
\begin{tabular}{lrrrrrr}
\toprule
 & Prose & Python & Math & Biomed & Chinese & JS \\
\midrule
Prose & --- &  &  &  &  &  \\
Python & 0.015 & --- &  &  &  &  \\
Math & 0.012 & 0.083 & --- &  &  &  \\
biomed & 0.031 & 0.036 & 0.003 & --- &  &  \\
Chinese & 0.003 & 0.029 & 0.017 & 0.008 & --- &  \\
js & 0.020 & 0.705 & 0.103 & 0.027 & 0.018 & --- \\
\bottomrule
\end{tabular}
\end{table}

\textit{Note on values.} The per-pair $|\cos|$ values above are transcribed directly from the underlying gradient-orthogonality data (referenced in the experimental-setup section). Cells show $|\cos|$ rounded to three decimal places. For \texttt{TFGN\_GPT2M\_FS}, distributional summary statistics are reported in lieu of a per-pair $6\times 6$ matrix; the full matrix is reserved under NDA along with the per-layer decompositions and the routing-state JSONs.

\section{Full PPL Matrices with bwt\_d Rows}
\label{app:ppl}

The PPL matrices for the eleven primary conditions are presented in §\ref{sec:results:1pagers} and §\ref{sec:results:baselines} of the main results section, where they are paired with their respective per-condition narratives. Cross-referencing them in this appendix would duplicate body content.

\textbf{Pointers:} \texttt{TFGN\_LLAMA8B\_RETROFIT} matrix in §\ref{sec:results:llama8b_retrofit}; \texttt{TFGN\_LLAMA8B\_FS} in §\ref{sec:results:llama8b_fs}; \texttt{TFGN\_GPT2M\_FS} in §\ref{sec:results:gpt2m_fs}; \texttt{TFGN\_GPT2M\_RETROFIT} in §\ref{sec:results:gpt2m_rf}; \texttt{TFGN\_GPT2S\_FS} in §\ref{sec:results:gpt2s_fs}; the five matched baselines in §\ref{sec:results:baselines}.

\textbf{Reading rule for every PPL matrix in this paper:} rows are phase trained (P1, P2, \ldots), columns are evaluation domains. Cell $M[t, d]$ is the held-out perplexity on domain $d$ after phase $t$. The bottom row reports per-domain $\text{bwt}_d$ for the trained domains. The diagonal is the just-trained PPL; off-diagonal cells in the trained-domain columns are forgetting (or, for evaluation-only columns, out-of-distribution evaluation PPL not CL forgetting).

\section{Extension A Canonical Values and 11-Condition Matrix}
\label{app:exta}

The full per-condition Extension A BWT/FM table is in §\ref{sec:exta:all11} (Table~\ref{tab:exta_canonical}). The three-axis decomposition of the 81\% headline reduction is in §\ref{sec:exta:headline} (Table~\ref{tab:exta_decomp}). The per-phase BWT decomposition for the three reference conditions is in Table~\ref{tab:exta_perphase}.

\textbf{Configuration matrix.} Every Extension A condition pairs a routing substrate (basic vs enhanced) with a self-regulation stack subset (none / sensing+gating / internal-world-model-only / full-stack-diagnostic / full-stack-active). The full configuration assignment per condition is reserved; access terms are in \S\ref{sec:limitations:nda}. The numerical effect of the assignment on BWT is reported in Table~\ref{tab:exta_canonical}.

\textbf{Per-layer gradient orthogonality.} All twelve layers of the GPT-2 Small backbone report gradient orthogonal-fraction $> 0.995$ on every Extension A condition --- orthogonality is near-uniform across depth. The full per-layer decomposition is in the gradient-orthogonality JSON artifact reserved under NDA.

\section{Tier A 200\,M Per-Condition PPL Matrices}
\label{app:exta_tier_a_ppl}

The four publicly-disclosed Tier A 200\,M-tokens-per-phase conditions are reported here for completeness. They run on Prose~$\to$~Python~$\to$~Math at GPT-2 Small ($\sim$398\,M total) with the basic-routing main-paper substrate; they differ in the self-regulation stack subset enabled. The 1\,B-token Tier B and Tier C matrices are in §\ref{sec:exta:ppl_matrices_main}; the canonical BWT/FM scalars across all eleven Extension A conditions are in Table~\ref{tab:exta_canonical}.

\begin{table}[!htbp]
\centering
\scriptsize
\setlength{\tabcolsep}{3pt}
\renewcommand{\arraystretch}{1.05}
\caption{\texttt{TFGN\_EXTA\_A\_BASELINE} (Tier A matched control, base-consolidation only) --- per-domain PPL matrix and scalar metrics (3-phase Prose~$\to$~Python~$\to$~Math, 200\,M tokens/phase). \textbf{$\text{BWT}_3 = -0.00942$}.}
\label{tab:exta_tier_a_a_baseline}

\begin{tabular}{lrrr}
\toprule
\textbf{Phase trained} & Prose & Python & Math \\
\midrule
P1 Prose      & 77.75 & 51.63 & 126.56 \\
P2 Python     & 77.75 & 32.93 & 126.32 \\
P3 Math       & 77.79 & 33.53 & 116.46 \\
\midrule
$\text{bwt}_d$ & $-0.00051$ & $-0.01822$ & --- \\
\bottomrule
\end{tabular}%
\\[0.4em]

\begin{tabular}{p{0.30\textwidth}p{0.62\textwidth}}
\toprule
\textbf{Metric} & \textbf{Value} \\
\midrule
$\text{BWT}_3$ / FM & $-0.00942$ / $+0.00942$ \\
Per-domain $\text{bwt}_d$ & Prose $-0.00051$;\; Python $-0.01822$ \\
Other & Mean$|\cos|$ 0.0436; L2-orth 99.905\%; HellaSwag P1$\to$P3: 0.268 $\to$ 0.274 ($+2.2\%$, maintained) \\
\bottomrule
\end{tabular}%

\end{table}

\begin{table}[!htbp]
\centering
\scriptsize
\setlength{\tabcolsep}{3pt}
\renewcommand{\arraystretch}{1.05}
\caption{\texttt{TFGN\_EXTA\_A\_SENSEACT} (sensing + gating pair, minimal) --- per-domain PPL matrix and scalar metrics (3-phase Prose~$\to$~Python~$\to$~Math, 200\,M tokens/phase). \textbf{$\text{BWT}_3 = -0.00528$}.}
\label{tab:exta_tier_a_a_senseact}

\begin{tabular}{lrrr}
\toprule
\textbf{Phase trained} & Prose & Python & Math \\
\midrule
P1 Prose      & 73.40 & 45.32 & 118.99 \\
P2 Python     & 73.40 & 31.86 & 118.69 \\
P3 Math       & 73.46 & 32.17 & 112.08 \\
\midrule
$\text{bwt}_d$ & $-0.00082$ & $-0.00973$ & --- \\
\bottomrule
\end{tabular}%
\\[0.4em]

\begin{tabular}{p{0.30\textwidth}p{0.62\textwidth}}
\toprule
\textbf{Metric} & \textbf{Value} \\
\midrule
$\text{BWT}_3$ / FM & $-0.00528$ / $+0.00528$ \\
Per-domain $\text{bwt}_d$ & Prose $-0.00082$;\; Python $-0.00973$ \\
Other & HellaSwag P1$\to$P3: 0.268 $\to$ 0.264 ($-1.5\%$, mildly degraded) \\
\bottomrule
\end{tabular}%

\end{table}

\begin{table}[!htbp]
\centering
\scriptsize
\setlength{\tabcolsep}{3pt}
\renewcommand{\arraystretch}{1.05}
\caption{\texttt{TFGN\_EXTA\_A\_FULL\_DIAG} (full stack, diagnostic consolidation; critical differentiator) --- per-domain PPL matrix and scalar metrics (3-phase Prose~$\to$~Python~$\to$~Math, 200\,M tokens/phase). \textbf{$\text{BWT}_3 = -0.01041$}.}
\label{tab:exta_tier_a_a_full_diag}

\begin{tabular}{lrrr}
\toprule
\textbf{Phase trained} & Prose & Python & Math \\
\midrule
P1 Prose      & 68.87 & 39.66 & 107.95 \\
P2 Python     & 68.87 & 27.12 & 107.52 \\
P3 Math       & 68.88 & 27.67 & 100.24 \\
\midrule
$\text{bwt}_d$ & $-0.00015$ & $-0.02028$ & --- \\
\bottomrule
\end{tabular}%
\\[0.4em]

\begin{tabular}{p{0.30\textwidth}p{0.62\textwidth}}
\toprule
\textbf{Metric} & \textbf{Value} \\
\midrule
$\text{BWT}_3$ / FM & $-0.01041$ / $+0.01041$ \\
Per-domain $\text{bwt}_d$ & Prose $-0.00015$;\; Python $-0.02028$ \\
Other & HellaSwag P1$\to$P3: 0.286 $\to$ 0.294 ($+2.8\%$, maintained) \\
\bottomrule
\end{tabular}%

\end{table}

\begin{table}[!htbp]
\centering
\scriptsize
\setlength{\tabcolsep}{3pt}
\renewcommand{\arraystretch}{1.05}
\caption{\texttt{TFGN\_EXTA\_A\_CHAMPION} (\textbf{Tier A champion} --- full self-regulation, active consolidation) --- per-domain PPL matrix and scalar metrics (3-phase Prose~$\to$~Python~$\to$~Math, 200\,M tokens/phase). \textbf{$\text{BWT}_3 = -0.00277$}.}
\label{tab:exta_tier_a_a_champion}

\begin{tabular}{lrrr}
\toprule
\textbf{Phase trained} & Prose & Python & Math \\
\midrule
P1 Prose      & 69.06 & 37.92 & 106.49 \\
P2 Python     & 69.06 & 27.10 & 106.13 \\
P3 Math       & 69.07 & 27.24 & 101.99 \\
\midrule
$\text{bwt}_d$ & $-0.00014$ & $-0.00517$ & --- \\
\bottomrule
\end{tabular}%
\\[0.4em]

\begin{tabular}{p{0.30\textwidth}p{0.62\textwidth}}
\toprule
\textbf{Metric} & \textbf{Value} \\
\midrule
$\text{BWT}_3$ / FM & \textbf{$-0.00277$} / $+0.00277$ \\
Per-domain $\text{bwt}_d$ & Prose $-0.00014$;\; Python $-0.00517$ \\
Other & Mean$|\cos|$ 0.0526; \textbf{L2-orth 99.737\%}; HellaSwag P1$\to$P3: 0.280 $\to$ 0.290 (\textbf{$+3.6\%$, improved}) \\
\bottomrule
\end{tabular}%

\end{table}

\section{Definitions and Equations Index}
\label{app:eqindex}

This appendix is a one-stop index for every load-bearing symbol, metric, and equation referenced in the paper. Each entry has a first-use section reference.

\textbf{Evaluation metrics (first-use Appendix~\ref{app:metrics})}
\begin{itemize}
\item BWT (Lopez-Paz adapted to PPL) --- average per-domain relative degradation from just-trained PPL to final-phase PPL.
\item FM --- average per-domain relative degradation from just-trained PPL to worst observed PPL across later phases.
\item L2-orthogonal fraction --- $\sqrt{1 - \overline{|\cos|}^2}$ averaged across cross-domain pairs.
\item HellaSwag accuracy --- zero-shot likelihood-of-correct-completion on HellaSwag validation.
\end{itemize}

\textbf{Equations referenced in this paper}

\textbf{Load-bearing numerical constants}
\begin{itemize}
\item BWT$_{3}$ = $-0.007$ at LLaMA 3.1 8B Retrofit (tightest in paper) --- §\ref{sec:results:llama8b_retrofit}.
\item L2-orthogonal floor = 99.59\% (TFGN\_GPT2M\_RETROFIT) --- §\ref{sec:results:gradortho}.
\item L2-orthogonal ceiling = 99.94\% (TFGN\_GPT2M\_FS) --- §\ref{sec:results:gpt2m_fs}.
\item Extension A 81\% reduction (BWT $-0.01140$ vs anchor $-0.06010$) --- §\ref{sec:exta:headline}.
\item Extension B 99.96\% reshape (mean cosine 0.9996 over 30 pairs at $\sim$398\,M; 0.9995 at $\sim$739\,M) --- §\ref{sec:extb:reshape}.
\item Sub-task pair cosine ceiling = 0.87 (no Python or Math sub-task pair exceeds this) --- (detailed reference reserved; see §\ref{sec:limitations:nda}).
\end{itemize}


\begin{thebibliography}{99}
\setlength{\itemsep}{2pt}
\small

\bibitem[OpenAI(2023)]{openai_gpt4_2023}
OpenAI.
\newblock GPT-4 Technical Report.
\newblock \emph{arXiv:2303.08774}, 2023.

\bibitem[Lopez-Paz \& Ranzato(2017)]{lopezpaz_gem_2017}
D.~Lopez-Paz and M.~Ranzato.
\newblock Gradient Episodic Memory for Continual Learning.
\newblock \emph{NeurIPS}, 2017.

\bibitem[Dupoux et~al.(2026)]{dupoux_lecun_malik_2026}
E.~Dupoux, Y.~LeCun, and J.~Malik.
\newblock Why AI Systems Don't Learn and What to Do About It: Lessons on Autonomous Learning from Cognitive Science.
\newblock \emph{arXiv:2603.15381}, 2026.

\bibitem[Kirkpatrick et~al.(2017)]{kirkpatrick_ewc_2017}
J.~Kirkpatrick et~al.
\newblock Overcoming catastrophic forgetting in neural networks.
\newblock \emph{PNAS}, 2017.

\bibitem[Aljundi et~al.(2018)]{aljundi_mas_2018}
R.~Aljundi et~al.
\newblock Memory Aware Synapses: Learning what (not) to forget.
\newblock \emph{ECCV}, 2018.

\bibitem[Zenke et~al.(2017)]{zenke_si_2017}
F.~Zenke, B.~Poole, and S.~Ganguli.
\newblock Continual Learning Through Synaptic Intelligence.
\newblock \emph{ICML}, 2017.

\bibitem[Chaudhry et~al.(2019a)]{chaudhry_agem_2019}
A.~Chaudhry et~al.
\newblock Efficient Lifelong Learning with A-GEM.
\newblock \emph{ICLR}, 2019.

\bibitem[Chaudhry et~al.(2019b)]{chaudhry_er_2019}
A.~Chaudhry et~al.
\newblock On Tiny Episodic Memories in Continual Learning.
\newblock \emph{arXiv:1902.10486}, 2019.

\bibitem[Buzzega et~al.(2020)]{buzzega_der_2020}
P.~Buzzega et~al.
\newblock Dark Experience for General Continual Learning: A Strong, Simple Baseline.
\newblock \emph{NeurIPS}, 2020.

\bibitem[Aljundi et~al.(2019)]{aljundi_mir_2019}
R.~Aljundi et~al.
\newblock Online Continual Learning with Maximally Interfered Retrieval.
\newblock \emph{NeurIPS}, 2019.

\bibitem[Farajtabar et~al.(2020)]{farajtabar_ogd_2020}
M.~Farajtabar et~al.
\newblock Orthogonal Gradient Descent for Continual Learning.
\newblock \emph{AISTATS}, 2020.

\bibitem[Saha et~al.(2021)]{saha_gpm_2021}
G.~Saha, I.~Garg, and K.~Roy.
\newblock Gradient Projection Memory for Continual Learning.
\newblock \emph{ICLR}, 2021.

\bibitem[Wang et~al.(2021)]{wang_adamnscl_2021}
S.~Wang et~al.
\newblock Training Networks in Null Space of Feature Covariance for Continual Learning.
\newblock \emph{CVPR}, 2021.

\bibitem[Mallya \& Lazebnik(2018)]{mallya_packnet_2018}
A.~Mallya and S.~Lazebnik.
\newblock PackNet: Adding Multiple Tasks to a Single Network by Iterative Pruning.
\newblock \emph{CVPR}, 2018.

\bibitem[Mallya et~al.(2018)]{mallya_piggyback_2018}
A.~Mallya, D.~Davis, and S.~Lazebnik.
\newblock Piggyback: Adapting a Single Network to Multiple Tasks by Learning to Mask Weights.
\newblock \emph{ECCV}, 2018.

\bibitem[Serra et~al.(2018)]{serra_hat_2018}
J.~Serra et~al.
\newblock Overcoming Catastrophic Forgetting with Hard Attention to the Task.
\newblock \emph{ICML}, 2018.

\bibitem[Rusu et~al.(2016)]{rusu_progressive_2016}
A.~Rusu et~al.
\newblock Progressive Neural Networks.
\newblock \emph{arXiv:1606.04671}, 2016.

\bibitem[Hu et~al.(2022)]{hu_lora_2022}
E.~Hu et~al.
\newblock LoRA: Low-Rank Adaptation of Large Language Models.
\newblock \emph{ICLR}, 2022.

\bibitem[Wang et~al.(2023)]{olora_2024}
X.~Wang et~al.
\newblock Orthogonal Subspace Learning for Language Model Continual Learning (O-LoRA).
\newblock \emph{Findings of EMNLP}, 2023. arXiv:2310.14152.

\bibitem[Qian et~al.(2025)]{treelora_2025}
Y.-Y. Qian, Y.-Z. Xu, Z.-Y. Zhang, P.~Zhao, and Z.-H. Zhou.
\newblock TreeLoRA: Efficient Continual Learning via Layer-Wise LoRAs Guided by a Hierarchical Gradient-Similarity Tree.
\newblock \emph{ICML}, 2025. arXiv:2506.10355.

\bibitem[Hoy \& Celik(2025)]{stable_2025}
W.~Hoy and N.~Celik.
\newblock STABLE: Gated Continual Learning for Large Language Models.
\newblock \emph{arXiv:2510.16089}, 2025.

\bibitem[Chen et~al.(2024)]{longlora_2024}
Y.~Chen et~al.
\newblock LongLoRA: Efficient Fine-tuning of Long-Context Large Language Models.
\newblock \emph{ICLR}, 2024.

\bibitem[Chen et~al.(2023)]{lifelong_moe_2023}
W.~Chen et~al.
\newblock Lifelong Language Pretraining with Distribution-Specialized Experts (Lifelong-MoE).
\newblock \emph{ICML}, 2023. arXiv:2305.12281.

\bibitem[Liu et~al.(2024)]{liu_loramoe_2024}
Q.~Liu et~al.
\newblock LoRAMoE: Revolutionizing Mixture of Experts for Maintaining World Knowledge in Language Model Alignment.
\newblock \emph{ACL}, 2024.

\bibitem[Smith et~al.(2023)]{coda_prompt_2023}
J.~Smith et~al.
\newblock CODA-Prompt: Continual Decomposed Attention-based Prompting for Rehearsal-Free Continual Learning.
\newblock \emph{CVPR}, 2023.

\bibitem[von Oswald et~al.(2020)]{vonoswald_hypernet_2020}
J.~von Oswald et~al.
\newblock Continual Learning with Hypernetworks.
\newblock \emph{ICLR}, 2020.

\bibitem[Beaulieu et~al.(2020)]{beaulieu_anml_2020}
S.~Beaulieu et~al.
\newblock Learning to Continually Learn (ANML).
\newblock \emph{ECAI}, 2020.

\bibitem[Javed \& White(2019)]{javed_oml_2019}
K.~Javed and M.~White.
\newblock Meta-Learning Representations for Continual Learning (OML).
\newblock \emph{NeurIPS}, 2019.

\bibitem[Miconi et~al.(2018)]{miconi_difplas_2018}
T.~Miconi, K.~Stanley, and J.~Clune.
\newblock Differentiable Plasticity: Training plastic neural networks with backpropagation.
\newblock \emph{ICML}, 2018.

\bibitem[Miconi et~al.(2020)]{miconi_backpropamine_2020}
T.~Miconi, A.~Rawal, J.~Clune, and K.~Stanley.
\newblock Backpropamine: Training self-modifying neural networks with differentiable neuromodulated plasticity.
\newblock \emph{ICLR}, 2020.

\bibitem[Rodriguez et~al.(2022)]{rodriguez_stpn_2022}
H.~Rodriguez et~al.
\newblock Short-Term Plasticity Neurons Learning to Learn and Forget.
\newblock \emph{ICML}, 2022. arXiv:2206.14048.

\bibitem[Miconi \& Kay(2025)]{miconi_kay_natneuro_2025}
T.~Miconi and K.~Kay.
\newblock Neural mechanisms of relational learning and fast knowledge reassembly in plastic neural networks.
\newblock \emph{Nature Neuroscience}, 28:406--414, 2025. doi:10.1038/s41593-024-01852-8.

\bibitem[Dohare et~al.(2024)]{dohare_sutton_nature_2024}
S.~Dohare et~al.
\newblock Loss of plasticity in deep continual learning.
\newblock \emph{Nature}, 2024.

\bibitem[Meng et~al.(2022)]{meng_rome_2022}
K.~Meng et~al.
\newblock Locating and Editing Factual Associations in GPT (ROME).
\newblock \emph{NeurIPS}, 2022.

\bibitem[Meng et~al.(2023)]{meng_memit_2023}
K.~Meng et~al.
\newblock Mass-Editing Memory in a Transformer (MEMIT).
\newblock \emph{ICLR}, 2023.

\bibitem[Jiang et~al.(2024)]{nse_2024}
H.~Jiang et~al.
\newblock Neuron-Level Sequential Editing for Large Language Models.
\newblock \emph{ACL}, 2025. arXiv:2410.04045.

\bibitem[Wang et~al.(2024)]{wise_2024}
P.~Wang et~al.
\newblock WISE: Rethinking the Knowledge Memory for Lifelong Model Editing of Large Language Models.
\newblock \emph{NeurIPS}, 2024.

\bibitem[Park et~al.(2025)]{make_2024}
S.~Park, S.~Park, J.~Kim, and H.~Kim.
\newblock MAKE: Memory-Associated Knowledge Editing.
\newblock \emph{Transactions of the Association for Computational Linguistics}, 13:938--952, 2025. doi:10.1162/TACL.a.26.

\bibitem[Wang et~al.(2026)]{hiedit_2026}
Y.~Wang, T.~Sun, C.~Tang, et~al.
\newblock HiEdit: Lifelong Model Editing with Hierarchical Reinforcement Learning.
\newblock \emph{arXiv:2604.11214}, 2026.

\bibitem[Shi et~al.(2025)]{shi_clllm_csur_2025}
H.~Shi et~al.
\newblock Continual Learning for Large Language Models: A Survey.
\newblock \emph{ACM Computing Surveys}, 2025.

\bibitem[Wang et~al.(2024)]{wang_clsurvey_csur_2024}
L.~Wang, X.~Zhang, H.~Su, and J.~Zhu.
\newblock A Comprehensive Survey of Continual Learning: Theory, Method and Application.
\newblock \emph{IEEE TPAMI}, 46(8):5362--5383, 2024. arXiv:2302.00487.

\bibitem[Imanov(2026)]{mech_forgetting_2026}
O.~Y.~L. Imanov.
\newblock Mechanistic Analysis of Catastrophic Forgetting in Large Language Models During Continual Fine-Tuning.
\newblock \emph{arXiv:2601.18699}, 2026.

\bibitem[Li \& Lee(2024)]{examining_forgetting_2024}
C.-A. Li and H.-Y. Lee.
\newblock Examining Forgetting in Continual Pre-training of Aligned Large Language Models.
\newblock \emph{arXiv:2401.03129}, 2024.

\bibitem[Chen et~al.(2024)]{llama3_syne_2024}
J.~Chen, Z.~Chen, J.~Wang, K.~Zhou, Y.~Zhu, J.~Jiang, Y.~Min, W.~X.~Zhao, et~al.
\newblock Towards Effective and Efficient Continual Pre-training of Large Language Models (Llama-3-SynE).
\newblock \emph{arXiv:2407.18743}, 2024.

\bibitem[Abbes et~al.(2025)]{revisit_replay_2025}
I.~Abbes, G.~Subbaraj, M.~Riemer, et~al.
\newblock Revisiting Replay and Gradient Alignment for Continual Pre-Training of Large Language Models.
\newblock \emph{arXiv:2508.01908}, 2025.

\bibitem[\v{S}liogeris et~al.(2025)]{ewc_gemma2_2025}
V.~\v{S}liogeris, P.~Daniu\v{s}is, and A.~Nakvosas.
\newblock Full-Parameter Continual Pretraining of Gemma2: Insights into Fluency and Domain Knowledge.
\newblock \emph{arXiv:2505.05946}, 2025.

\bibitem[Wang et~al.(2023)]{trace_2023}
X.~Wang, Y.~Zhang, T.~Chen, S.~Gao, S.~Jin, X.~Yang, Z.~Xi, R.~Zheng, Y.~Zou, T.~Gui, Q.~Zhang, X.~Huang.
\newblock TRACE: A Comprehensive Benchmark for Continual Learning in Large Language Models.
\newblock \emph{arXiv:2310.06762}, 2023.

\bibitem[Zellers et~al.(2019)]{hellaswag_2019}
R.~Zellers et~al.
\newblock HellaSwag: Can a Machine Really Finish Your Sentence?
\newblock \emph{ACL}, 2019.

\bibitem[Penedo et~al.(2024)]{penedo_fineweb_2024}
G.~Penedo, H.~Kydl\'i\v{c}ek, L.~Ben~Allal, A.~Lozhkov, M.~Mitchell, C.~Raffel, L.~von~Werra, and T.~Wolf.
\newblock The FineWeb Datasets: Decanting the Web for the Finest Text Data at Scale.
\newblock \emph{arXiv:2406.17557}, 2024.

\bibitem[Li et~al.(2023)]{li_starcoder_2023}
R.~Li, L.~Ben~Allal, Y.~Zi, N.~Muennighoff, D.~Kocetkov, C.~Mou, M.~Marone, et~al.
\newblock StarCoder: may the source be with you!
\newblock \emph{arXiv:2305.06161}, 2023.
\newblock The StarCoderData training corpus is the deduplicated, decontaminated derivative of The Stack used here for both Python and JavaScript.

\bibitem[Paster et~al.(2023)]{paster_openwebmath_2023}
K.~Paster, M.~Dos~Santos, Z.~Azerbayev, and J.~Ba.
\newblock OpenWebMath: An Open Dataset of High-Quality Mathematical Web Text.
\newblock \emph{arXiv:2310.06786}, 2023.

\bibitem[Sayers et~al.(2024)]{sayers_pubmed_2024}
E.~W. Sayers, J.~Beck, E.~E. Bolton, J.~R. Brister, J.~Chan, D.~C. Comeau, et~al.
\newblock Database resources of the National Center for Biotechnology Information in 2024.
\newblock \emph{Nucleic Acids Research}, 52(D1):D33--D43, 2024.

\bibitem[Nguyen et~al.(2024)]{nguyen_culturax_2024}
T.~Nguyen, C.~Van~Nguyen, V.~Lai, H.~Man, N.~T. Ngo, F.~Dernoncourt, R.~A. Rossi, and T.~H. Nguyen.
\newblock CulturaX: A Cleaned, Enormous, and Multilingual Dataset for Large Language Models in 167 Languages.
\newblock \emph{arXiv:2309.09400}, LREC-COLING 2024.

\bibitem[Wei et~al.(2022)]{cot_2022}
J.~Wei, X.~Wang, D.~Schuurmans, M.~Bosma, B.~Ichter, F.~Xia, E.~Chi, Q.~V. Le, and D.~Zhou.
\newblock Chain-of-Thought Prompting Elicits Reasoning in Large Language Models.
\newblock \emph{NeurIPS}, 2022. arXiv:2201.11903.

\bibitem[Yao et~al.(2023)]{tot_2023}
S.~Yao, D.~Yu, J.~Zhao, I.~Shafran, T.~Griffiths, Y.~Cao, and K.~Narasimhan.
\newblock Tree of Thoughts: Deliberate Problem Solving with Large Language Models.
\newblock \emph{NeurIPS}, 2023. arXiv:2305.10601.

\bibitem[Hao et~al.(2024)]{coconut_2024}
S.~Hao, S.~Sukhbaatar, D.~Su, X.~Li, Z.~Hu, J.~Weston, and Y.~Tian.
\newblock Training Large Language Models to Reason in a Continuous Latent Space (Coconut).
\newblock arXiv:2412.06769, 2024.

\bibitem[Hafner et~al.(2025)]{dreamerv3}
D.~Hafner, J.~Pasukonis, J.~Ba, and T.~Lillicrap.
\newblock Mastering Diverse Control Tasks through World Models.
\newblock \emph{Nature}, 640:647--653, 2025. arXiv:2301.04104.

\bibitem[Schrittwieser et~al.(2020)]{muzero}
J.~Schrittwieser, I.~Antonoglou, T.~Hubert, K.~Simonyan, L.~Sifre, et~al.
\newblock Mastering Atari, Go, Chess and Shogi by Planning with a Learned Model (MuZero).
\newblock \emph{Nature}, 588(7839):604--609, 2020. arXiv:1911.08265.

\bibitem[LeCun(2022)]{jepa_lecun_2022}
Y.~LeCun.
\newblock A Path Towards Autonomous Machine Intelligence (JEPA).
\newblock \emph{OpenReview}, Version 0.9.2, 2022.

\bibitem[Assran et~al.(2025)]{vjepa2_2025}
M.~Assran, A.~Bardes, D.~Fan, Q.~Garrido, R.~Howes, et~al., and Y.~LeCun.
\newblock V-JEPA 2: Self-Supervised Video Models Enable Understanding, Prediction and Planning.
\newblock arXiv:2506.09985, 2025.

\bibitem[Janner et~al.(2022)]{diffuser_2022}
M.~Janner, Y.~Du, J.~B. Tenenbaum, and S.~Levine.
\newblock Planning with Diffusion for Flexible Behavior Synthesis (Diffuser).
\newblock \emph{ICML}, 2022. arXiv:2205.09991.

\bibitem[Fedus et~al.(2022)]{fedus_switch_2022}
W.~Fedus, B.~Zoph, and N.~Shazeer.
\newblock Switch Transformers: Scaling to Trillion Parameter Models with Simple and Efficient Sparsity.
\newblock \emph{JMLR}, 23, 2022. arXiv:2101.03961.

\bibitem[Dai et~al.(2024)]{dai_deepseekmoe_2024}
D.~Dai, C.~Deng, C.~Zhao, R.~X. Xu, H.~Gao, et~al.
\newblock DeepSeekMoE: Towards Ultimate Expert Specialization in Mixture-of-Experts Language Models.
\newblock \emph{ACL}, 2024. arXiv:2401.06066.

\bibitem[Turner et~al.(2023)]{actadd_2023}
A.~M. Turner, L.~Thiergart, D.~Udell, G.~Leech, U.~Mini, and M.~MacDiarmid.
\newblock Activation Addition: Steering Language Models Without Optimization.
\newblock arXiv:2308.10248, 2023.

\bibitem[Zou et~al.(2023)]{repe_2023}
A.~Zou, L.~Phan, S.~Chen, J.~Campbell, P.~Guo, et~al., D.~Song, M.~Fredrikson, J.~Z. Kolter, and D.~Hendrycks.
\newblock Representation Engineering: A Top-Down Approach to AI Transparency.
\newblock arXiv:2310.01405, 2023.

\bibitem[Li et~al.(2023)]{li2023iti}
K.~Li, O.~Patel, F.~Vi\'egas, H.~Pfister, and M.~Wattenberg.
\newblock Inference-Time Intervention: Eliciting Truthful Answers from a Language Model.
\newblock \emph{NeurIPS}, 2023. arXiv:2306.03341.

\bibitem[Todd et~al.(2024)]{function_vectors_2024}
E.~Todd, M.~L. Li, A.~Sen Sharma, A.~Mueller, B.~C. Wallace, and D.~Bau.
\newblock Function Vectors in Large Language Models.
\newblock \emph{ICLR}, 2024. arXiv:2310.15213.

\bibitem[Templeton et~al.(2024)]{templeton2024scaling}
A.~Templeton, T.~Conerly, J.~Marcus, J.~Lindsey, T.~Bricken, et~al., and T.~Henighan.
\newblock Scaling Monosemanticity: Extracting Interpretable Features from Claude 3 Sonnet.
\newblock \emph{Transformer Circuits Thread}, Anthropic, May 2024.

\bibitem[Panickssery et~al.(2024)]{panickssery2024}
N.~Panickssery, N.~Rimsky, M.~Gabrieli, J.~Schulz, M.~Tong, E.~Hubinger, and A.~M. Turner.
\newblock Steering Llama 2 via Contrastive Activation Addition (CAA).
\newblock \emph{ACL}, 2024. arXiv:2312.06681.

\bibitem[Arditi et~al.(2024)]{arditi2024}
A.~Arditi, O.~Obeso, A.~Syed, D.~Paleka, N.~Panickssery, W.~Gurnee, and N.~Nanda.
\newblock Refusal in Language Models is Mediated by a Single Direction.
\newblock \emph{NeurIPS}, 2024. arXiv:2406.11717.

\bibitem[Zhang et~al.(2025)]{do_latent_tokens_think_2025}
Y.~Zhang, B.~Tang, T.~Ju, S.~Duan, and G.~Liu.
\newblock Do Latent Tokens Think? A Causal and Adversarial Analysis of Chain-of-Continuous-Thought.
\newblock arXiv:2512.21711, 2025.

\end{thebibliography}
\end{document}